\pdfoutput=1

\documentclass[twoside,11pt]{article}

\usepackage{jmlr2e}

\usepackage{microtype}
\usepackage{booktabs} 

\usepackage{algorithm}
\usepackage{algorithmic}
\usepackage{color}
\usepackage{enumitem}
\usepackage{caption,subcaption}

\usepackage[jmlr,parskip=false]{autoconfig}

\AtBeginDocument{
    \theoremstyle{theorem}
    \newtheorem{problem}{Problem}[section]
}

\def\II{\mathbb{I}}

\def\Bset{\mathcal{B}}
\def\Cset{\mathcal{C}}

\def\Eset{\mathcal{E}}
\def\Iset{\mathcal{I}}
\def\Sset{\mathcal{S}}
\def\Xset{\mathcal{X}}

\def\gmax{\gamma_{\max{}}}
\def\gmin{\gamma_{\min{}}}

\def\Xtilde{\bX}

\def\bA{{\bf A}}
\def\bB{{\bf B}}
\def\bD{{\bf D}}

\def\bO{{\bf O}}
\def\bP{{\bf P}}

\def\bU{{\bf U}}
\def\bV{{\bf V}}
\def\bX{{\bf X}}
\def\bZ{{\bf Z}}
\def\bSigma{{\boldsymbol \Sigma}}
\def\tSigma{\overline{\boldsymbol \Sigma}}

\def\Eye{{\bf I}}
\def\Param{{\bf B^\star}}
\def\ParamBound{{b^\star}}
\def\Cutoff{{\mathfrak{c}}}
\def\Loss{\mathcal{L}}
\def\CV{{\operatorname{cv}}}
\def\Estimator{\widehat{\bB}}
\def\EstimatorCV{\Estimator_\CV}
\def\SampleDistr{\Pi}
\def\SamplingOp{\mathfrak{X}}
\def\Noise{\varepsilon}
\def\Rademacher{\zeta}
\def\SpikeNorm{\mathfrak{N}}
\def\ErrorCV{\widehat{E}}
\def\Penalty{\lambda}
\def\PenaltyRand{\widehat{\lambda}}
\def\PenaltyCV{\PenaltyRand_\CV}

\def\bUhat{\widehat{\bU}}
\def\bVhat{\widehat{\bV}}

\def\lambdaellzero{\lambda_{\ell_0}}

\def\barNoise{\bar{\Noise}}

\def\tW{\widetilde{W}}

\NewNorm\NormLTP{_{L^2(\SampleDistr)}}
\NewNorm\NormSGB{_{\psi_2(\SampleDistr)}}

\hypersetup{hidelinks}

\usepackage{lastpage}
\jmlrheading{23}{2022}{1-\pageref{LastPage}}{9/19; Revised
6/22}{7/22}{19-795}{Nima Hamidi and Mohsen Bayati}

\ShortHeadings{On Low-rank Trace Regression}{Hamidi and Bayati}
\firstpageno{1}

\begin{document}

\title{On Low-rank Trace Regression under General Sampling Distribution}

\author{\name Nima Hamidi \email hamidi@stanford.edu \\
       \addr Department of Statistics\\
       Stanford University\\
       Stanford, CA 94305, USA
       \AND
       \name Mohsen Bayati \email bayati@stanford.edu \\
       \addr Graduate School of Business\\
       Stanford University\\
       Stanford, CA 94305, USA}

\editor{Inderjit Dhillon}

\maketitle

\begin{abstract}
In this paper, we study the trace regression when a matrix of parameters $\mathbf{B}^\star$ is estimated via the \emph{convex relaxation of a rank-regularized regression} or via \emph{regularized non-convex optimization}. It is known that these estimators satisfy near-optimal error bounds under assumptions on the rank, coherence, and spikiness of $\mathbf{B}^\star$. We start by introducing a general notion of spikiness for $\mathbf{B}^\star$ that provides a generic recipe to prove the \emph{restricted strong convexity} of the sampling operator of the trace regression and obtain near-optimal and non-asymptotic error bounds for the estimation error. Similar to the existing literature, these results require the regularization parameter to be above a certain \emph{theory-inspired} threshold that depends on observation noise that may be unknown in practice. Next, we extend the error bounds to cases where the regularization parameter is chosen via cross-validation. This result is significant in that existing theoretical results on cross-validated estimators
\citep{kale2011crossvalidation,kumar2013nearoptimal,aboumoustafa2017apriori} do not apply to our setting since the estimators we study are not known to satisfy their required notion of stability. Finally, using simulations on synthetic and real data, we show that the cross-validated estimator selects a near-optimal penalty parameter and outperforms the theory-inspired approach of selecting the parameter.
\end{abstract}


\section{Introduction}

We consider the problem of estimating an unknown parameter matrix $\Param\in\IR^{d_r\times d_c}$ from $n$ noisy observations
\begin{align}
Y_i = \Trace{\Param\bX_i^\top}+\Noise_i\,,\label{eq:prob-form1}
\end{align}
for $i=1,\ldots,n$ where each $\Noise_i\in \IR$ is a zero-mean noise and each $\bX_i\in\IR^{d_r\times d_c}$ is a known measurement matrix, sampled independently from a distribution $\SampleDistr$ over $\IR^{d_r\times d_c}$. We also assume the estimation problem is high-dimensional (when $n\ll d_r\times d_c$).

Over the past decade, this problem has been studied for several families of distributions $\SampleDistr$ that span a range of applications. It is constructive to look at the following four special cases of the problem:
\begin{itemize}
\item {\bf Matrix completion:} Let $\SampleDistr$ be the uniform distribution over the set of canonical basis matrices for $\IR^{d_r\times d_c}$, i.e., over the set of all matrices that have only a single non-zero entry that is equal to $1$. In this case we recover the well-known \emph{matrix completion} problem, which estimates $\Param$ when $n$ noisy observations of (uniformly randomly) selected entries are available \citep{candes2009exact,candes2010matrix,keshavan2009matrix,keshavan2010matrix}. A more general version of this problem is when $\SampleDistr$ is a non-uniform probability distribution over the basis matrices \citep{srebro2010collaborative,negahban2011estimation,klopp2014noisy}.

\item {\bf Multi-task learning:} When the support of $\SampleDistr$ includes only matrices that have a single non-zero row, then the problem reduces to the multi-task learning problem. Specifically, once can consider a setting with $n$ observations of multiple learning tasks, $d_r$ different supervised learning problems, that are represented by $d_r$ linear regression models with unknown $d_c$-dimensional parameters $B^*_1,\ldots,B^*_{d_r}$ that form rows of $\bB^*$. Equivalently, when the $i_r$-th row of $\bX_i$ is non-zero, we can assume that $Y_i$ is a noisy observation of the $i_r$-th task, with a feature vector equal to the $i_r$-th row of $\bX_i$. In multi-task learning the goal is to learn the parameters (matrix $\Param$) by leveraging the structural properties (similarities) of the tasks \citep{caruana1997multitask}.

\item {\bf Compressed sensing via Gaussian ensembles:} If we view the matrix as a high-dimensional vector of size $d_rd_c$, then the estimation problem can be viewed as an example of the compressed sensing problem, given certain structural assumptions on $\Param$. In this literature it is known that Gaussian ensembles, when each $\bX_i$ is a random matrix with entries filled with i.i.d. samples from $\Normal{0,1}$, are a suitable family of measurement matrices \citep{candes2011tight}.

\item {\bf Compressed sensing via factored measurements:} Consider the previous example. One drawback of the Gaussian ensembles is the need to store $n$ large matrices which requires memory of size $O(nd_rd_c)$.
    \cite{recht2010guaranteed} propose factored measurements to reduce this memory requirement. They suggest using rank-1 matrices $\bX_i$ of the form $UV^\top$, where $U\in\IR^{d_r}$ and $V\in\IR^{d_c}$ are random vectors that reduces the memory requirement to $O(nd_r+nd_c)$.
\end{itemize}

A popular estimator, using the observations in \cref{eq:prob-form1}, is given by solution to the following convex program,
\begin{align}
\min_{\bB\in\Sset}\frac{1}{n}\sum_{i=1}^{n}\left[Y_i-\Trace{\bB\bX_i^\top}\right]^2 + \lambda\NormNuc{\bB}\,,\label{eq:convex-opt-version-1}
\end{align}
where $\Sset\subseteq\IR^{d_r\times d_c}$ is an arbitrary convex set of matrices with $\Param\in\Sset$,
$\lambda$ is a regularization parameter, and $\NormNuc{\bB}$ is the trace norm of $\bB$ (defined in \cref {sec:notation}) that favors low-rank matrices. This type of estimator was initially introduced by \citet{candes2009exact} for the noise-free version of the matrix completion problem and was later expanded to more general cases. An admittedly incomplete list of follow-up work is \citep{candes2010power,mazumder2010spectral,gross2011recovering,recht2011simpler, rohde2011estimation, koltchinskii2011nuclear, negahban2011estimation, negahban2012restricted,klopp2014noisy}. Another class of estimators, studied by \cite{srebro2005generalization}, \cite{keshavan2009matrix}, and \cite{keshavan2010matrix}, replaces the variable $\bB$ in \cref{eq:convex-opt-version-1} by $\bU\bV^\top$, where $\bU$ and $\bV$ are explicitly low-rank matrices, and the trace-norm penalty is replaced with a ridge-type penalty term on entries of $\bU$ and $\bV$, as shown in \cref{eq:alt-min-loss} of \cref{subsec:estimator}. These two bodies of literature provide tail bounds for the corresponding estimators, under certain assumptions on the rank, coherence, and spikiness of $\Param$ for a few classes of sampling distributions $\SampleDistr$. We refer the reader to the detailed discussion of this literature in \citep{davenport2016overview,hastie2015statistical}, and the references therein.

\paragraph{Contributions.} Our paper extends the above literature, and makes the following contributions:

\begin{enumerate}
\item[(i)] We introduce a general notion of spikiness for $\Param$, and construct error bounds (building on the analysis of \cite{klopp2014noisy}) for the estimation error of a large family of estimators. Our main contribution is a general recipe for proving the well-known restricted strong convexity (RSC) condition, defined in \cref{sec:trace-reg}.

\item[(ii)] Next, we prove the first error bound (to the best of our knowledge) for the cross-validated version of the above family of estimators. Specifically, all bounds in for matrix estimation in the literature, as well as our bounds in (i), require the regularization parameter $\lambda$ to be larger than a constant multiple of the operator norm of $\sum_{i=1}^{n}\Noise_i\bX_i$, which is not feasible in practice due to lack of access to $\{\Noise_i\}_{i=1}^n$. In fact, instead of using these  theory-inspired estimators, practitioners select $\lambda$ via cross-validation. We prove that this cross-validated estimator satisfies similar error bounds to the ones in (i).
    We also show, via simulations on real and synthetic data, that the cross-validated estimator outperforms the theory-inspired estimators, and is nearly as good as the oracle estimator that chooses $\lambda$ based on its knowledge of $\bB^*$.

    We note that existing analyses of cross-validated estimators \citep[e.g.,][]{kale2011crossvalidation,kumar2013nearoptimal,aboumoustafa2017apriori} do not apply to our setting since they impose certain stability criteria on the estimation algorithm. However, establishing these criteria in our case is highly non-trivial for two reasons. First, we are studying a family of algorithms and not a specific algorithm. Second, it is not known whether stability holds even for a single low-rank matrix estimation method (including both convex and non-convex optimization).\footnote{In the convex relaxion case, it might be possible to analyze the cross-validated estimator by extending the analysis of \cite{chetverikov2016crossvalidated}  for the LASSO estimator.  But even if such an extension is possible, it would be only for a single estimator, and not the larger family of estimators we study here. In addition, it would be a long proof (the proof for the LASSO estimator is over 30 pages), whereas our proof is only a few pages.}

\item[(iii)] We apply our results from (i) to the four classes of problems discussed above and obtain the first such error bounds (to the best of our knowledge) for the multi-task learning problem. For matrix completion and both cases of compressed sensing (with Gaussian ensembles and with factored measurements), we obtain matching and near-optimal error bounds as the ones in the existing literature, e.g., by \cite{negahban2012restricted}, \cite{klopp2014noisy}, \cite{candes2011tight}, and \cite{cai2015rob}.

    We note that \cite{rohde2011estimation} and \cite{negahban2011estimation} also consider the
    trace regression problem under general sampling distributions. Moreover, they only provide error bounds for the estimation error when the corresponding sampling operator satisfies the restricted isometry property (RIP) or RSC. However, they do not prove whether RIP or RSC hold for the multi-task learning problem. In fact, \cite{rohde2011estimation} state that their analysis cannot prove RIP for the multi-task learning problem. 
    By contrast, we prove that RSC holds for all four classes of problems, by leveraging our unifying method of proving the RSC condition.

    \cite{koltchinskii2011nuclear} study the trace regression problem in its general form as well. However, they study a different estimator that is not practical to implement. Specifically, \cite{koltchinskii2011nuclear} propose the following  estimator: 
    \[
    \Estimator_{\text{KLT}}:=\Argmin_{\bB}\left\{\Expect[\big]{\Inner{\bB,\bX}^2} - \Inner{\frac{2}{n}\sum_{i=1}^{n}Y_i\bX_i,\bB} + \lambda\NormNuc{\bB}\right\}\,.
    \]
     While \cite{koltchinskii2011nuclear} show that $\Estimator_{\text{KLT}}$ enjoys a general sharp oracle inequality, they need knowledge of distribution of $\bX$ to calculate $\Expect{\Inner{\bB,\bX}^2}$.
    But, in our study (like majority of the literature on low-rank matrix estimation), the estimators do not need any knowledge of the distribution of $\bX$ and work with the data at hand, which is more practical.

    Finally, for Gaussian ensembles and factored measurements, when there is no noise, our results also demonstrate that $\bB^*$ can be exactly recovered when the number of observations is above a certain threshold. Our recovery thresholds match the ones by \cite{candes2009exact} and \cite{cai2015rob}.


\end{enumerate}

\paragraph{Organization of the paper.} We introduce additional notation and state the precise formulation of the problem in \cref{sec:notation}. In \cref{sec:trace-reg} we introduce a family of estimators and prove tail bounds on their estimation error. \cref{sec:cross-val} contains our results for the cross-validated estimator and corresponding numerical simulations. Application of our main error bounds to the aforementioned four classes of problems is given in \cref{sec:applications}, and exact recovery results are given in \cref{sec:exact-recovery}. Details of the proofs are discussed in \cref{sec:aux,sec:klopp-pf} and additional simulations are provided in \cref{sec:supp-simulations}.

\section{Notation and Problem Formulation}
\label{sec:notation}

Notation in bold roman capital letters (e.g., $\bA$) denote matrices and non-bold italic capital letters denote vectors (e.g., $V$). For any positive integer $m$, the notation $e_1,e_2,\ldots,e_m$ refers to the standard basis for $\IR^m$, and $\Eye_m$ is the $m\times m$ identity matrix. The \emph{trace inner product} of matrices $\bA_1$ and $\bA_2$ with the same dimensions is defined as
\begin{align*}
\Inner{\bA_1,\bA_2} := \Trace{\bA_1\bA_2^\top}.
\end{align*}
For $d_r\times d_c$ matrices $\bX_1,\bX_2,\cdots,\bX_n$, let the sampling operator $\SamplingOp:\IR^{d_r\times d_c}\to\IR^n$ be given by
\begin{align*}
\left[\SamplingOp(\bB)\right]_i:=\Inner{\bB,\bX_i}\qquad\text{for all $i\in[n]$,}
\end{align*}
where by $[k]$, we denote the set $\{1,2,\ldots,k\}$. For any two real numbers $a$ and $b$, $a\vee b$ and $a\wedge b$ denote $\max(a,b)$ and $\min(a,b)$ respectively. Also, a real-valued random variable $z$ is $\sigma$-sub-Gaussian if
$\Expect{\exp(\eta z)}\leq \exp(\sigma^2 z^2/2)$ for all $\eta\in \IR$.

For a norm\footnote{$\SpikeNorm$ can also be a semi-norm.} $\SpikeNorm:\Xset\to\IR^+\cup\{0\}$ defined on a vector space $\Xset$, let $\SpikeNorm^*:\Xset\to\IR^+\cup\{0,\infty\}$ be its \emph{dual norm} defined as
\begin{align*}
\SpikeNorm^*(X)=\sup_{\SpikeNorm(Y)\leq 1}\Inner{X,Y}\qquad\text{for all $X\in\Xset$.}
\end{align*}
In this paper, we use several different matrix norms. A brief explanation of these norms is as follows. Let $\bB$ be a matrix with $d_r$ rows and $d_c$ columns.
\begin{enumerate}
\item \textbf{$\mathbf{L}^{\boldsymbol{\infty}}$-norm} is defined by
$
\NormInf{\bB}:=\max_{(i,j)\in[d_r]\times[d_c]}\left\{\Abs{\bB_{ij}} \right\}
$.

\item \textbf{Frobenius norm} is defined as $\NormF{\bB}:=\sqrt{\sum_{(i,j)\in[d_r]\times[d_c]}\bB_{ij}^2}$.

\item \textbf{Operator norm} is defined as $\NormOp{\bB}:=\sup_{\NormTwo{V}=1\, \&\, V\in\IR^{d_c}}\NormTwo{\bB V}$.
An alternative definition of the operator norm is given by using the \emph{singular value decomposition} (SVD) of $\bB=\bU\bD\bV^\top$, where $\bD$ is an $r\times r$ diagonal matrix and $r$ denotes the rank of $\bB$. In this case, it is well known that
$\NormOp{\bB}=\bD_{11}$.

\item \textbf{Trace norm} is defined as $\NormNuc{\bB}:=\sum_{i=1}^{r}\bD_{ii}$.

\item \textbf{$\mathbf{L_{p,q}}$-norm}, for $p,q\ge 1$, is defined as $
\Norm{\bB}_{p,q}:=\left(\sum_{r=1}^{d_r}\left(\sum_{c=1}^{d_c}|B_{rc}|^p\right)^{q/p}\right)^{1/q}$.

\item \textbf{$\mathbf{L^2\boldsymbol{(\SampleDistr)}}$-norm} is defined as
$\NormLTP{\bB}:=\sqrt{\Expect{\Inner{\bB,\bX}^2}}$, where $\bX$ is sampled from a probability measure $\SampleDistr$ on $\IR^{d_r\times d_c}$.

\item \textbf{Exponential Orlicz norm} is defined for any $p\geq1$ and probability measure $\SampleDistr$ on $\IR^{d_r\times d_c}$ as
\begin{align*}
\Norm{\bB}_{\psi_p(\SampleDistr)}
&:=
\Norm{\Inner{\bB,\bX}}_{\psi_p} =
\inf\left\{t>0:\Expect{e^{\left(\frac{\Abs{\Inner{\bB,\bX}}}{t}\right)^p}-1}\leq1\right\},
\end{align*}
where $\bX$ has distribution $\SampleDistr$.

\end{enumerate}

Now, we will state the main trace regression problem that is studied in this paper.

\begin{problem}\label{prob:main} Let $\Param$ be an unknown $d_r\times d_c$ matrix that is low-rank and has real-valued entries, specifically, $r\ll\min(d_r,d_c)$. Moreover, assume that $\SampleDistr$ is a distribution on $\IR^{d_r\times d_c}$, and that $\bX_1,\bX_2,\cdots,\bX_n$ are i.i.d. samples from $\SampleDistr$ and their corresponding sampling operator is $\SamplingOp:\IR^{d_r\times d_c}\to\IR^n$. Our regression model is given by
\begin{align}
\label{eq:generative-model}
Y=\SamplingOp(\Param)+E,
\end{align}
where observation $Y$ and noise $E$ are both vectors in $\IR^n$. Elements of $E$ are denoted by $\Noise_1,\ldots,\Noise_n$, where $\{\Noise_i\}_{i=1}^n$ is a sequence of independent mean-zero random variables with variance at most $\sigma^2$. The goal is to estimate $\Param$ from the observations $Y$.
\end{problem}
We also use the following two notations: 
\[
\bSigma:=\frac{1}{n}\sum_{i=1}^{n}\Noise_i\Xtilde_i~~~~\text{and}~~~\bSigma_R := \frac{1}{n}\sum_{i=1}^{n}\Rademacher_i\bX_i\,,
\] 
where $\{\Rademacher_i\}_{i=1}^n$ is an i.i.d. sequence with a Rademacher distribution. 

\section{Estimation Method and Corresponding Tail Bounds}
\label{sec:trace-reg}

This section is dedicated to the tail bounds for the trace regression problem.
The results and the proofs in this section are based on those found in \citet{klopp2014noisy}, with a slightly sharper analysis. For the sake of completeness, all proofs are reproduced (adapted) for our setting and are presented in \cref{sec:klopp-pf}.

\subsection{General notions of spikiness} \label{subsec:g-spikiness-rank}

It is a well-known fact that, in Problem \ref{prob:main}, the low-rank assumption is not sufficient for estimating $\Param$ from the observations $Y$. For example, changing one entry of $\Param$ increases the rank of the matrix by (at most) 1, whereas it would be impossible to distinguish between these two matrices without observing the modified single entry. To remedy this difficulty, \citet{candes2009exact} and \cite{keshavan2009matrix} propose an incoherence assumption. If singular value decomposition (SVD) of $\Param$ is $\bU\bSigma\bV$, then the incoherence assumption roughly means that all rows of $\bU$ and $\bV$ have norms of the same order. Alternatively, \citet{negahban2012restricted} study the problem under a (less restrictive) assumption that bounds the \emph{spikiness} of the matrix $\Param$. Here, we define a general notion of spikiness as a matrix that
is sufficient for the estimation of $\Param$ under the general sampling distribution $\SampleDistr$, and
includes the spikiness studied by \citet{negahban2012restricted} as a special case. We define the \emph{spikiness} and \emph{low rankness} of a matrix $\bB\in\IR^{d_r\times d_c}$ as
\[
\text{spikiness of }\bB:=\frac{\SpikeNorm(\bB)}{\NormF{\bB}}
~~~~~~~~~~ \text{and} ~~~~~~~~~~
\text{low rankness of }\bB:=\frac{\NormNuc{\bB}}{\NormF{\bB}}\,.
\]
The spikiness used in \citet{negahban2012restricted} can be recovered by setting $\SpikeNorm(\bB)=\sqrt{d_rd_c}\Norm{\bB}_\infty$. This choice of norm, however, is not suitable for many distributions of $\bX$. For example, we will see in  \cref{subsec:bandit-regression} in the multi-task learning setting that
$\SpikeNorm(\bB)=2\sqrt{d}\Norm{\bB}_{2,\infty}$. In general, we will also show that
 in all cases, the exponential Orlicz norm can be used to guide the selection of $\SpikeNorm(\bB)$. Specifically, $\NormSGB{\cdot}$ is used in \crefrange{subsec:matrix-completion}{subsec:gauss-ensemble},  and $\Norm{\cdot}_{\psi_1(\SampleDistr)}$ is used in \cref{subsec:factored-meas}.

\subsubsection{Intuition for Selecting $\SpikeNorm$}

Here we provide some intuition for the use of the exponential Orlicz norm to select $\SpikeNorm$. \cite{negahban2011estimation} show how an error bound can be obtained from the RSC condition (defined in \cref{subsec:bounds} below) on a suitable set of matrices. The condition roughly requires that $
\Norm{\SamplingOp(\bB)}_2^2/n
\geq
\alpha\NormF{\bB}^2$ for a constant $\alpha$. Assume that random variables $\Inner{\bX_i,\bB}$ are not heavy-tailed; then $\Norm{\SamplingOp(\bB)}_2^2/n$ concentrates around its mean, $\NormLTP{\bB}$. The
Orlicz norm, which measures how heavy-tailed a distribution is, helps us construct a suitable ``constraint'' set of matrices where the aforementioned concentration holds simultaneously.

To be more concrete, consider the multi-tasking example (studied in \cref{subsec:bandit-regression}) for the  case of $d_r=d_c=d$. In particular, $\bX_i=e_iX_i^\top$, where $i$ is a uniformly random index in $[d]$ and  $X_i$ is a $d$ vector with i.i.d. $\Normal{0,d}$ entries.

In this example, we have $\NormLTP{\bB}=\NormF{\bB}^2$ for all $\bB\in\IR^{d\times d}$. Now, if a fixed $\bB$ is such that $\Inner{\bX_i,\bB}$ has a light tail, one can show that for sufficiently large $n$, due to concentration, $\Norm{\SamplingOp(\bB)}_2^2/n$ is at least $\NormF{\bB}^2/2$. 
To see this, consider two extreme cases: let $\bB_1$ be a matrix such that its first row has an $\ell^2$-norm equal to one and its remaining entries are zero, and let $\bB_2$ be a matrix such that all of its rows have an $\ell_2$-norm equal to $1/\sqrt d$. Both of these matrices have a Frobenious norm equal to one. But,  $\Norm{\SamplingOp(\bB_1)}_2^2$ has a heavier tail than $\Norm{\SamplingOp(\bB_2)}_2^2$, because $\Inner{\bX_i,\bB_1}$ is zero most of the time, but it is very large occasionally, whereas almost all realizations of $\Inner{\bX_i,\bB_2}$ are roughly of the same size. This intuition implies that matrices whose rows are almost the same size are more likely to satisfy RSC than the other ones. However, since $X_i^\top$ is invariant under rotation, one can see that the only thing that matters for RSC is the \emph{norm} of the rows. Indeed, the Orlicz norm verifies our intuition and after doing the computation (see \cref{subsec:bandit-regression} for details), one can see that $\Norm{\bB}_{\psi_2(\SampleDistr)}=O(\sqrt{d}\Norm{\bB}_{2,\infty})$. We will later see in \cref{subsec:bandit-regression} that by  defining $\SpikeNorm(\bB)$ to be a constant multiple of $\sqrt{d}\Norm{\bB}_{2,\infty}$, we obtain a suitable choice for $\Norm{\bB}$. Note that, for the matrix completion application, one can argue similarly that, in order for a matrix to satisfy the RSC condition with high probability, it cannot have a few very large rows. However, the second part of the above argument that uses rotation invariance does not apply, and actually a similar argument to the former one implies that each row cannot also have a few very large entries. Therefore, all entries of $\bB$ should be roughly of the same size, which would match the spikiness notion of \cite{negahban2012restricted}.

\subsection{Estimation} \label{subsec:estimator}

Before introducing the estimation approach, we state our first assumption.
\begin{asmp}
\label{as:x-tilde-boundedness-deterministic}
Assume that $\SpikeNorm(\Param)\leq \ParamBound$ for some $\ParamBound>0$.
\end{asmp}
Note that in \cref{as:x-tilde-boundedness-deterministic}, we  require only a bound on $\SpikeNorm(\Param)$ and \emph{not} the general spikiness of $\Param$.

Our theoretical results enjoy a certain notion of algorithmic independence. To make this point precise, we start by considering the trace-norm penalized least squares loss functions, also stated in a different format in \cref{eq:convex-opt-version-1},
\begin{align}
\Loss(\bB):=\frac{1}{n}\Norm{Y-\SamplingOp(\bB)}_2^2+\lambda\NormNuc{\bB}\,.\label{eq:main-loss}
\end{align}
However, we do not necessarily need to find the global minimum of \cref{eq:main-loss}. Let $\Sset\subseteq\IR^{d_r\times d_c}$ be an arbitrary convex set of matrices with $\Param\in\Sset$. All of our bounds are stated for any $\Estimator$ that satisfies
\begin{align}
\label{eq:goodness-condition}
\Estimator\in\Sset\quad\text{and}\quad\SpikeNorm(\Estimator)\leq \ParamBound\quad\text{and}\quad\Loss(\Estimator)\leq\Loss(\Param)\,.
\end{align}
While the global minimizer, $\Argmin_{\bB\in\Sset:\SpikeNorm(\bB)\leq b^*}\Loss(\bB)$, satisfies \cref{eq:goodness-condition}, we can also achieve \cref{eq:goodness-condition} by using other loss minimization problems. A notable example is the \emph{alternating minimization} approach that aims to solve
\begin{align}
(\bUhat,\bVhat)=\Argmin_{\bU\in\IR^{d_r\times r},\bV\in\IR^{d_c\times r},\bU\bV^\top\in\Sset}\frac{1}{n}\Norm{Y-\SamplingOp(\bU\bV^\top)}_2^2+\frac{\lambda}{2}\left(\NormF{\bU}^2+\NormF{\bV}^2\right)\,,\label{eq:alt-min-loss}
\end{align}
where $r$ is a preselected value for the rank. If we find the minimizer of \cref{eq:alt-min-loss}, then it is known that $\Estimator=\bUhat\bVhat^\top$ satisfies \cref{eq:goodness-condition}. For example, this fact is shown by  \cite{keshavan2009matrix} and \cite{mazumder2010spectral}.


\subsection{Restricted strong convexity and the tail bounds} \label{subsec:bounds}

The upper bound for the estimation error that will be stated next relies on a \emph{restricted strong convexity} (RSC) condition. This condition is defined below and will be shown to hold with high probability, under some assumptions on $\Param$ and $\SampleDistr$.
\begin{defn}[Restricted Strong Convexity Condition]\label{def:RSC}
 For a \emph{constraint} set $\Cset\subseteq\IR^{d_r\times d_c}$, we say that $\SamplingOp(\cdot)$ satisfies the RSC condition over the set $\Cset$ if there exist constants $\alpha(\SamplingOp)>0$ and $\beta(\SamplingOp)$ such that, for all $\bB\in\Cset$,
\begin{align*}
\frac{\Norm{\SamplingOp(\bB)}_2^2}{n}\geq\alpha(\SamplingOp)\NormF{\bB}^2-\beta(\SamplingOp)\,.
\end{align*}
\end{defn}
For the upper bound, we need the RSC condition to hold for a specific family of constraint sets that are parameterized by two positive parameters $\nu,\eta$. Define $\Cset(\nu,\eta)$ as
\begin{align}
\Cset(\nu,\eta)
:=
\left\{
\bB\in\IR^{d_r\times d_c}
~|~
\SpikeNorm(\bB)=1,~
\NormF{\bB}^2\geq\nu,~
\NormNuc{\bB}\leq\sqrt{\eta}\NormF{\bB}
\right\}.\label{eq:Cset-def}
\end{align}
The next result provides the upper bound on the estimation error when $\lambda$ is large enough and the RSC condition holds on $\Cset(\nu,\eta)$ for some constants $\alpha$ and $\beta$. 
\begin{thm}
\label{thm:deterministic-error-bound}
Let $\Param$ be a matrix of rank $r$ and define $\eta:=72r$. Also assume that $\SamplingOp(\cdot)$ satisfies the RSC condition for $\Cset(\nu,\eta)$ as defined  in \cref{def:RSC} with constants $\alpha=\alpha(\SamplingOp)$ and $\beta=\beta(\SamplingOp)$. In addition, assume that $\lambda=\lambda(n)$ is chosen such that
\begin{align}
\label{eq:lambda-condition}
\lambda\geq3\NormOp{\bSigma}\,,
\end{align}
where $\bSigma=\frac1n\sum_{i=1}^{n}\Noise_i\Xtilde_i$. Then, for any matrix $\Estimator$ satisfying \cref{eq:goodness-condition}, we have
\begin{align}
\NormF[\big]{\Estimator-\Param}^2
&\leq
\left(\frac{100\lambda^2r}{3\alpha^2}+\frac{8\ParamBound^2\beta}\alpha\right)\vee 4\ParamBound^2\nu\,.
\label{eq:trace-reg-bound}
\end{align}
\end{thm}
To simplify the notation, we will write $\lambda$ instead of $\lambda(n)$. The
proof of \cref{thm:deterministic-error-bound} is given in \cref{subsec:pf-thm-deterministic-error-bound}. However, here we provide a summary of the main steps to help in understanding the source of each term on the right-hand side of \cref{eq:trace-reg-bound}. First, using \cref{eq:goodness-condition} and \cref{eq:lambda-condition} and a series of algebraic manipulations, one can show that
$\Norm[\big]{\SamplingOp(\Param-\Estimator)}_2^2/n \le \lambda\sqrt{r}\NormF[\big]{\Estimator-\Param}$ holds, up to constants.
Combining this with the RSC condition for $(\Estimator-\Param)/\SpikeNorm(\Estimator-\Param)$ yields
the inequality $\alpha\NormF[\big]{\Estimator-\Param}^2-4\beta\ParamBound^2\leq \lambda\sqrt{r}\NormF[\big]{\Estimator-\Param}$, which can be translated, up to constants, into the term
\[
\frac{\lambda^2r}{\alpha^2}+\frac{\ParamBound^2\beta}\alpha
\]
in \cref{eq:trace-reg-bound}. However, to legitimize application of the RSC condition, one needs a lower bound $\sqrt{\nu}$ on the Frobenius norm of  $(\Estimator-\Param)/\SpikeNorm(\Estimator-\Param)$ to help show that $(\Estimator-\Param)/\SpikeNorm(\Estimator-\Param)$ is in $\Cset(\nu,\eta)$, where $\eta$ is a constant multiple of $r$. In the absence of this lower bound, we can directly use $\ParamBound^2\nu$ as an upper bound for $\alpha\NormF[\big]{\Estimator-\Param}^2$, which explains the last term on the right-hand side of \cref{eq:trace-reg-bound}.

%
Note that even though the assumptions of \cref{thm:deterministic-error-bound}
involve the noise and the observation matrices $\Xtilde_i$, no distributional assumption is required and the result is deterministic. However, we will employ probabilistic results later to show that the assumptions of the theorem hold. Specifically, we can show that \cref{eq:lambda-condition} holds with high probability, using a version of the Bernstein tail inequality for the operator norm of matrix martingales. This is stated as \cref{lem:freedman} in \cref{subsec:bernstein} and also appears as Proposition 11 of \cite{klopp2014noisy}.

The other condition in \cref{thm:deterministic-error-bound}, RSC for $\Cset(\nu,\eta)$, will be shown to hold with high probability by \cref{thm:rsc-condition} below. Before stating this result, we need two distributional assumptions for $\SamplingOp(\cdot)$. Recall that the distribution (over $\IR^{d_r\times d_c}$) from which our observation matrices $\{\bX_i\}_{i=1}^n$ are sampled is denoted by $\SampleDistr$.
\begin{asmp}
\label{as:gtmin-gtmax-bounds}
For positive constants $\gmin$ and $\gmax$, and for all $\bB\in\IR^{d_r\times d_c}$, the following holds:
\[
\gmin\NormF{\bB}^2
\leq
\Norm{\bB}_{L^2(\SampleDistr)}^2
\leq
\gmax\NormF{\bB}^2\,.
\]
\end{asmp}
As we will see later in this section, all upper bounds include $\gmin$ in the denominator. Therefore, \cref{as:gtmin-gtmax-bounds} provides a sufficient condition on the sampling distribution, which will make the estimation problem tractable. For example, in the special case of matrix completion, as demonstrated empirically by \cite{srebro2010collaborative} and \cite{foygel2011learning}, when some rows or columns of $\Param$ are sampled disproportionately more frequently, the estimation becomes more difficult. The assumption also resembles the bounds on leverage scores as shown by \cite{chen2015completing}. The condition involving $\gmax$ will be needed in \cref{sec:cross-val} for the analysis of the cross-validated estimator.
\begin{asmp}
\label{as:x-tilde-boundedness-stochastic}
There exists $\Cutoff>0$, such that
\begin{align*}
\Expect{\Inner{\bX,\bB}^2\cdot\II(\Abs{\Inner{\bX,\bB}}\leq\Cutoff)}\geq
\frac{1}{2}\Expect{\Inner{\bX,\bB}^2}\,,
\end{align*}
for all $\bB$ with $\SpikeNorm(\bB)\leq1$, where the expectations are with respect to $\SampleDistr$.
\end{asmp}
\cref{as:x-tilde-boundedness-stochastic} imposes a restriction on the sampling distribution such that the random variable $\Inner{\bX,\bB}$ will not have a heavy tail. This allows us to show that $\Inner{\bX,\bB}$ is not much smaller than its expectation, which in turn allows us to prove the RSC condition. In the special case of matrix completion, this assumption plays a similar role to the ``leveraged sampling''  assumption in \cite{chen2015completing}. The next remark shows that the variance of $\Inner{\bX,\bB}$ plays an important role in proving \cref{as:x-tilde-boundedness-stochastic}.
\begin{rem}
We will show later (in \cref{cor:orlicz-trun-second-moment-bound} of \cref{sec:aux}) that whenever $\Var{\Inner{\bX,\bB}}\approx1$ uniformly over $\bB$ with $\SpikeNorm(\bB)=1$, when $\SpikeNorm(\cdot)=\Norm{\cdot}_{\psi_p(\SampleDistr)}$ and $p\ge 1$, then $\Cutoff$ is a small constant that does not depend on the dimensions.
\end{rem}

The next result shows that a slightly more general form of the RSC condition holds with high probability.
\begin{thm}[Restricted Strong Convexity]
\label{thm:rsc-condition}
Define
\begin{align*}
\Cset'(\theta,\eta)
:=
\left\{
\bA\in\IR^{d_r\times d_c}
~|~
\SpikeNorm(\bA)=1,
\NormLTP{\bA}^2\geq\theta,
\NormNuc{\bA}\leq\sqrt{\eta}\NormF{\bA}
\right\}\,.
\end{align*}
If \cref{as:gtmin-gtmax-bounds,as:x-tilde-boundedness-stochastic} hold, then for all $\bA\in\Cset'(\theta,\eta)$, the inequality
\begin{align}
\frac1n\Norm{\SamplingOp(\bA)}_2^2\geq\frac14\Norm{\bA}_{L^2(\SampleDistr)}^2-\frac{93\eta\Cutoff^2}{\gmin}\Expect{\NormOp[\big]{\bSigma_R}}^2
\end{align}
holds with probability greater than $1-2\exp\left(-\frac{Cn\theta}{\Cutoff^2}\right)$ where $C>0$ is an absolute constant, provided that $Cn\theta>\Cutoff^2$, and the equality $\bSigma_R := \frac{1}{n}\sum_{i=1}^{n}\zeta_i\bX_i$ holds where $\{\zeta_i\}_{i=1}^n$ is an i.i.d. sequence with a Rademacher distribution.
\end{thm}
\begin{rem}
\label{rem:RSC-truncation}
Our proof of \cref{thm:rsc-condition}, stated in \cref{subsec:pf-rsc-condition}, follows a similar strategy as in \citep{klopp2014noisy}. However, in our general setting the random variable $\Inner{\bX,\bA}$, where $\bA\in\Cset(\nu,\eta)$, can be large by a factor of $\sqrt{d}$, e.g., in the multi-task learning case. In this situation, a direct application of the approach in \citep{klopp2014noisy} leads to an extra factor $\sqrt{d}$ in $\beta(\SamplingOp)$ that also appears in the final upper bound of Theorem \ref{thm:deterministic-error-bound}. In order to avoid this extra factor, we sharpen the application of the Massart concentration inequality, as shown in the proof of \cref{lem:bound-Bset-l}.
\end{rem}
Note that \cref{thm:rsc-condition} states RSC holds for $\Cset'(\theta,\eta)$ which is slightly different than the set $\Cset(\nu,\eta)$ defined in \cref{eq:Cset-def}. But, using \cref{as:gtmin-gtmax-bounds}, we can see that
\begin{align*}
\Cset(\nu,\eta)\subseteq\Cset'(\gmin\nu,\eta)\,.
\end{align*}
Therefore, the following variant of the RSC condition holds.
\begin{cor}
\label{cor:rsc-condition}
If \cref{as:gtmin-gtmax-bounds,as:x-tilde-boundedness-stochastic} hold, with probability greater than $1-2\exp\left(-{Cn\gmin\nu}/{\Cutoff^2}\right)$, the inequality
\begin{align}
\frac1n\Norm{\SamplingOp(\bA)}_2^2\geq\frac\gmin4\NormF{\bA}^2-\frac{93\eta\Cutoff^2}{\gmin}\Expect{\NormOp[\big]{\bSigma_R}}^2
\end{align}
holds for all $\bA\in\Cset(\nu,\eta)$, where $C>0$ is an absolute constant satisfying $Cn\gmin\nu>\Cutoff^2$, and $\bSigma_R := \frac{1}{n}\sum_{i=1}^{n}\zeta_i\bX_i$ with $\{\zeta_i\}_{i=1}^n$ is an i.i.d. sequence with Rademacher distribution.
\end{cor}

We conclude this section by stating the following corollary. This corollary combines the RSC condition (the version in \cref{cor:rsc-condition}) and the general deterministic error bound (\cref{thm:deterministic-error-bound}) to obtain the following probabilistic error bound.
\begin{cor}
\label{cor:prob-trace-reg-error-bound}
If \crefrange{as:x-tilde-boundedness-deterministic}{as:x-tilde-boundedness-stochastic} hold, and let $\lambda=\lambda(n)$ be larger than $C\, \Expect[\big]{\NormOp[\big]{\bSigma_R}}\ParamBound\Cutoff$ where $C$ is an arbitrary (and positive) constant, then we have that
\begin{align*}
\NormF[\big]{\Estimator-\Param}^2
&\leq
\frac{C'\lambda^2r}{\gmin^2}
\end{align*}
holds with probability at least
\begin{align}
1-\Prob{\lambda<3\NormOp{\bSigma}}-2\exp\left(-\frac{C''n\lambda^2r}{\Cutoff^2\ParamBound^2\gmin}\right)\label{eq:final-tail}
\end{align}
for numerical constants $C', C''>0$.
\end{cor}
\begin{proof}
First, we denote the threshold for $\lambda$ by $\lambda_1 =\lambda_1(n)= C\, \Expect{\NormOp[\big]{\bSigma_R}}\ParamBound\Cutoff$. Now, by defining
\begin{align*}
\alpha:=\frac{\gmin}{4},\qquad
\beta:=\frac{6696\cdot r\Cutoff^2}{\gmin}\Expect{\NormOp{\bSigma_R}}^2,
\qquad\text{and}\qquad
\nu:=\frac{\lambda_1^2r}{\gmin^2\ParamBound^2}\,,
\end{align*}
we observe that
\begin{align*}
\left(\frac{100\lambda^2r}{3\alpha^2}+\frac{8\ParamBound^2\beta}\alpha\right)\vee 4\ParamBound^2\nu&=
\left(\frac{1600\lambda^2r}{3\gmin^2}+\frac{32\times 6696\lambda_1^2 r}{C^2\gmin^2}\right)\vee \frac{4\lambda_1^2r}{\gmin^2}
\\
&\leq\frac{C'\lambda^2r}{\gmin^2},
\end{align*}
for a sufficiently large constant $C'>0$ (in fact, we would need $C'\ge 534+2.15\times 10^5\times C^{-2}$). The rest follows immediately from \cref{thm:deterministic-error-bound} and \cref{cor:rsc-condition}. We note that the condition of \cref{cor:rsc-condition} can be shown to hold by taking $C''$ such that $C''n\lambda^2r>\Cutoff^2\ParamBound^2\gmin$.
\end{proof}
\begin{rem}
\label{rem:lambda}
While \cref{cor:prob-trace-reg-error-bound} relies on two conditions for $\lambda$, namely, that $\lambda\ge \lambda_1$ and that $\Prob{\lambda<3\NormOp{\bSigma}}$ is small, only the latter condition is needed to obtain a tail bound like \cref{eq:trace-reg-bound} in \cref{thm:deterministic-error-bound}. The additional condition $\lambda\ge \lambda_1$ only helps to make the upper bound simpler, i.e., by allowing us to obtain ${C'\lambda^2r}/{\gmin^2}$ instead of the right-hand side of \cref{eq:trace-reg-bound}.
\end{rem}
\begin{rem}[Sampling complexity]
\label{rem:sampling-complexity}
The final upper bound obtained by \cref{cor:prob-trace-reg-error-bound} depends on $\lambda$, which is required to be larger than $C\, \Expect[\big]{\NormOp[\big]{\bSigma_R}}\ParamBound\Cutoff$.
We will show in \cref{sec:applications} that $\Expect[\big]{\NormOp[\big]{\bSigma_R}}$ is generally of order $\sqrt{d\log d/n}$, which yields an upper bound of order $rd\log/n$. We will also show that the probability of the upper bound, i.e., \cref{eq:final-tail}, is close to $1$ when $n\ge d\log^3 d$ (\crefrange{subsec:matrix-completion}{subsec:gauss-ensemble}) or
$n\ge d\log^4 d$ (\cref{subsec:factored-meas}), up to constants. Similarly, in the case of exact recovery, in \cref{sec:exact-recovery}, we will show that the sampling complexity is determined by a certain inequality that depends on $\Expect[\big]{\NormOp[\big]{\bSigma_R}}$, \cref{eq:exact-n-condition}, and is satisfied when (up to constants)
 $n\ge rd$ (for compressed sensing with Gaussian ensembles) and when $n\ge rd\log(d)$ (for compressed sensing with factored measurements).
\end{rem}
\begin{rem}[Lower bound and optimality]
\label{rem:lower-bound}
While \cref{cor:prob-trace-reg-error-bound} provides a general upper bound that applies to several families of problems, as we will showcase in \cref{sec:applications}, one may wonder whether the provided upper bound is tight. We will show in \cref{sec:applications} that the upper bound indeed matches provable lower bounds (known in the literature). Specifically, we will show that the upper bound of \cref{cor:prob-trace-reg-error-bound} is the same as that of Corollary 1 of \cite{negahban2012restricted}, for the matrix completion problem, and the same as the upper bound of Theorem 2.4 of \cite{candes2011tight}, for the compressed sensing problem. In both of these papers, the bounds are shown to be optimal, which means that the upper bound of \cref{cor:prob-trace-reg-error-bound} is also optimal for both classes of problems. But, it is an interesting open question whether one can prove a general lower bound, in the same spirit as the general upper bound of \cref{cor:prob-trace-reg-error-bound}.
\end{rem}

\section{Tail Bound for the Cross-validated Estimator}
\label{sec:cross-val}

One of the assumptions required for the tail bounds in \cref{sec:trace-reg} for $\Estimator=\Estimator(\lambda)$ is that $\lambda$ should be larger than $3\NormOp{\bSigma}$. However, in practice we do not have access to the latter parameter, which relies on knowledge of the noise values $\{\Noise_i\}_{i=1}^n$. Therefore, practitioners often use cross-validation to tune parameter $\lambda$. In this section, we prove that if $\lambda$ is selected via cross-validation, $\Estimator(\lambda)$ has similar tail bounds to those in \cref{sec:trace-reg}. This would provide theoretical support for the selection of $\lambda$ via cross-validation for our family of estimators.

Let $\{(\bX_i,y_i)\}_{i=1}^n$ be a set of observations and denote by $K$ the number of cross-validation folds. Let $\{\Iset_k\}_{k\in[K]}$ be a set of disjoint subsets of $[n]$ where $\cup_{k\in[K]}\Iset_k=[n]$. Also, we define $\Iset_{-k}:=[n]\setminus\Iset_{k}$.
Letting $n_k:=\Abs{\Iset_k}$, we have
\[
n=n_1+\cdots+n_K\,.
\]
Let $\SamplingOp_{k}(\cdot)$ and $\SamplingOp_{-k}(\cdot)$ be sampling operators for $\{\bX_i\}_{i\in \Iset_k}$ and $\{\bX_i\}_{i\in \Iset_{-k}}$, respectively. Similarly, $E_k$ and $E_{-k}$ are noise vectors on the corresponding sets of indices. Finally, $Y_k$ and $Y_{-k}$ denote the response vectors corresponding to $\SamplingOp_{k}(\Param)+E_k$ and $\SamplingOp_{-k}(\Param)+E_{-k}$, respectively. In our analysis, we assume that each partition contains a large fraction of all samples; namely, we assume that $n_k\geq N/(2K)$ for all $k\in[K]$.

Also, throughout this section, we assume that for any $\lambda>0$, the estimators $\Estimator_{-k}(\lambda)$ satisfy \cref{eq:goodness-condition} for observations $(\SamplingOp_{-k}(\Param),Y_{-k})$ for each $k\in[K]$. Note that this means that in
$\Loss(\Estimator_{-k}(\lambda))$, by \cref{eq:main-loss}, the proper scaling $1/(n-n_k)$ is used instead of $1/n$.

Define
\begin{align*}
\ErrorCV(\lambda):=\sum_{k=1}^{K}\Norm{Y_k-\SamplingOp_k(\Estimator_{-k}(\lambda))}_2^2\,.
\end{align*}
For any fixed $\lambda$, it can be observed that $\Norm[\big]{Y_k-\SamplingOp_k(\Estimator_{-k}(\lambda))}_2^2$ is an unbiased estimate of the prediction error for $\Estimator_{-k}(\lambda)$.
For every $\lambda$ we also define the estimator $\EstimatorCV(\lambda)$ as follows:
\begin{align}\label{eq:cvestimator}
\EstimatorCV(\lambda):=\sum_{k=1}^{K}\frac{n_k}n\cdot\Estimator_{-k}(\lambda)\,.
\end{align}
Cross-validation works by starting with a set $\Lambda=\{\lambda_1,\lambda_2,\cdots,\lambda_L\}$ of potential (positive) regularization parameters and then choosing
\[
\PenaltyCV\in\arg\min_{\lambda\in\Lambda}\widehat{E}(\lambda)\,.
\]
Then, the
\emph{$K$-fold cross-validated estimator} with respect to $\Lambda$ is $\EstimatorCV(\PenaltyCV)$.

Next, we state two main results. First, in \cref{thm:cv-generalization-bound}, we show a bound for $\EstimatorCV(\widehat{\lambda})$, where $\widehat{\lambda}$ can be any value in $\Lambda$. Then, in \cref{thm:cv-estimator-explicit-bound}, we combine \cref{thm:cv-generalization-bound} with \cref{cor:prob-trace-reg-error-bound} to obtain the main result of this section, which is an explicit tail bound for $\EstimatorCV(\PenaltyCV)$.
\begin{thm}
\label{thm:cv-generalization-bound}
Let $\Lambda=\{\lambda_1,\lambda_2,\cdots,\lambda_L\}$ be a set of positive regularization parameters, $\EstimatorCV$ be defined as in \cref{eq:cvestimator}, and $\PenaltyRand$ be a random variable such that $\PenaltyRand\in\Lambda$ almost surely. Moreover, define
\begin{align*}
\bar{\sigma}^2=\frac1{n}\sum_{i\in[n]}\Var{\Noise_i}\,,
\end{align*}
and assume that $(\Noise_i)_{i=1}^n$ are independent zero-mean $\sigma^2$-sub-Gaussian random variables. Then, for any $t>0$, we have
\begin{align}\label{eq:cv-generalization-bound}
\NormLTP[\big]{\EstimatorCV(\PenaltyRand)-\Param}^2
\leq
\widehat{E}(\PenaltyRand)-\bar{\sigma}^2+t\,,
\end{align}
with probability at least
\begin{align}\label{eq:cv-generalization-tail}
1-6KL\exp\left[-C\min\left(\frac{t^2}{\sigma^4\vee\ParamBound^4},\frac{t}{\sigma^2\vee\ParamBound^2}\right)\cdot\frac nK\right]\,,
\end{align}
where $C>0$ is a numerical constant. More generally, for all $\ell\in[L]$ and $k\in[K]$, in addition to \cref{eq:cv-generalization-bound}, the following inequality is satisfied with the same probability as in \cref{eq:cv-generalization-tail},
\begin{align*}
	\Abs[\Big]{\frac1{n_k}\Norm[\big]{Y_k-\SamplingOp_k(\Estimator_{-k}(\lambda_\ell))}_2^2-\left[\NormLTP{\Estimator_{-k}(\lambda_\ell)-\Param}^2+\bar{\sigma}_k^2\right]}\le t\,,
\end{align*}
where $\bar{\sigma}_k^2 =(1/{n_k})\sum_{i\in\Iset_k}\Var{\Noise_i}$.
\end{thm}
In order to prove \cref{thm:cv-generalization-bound}, we need to state and prove the following lemma.
\begin{lem}
\label{lem:gen-bound-single-b}
Let $\bB$ and $\Param$ be two $d_r\times d_c$ matrices with $\Norm{\bB-\Param}_{\psi_2(\SampleDistr)}\leq2\ParamBound$, and let $(\bX_i)_{i=1}^{n}$ be a sequence of i.i.d. samples drawn from $\SampleDistr$. Let $(\Noise_i)_{i=1}^{n}$ be a sequence of independent zero-mean $\sigma^2$-sub-Gaussian random variables and let $\bar{\sigma}^2$ be defined as follows
\begin{align*}
\bar{\sigma}^2:=\frac1n\sum_{i=1}^n\Var{\Noise_i}\,.
\end{align*}
Then, for any $t>0$, the inequality
\begin{align*}
\Abs[\Big]{\frac1{n}\Norm[\big]{Y-\SamplingOp(\bB)}_2^2-\left(\NormLTP{\bB-\Param}^2+\bar{\sigma}^2\right)}\leq t,
\end{align*}
holds with probability at least
\[
1-6\exp\left[-C\min\left(\frac{t^2}{\sigma^4\vee\ParamBound^4},\frac{t}{\sigma^2\vee\ParamBound^2}\right)n\right]\,,
\]
where $C>0$ is a numerical constant.
\end{lem}
\begin{proof} Recall from \cref{sec:notation} that the vector of all noise values $\{\Noise_i\}_{i=1}^n$ is denoted by $E$. Note that,
\begin{align*}
\Norm[\big]{Y-\SamplingOp(\bB)}_2^2
&=
\Norm[\big]{E-\SamplingOp(\bB-\Param)}_2^2\\
&=
\Norm[\big]{E}_2^2+\Norm[\big]{\SamplingOp(\bB-\Param)}_2^2-2\Inner[\big]{E,\SamplingOp(\bB-\Param)}\,.
\end{align*}
Next, using our \cref{lem:norm-se-exp-x-leq-norm-se-x} as well as Lemma 5.14 of \cite{vershynin2010introduction}, for each $i\in[n]$, we have
\begin{align*}
\NormSubE[\Big]{\Noise_i^2-\Expect{\Noise_i^2}}\le2\NormSubE{\Noise_i^2}\leq4\NormSubG{\Noise_i}^2\leq2\sigma^2\,.
\end{align*}
Also, $\Norm{\bB-\Param}_{\psi_2(\SampleDistr)}\leq2\ParamBound$ means that $\NormSubG{\Inner{\bX_i,\bB-\bB^\star}}\leq2\ParamBound$, which in turn yields that $\Inner{\bX_i,\bB-\bB^\star}^2$ is subexponential. We can now follow a similar logic as the above and obtain
\begin{align*}
\NormSubE[\big]{\Inner{\bX_i,\bB-\bB^\star}^2-\Expect{\Inner{\bX_i,\bB-\bB^\star}^2}}\leq16\ParamBound^2\,.
\end{align*}
In the same way, $\Noise_i\Inner{\bX_i,\bB-\Param}$, which is a product of two sub-Gaussian random variables, becomes subexponential with a zero mean:
\begin{align*}
\NormSubE{\Noise_i\Inner{\bX_i,\bB-\Param}}\leq8\sigma\ParamBound\,.
\end{align*}
It follows from Corollary 5.17 of \cite{vershynin2010introduction} that, by defining
\begin{align*}
\Eset_1(t)&:=\left\{\Abs[\Big]{\Norm[\big]{E}_2^2-\Expect[\big]{\Norm[\big]{E}_2^2}}\geq nt \right\}\,,\\
\Eset_2(t)&:=\left\{\Abs[\Big]{\Norm[\big]{\SamplingOp(\bB-\Param)}_2^2-\Expect[\big]{\Norm[\big]{\SamplingOp(\bB-\Param)}_2^2}}\geq nt \right\}\,,\\
\Eset_3(t)&:=\left\{\Abs[\Big]{2\Inner[\big]{E,\SamplingOp(\bB-\Param)}-\Expect[\big]{2\Inner[\big]{E,\SamplingOp(\bB-\Param)}}}\geq nt \right\}\,,
\end{align*}
we have
\begin{align*}
\Prob{\Eset_j}\leq2\exp\left[-C\min\left(\frac{t^2}{\sigma^4\vee\ParamBound^4},\frac{t}{\sigma^2\vee\ParamBound^2}\right)n\right]\,,
\end{align*}
for $j\in\{1, 2, 3\}$ where $C>0$ is a numerical constant. Applying the union bound, we get
\begin{align*}
\Prob[\Big]{\Abs[\Big]{\frac1{n}\Norm[\big]{Y-\SamplingOp(\bB)}_2^2-\left[\NormLTP{\bB-\Param}^2+\bar{\sigma}^2\right]}\geq t}
\leq
6\exp\left[C\min\left(\frac{t^2}{\sigma^4\vee\ParamBound^4},\frac{t}{\sigma^2\vee\ParamBound^2}\right)n\right]\,,
\end{align*}
for some (different) numerical constant $C>0$.
\end{proof}

Now we are ready to prove \cref{thm:cv-generalization-bound}.
\begin{proof}[Proof of \cref{thm:cv-generalization-bound}]
For all $\ell\in[L]$ and $k\in[K]$, define the bad event $\Eset_{\ell,k}$ to be
\begin{align*}
\Eset_{\ell,k}
:=
\left\{
\Abs[\Big]{\frac1{n_k}\Norm[\big]{Y_k-\SamplingOp_k(\Estimator_{-k}(\lambda_\ell))}_2^2-\left[\NormLTP{\Estimator_{-k}(\lambda_\ell)-\Param}^2+\bar{\sigma}_k^2\right]}> t
\right\}.
\end{align*}
It follows from \cref{lem:gen-bound-single-b}, the union bound, and the assumption $n_k\ge n/(2K)$, for all $k\in[K]$, that the bad event satisfies
\begin{align*}
\Prob{\bigcup_{k\in[K]}\bigcup_{\ell\in[L]}\Eset_{\ell,k}}
&\leq
6KL\exp\left(-c\min\left(\frac{t^2}{\sigma^4\vee\ParamBound^4},\frac{t}{\sigma^2\vee\ParamBound^2}\right)\cdot\frac nK\right)\,,
\end{align*}
for some constant $c\geq 0$.

Now, in the complement of the bad event, it follows from the convexity of $\NormLTP{\cdot}^2$ that
\begin{align*}
\NormLTP[\big]{\EstimatorCV(\PenaltyRand)-\Param}^2
&\leq
\sum_{k=1}^{K}\frac{n_k}n\cdot
\NormLTP[\big]{\Estimator_{-k}(\PenaltyRand)-\Param}^2\\
&\leq
\sum_{k=1}^{K}\frac{n_k}n\cdot\left[\frac1{n_k}\Norm[\big]{Y_k-\SamplingOp_k(\Estimator_{-k}(\widehat{\lambda})}_2^2-\bar{\sigma}_k^2+t\right]\\
&=
\widehat{E}(\PenaltyRand)-\bar{\sigma}^2+t\,,
\end{align*}
which is the desired result.
\end{proof}

Before stating the next result, we also define notations $\bSigma_{R,-k}$ and $\bSigma_{-k}$ as follows:
\[
\bSigma_{-k}=\sum_{i\in\Iset_{-k}}\Noise_i\bX_i
~~~~\text{and}~~~~\bSigma_{R,-k}=\sum_{i\in\Iset_{-k}}\zeta_i\bX_i\,,
\]
where, as in  \cref{sec:trace-reg}, $\{\zeta_i\}_{i\in[n]}$ are i.i.d. Rademacher random variables.

Now, we are ready to state the main result of this section, which is obtained by combining \cref{thm:cv-generalization-bound} with \cref{cor:prob-trace-reg-error-bound}.
\begin{thm}
\label{thm:cv-estimator-explicit-bound}
Let \crefrange{as:x-tilde-boundedness-deterministic}{as:x-tilde-boundedness-stochastic} hold, and let $\ell_0\in[L]$ be such that $\lambdaellzero$ (in $\Lambda$) is larger than $C\ParamBound\Cutoff \max_{k\in[K]}\Expect{\NormOp{\bSigma_{R,-k}}}$, where $C$ is an arbitrary (and positive) constant, and $\Cutoff{}$ is defined as in \cref{sec:trace-reg}. Also assume that $\Lambda$, $\PenaltyCV$, and $\EstimatorCV$ are defined as above. In addition, assume that  $\{\Noise_i\}_{i=1}^n$ are independent zero-mean $\sigma^2$-sub-Gaussian random variables. Then, for all $t>0$, we have
\begin{align*}
\NormLTP[\big]{\EstimatorCV(\PenaltyCV)-\Param}^2
&\leq
\frac{C_1\gmax\lambdaellzero^2r}{\gmin^2}+2t
\end{align*}
with probability at least
\begin{align}\label{eq:cv-good-probability}
\begin{split}
&1
-6KL\exp\left[-C_2\min\left(\frac{t^2}{\sigma^4\vee\ParamBound^4},\frac{t}{\sigma^2\vee\ParamBound^2}\right)\cdot\frac nK\right]\\
&~~~
-\sum_{k\in[K]}\Prob{\lambdaellzero\geq3\NormOp{\bSigma_{-k}}}\\
&~~~
-K\exp\left(-\frac{C_3n\lambdaellzero^2r}{\Cutoff^2\ParamBound^2\gmin}\right)\,,
\end{split}
\end{align}
where $C_1$, $C_2$, and $C_3$ are positive constants.
\end{thm}

\begin{proof}
The definition of $\PenaltyCV$, together with \cref{thm:cv-generalization-bound}, \cref{cor:prob-trace-reg-error-bound}, \cref{as:gtmin-gtmax-bounds}, and the union bound, yields
\begin{align*}
\NormLTP[\big]{\EstimatorCV(\PenaltyCV)-\Param}^2
&\leq
\widehat{E}(\PenaltyCV)-\bar{\sigma}^2+t\nonumber\\
&\leq
\widehat{E}(\lambdaellzero)-\bar{\sigma}^2+t\nonumber\\
&=
\sum_{k=1}^{K}\frac{n_k}n\cdot\left[\frac1{n_k}\Norm[\big]{Y_k-\SamplingOp_k(\Estimator_{-k}(\lambda_{\ell_0})}_2^2-\bar{\sigma}_k^2+t\right]\nonumber\\
&\leq
\sum_{k=1}^{K}\frac{n_k}n\cdot\left[\NormLTP{\Estimator_{-k}(\lambdaellzero)-\Param}^2+2t\right]
\\
&\leq
\sum_{k=1}^{K}\frac{n_k}n\cdot\left[\frac{ C_1\gmax \lambdaellzero^2r}{\gmin^2}+2t\right]\\
&=
\frac{C_1\gmax \lambdaellzero^2r}{\gmin^2}+2t
\end{align*}
with the probability stated in \cref{eq:cv-good-probability}. Note that we also used the fact that $|\Iset_{-k}|\geq n[1-1/(2K)]\geq n/2$ in the last term of \cref{eq:cv-good-probability}.
\end{proof}
\begin{rem}
\label{rem:cv-tail-L2SampleDistr}
While the tail bound of \cref{thm:cv-estimator-explicit-bound} is stated for $\NormLTP{\EstimatorCV(\PenaltyCV)-\Param}^2$, it is straightforward to use \cref{as:gtmin-gtmax-bounds} to obtain the following bound on $\NormF{\EstimatorCV(\PenaltyCV)-\Param}^2$ with the same probability
\begin{align*}
    \NormF{\EstimatorCV(\PenaltyCV)-\Param}^2
    \leq
    \frac{C_1\gmax\lambdaellzero^2r}{\gmin^3}+\frac{2t}\gmin.
\end{align*}
\end{rem}
\begin{rem}\label{rem:lower-bound-cv}
In light of \cref{rem:lower-bound}, and the fact that the above upper bound is the same as that of \cref{cor:prob-trace-reg-error-bound} up to constants, our bounds for the cross-validated estimator are also tight for the matrix completion and the compressed sensing settings.
\end{rem}
\begin{rem}[Variants of cross-validation]\label{rem:other-cv-forms}
Recall that \cref{thm:cv-generalization-bound} shows that
the test error
$(1/{n_k})\Norm{Y_k-\SamplingOp_k(\Estimator_{-k}(\lambda_\ell))}_2^2$
concentrates around its expectation
$\NormLTP{\Estimator_{-k}(\lambda_\ell)-\Param}^2+\bar{\sigma}_k^2$,
for every fold $k$ and every regularization parameter $\lambda_\ell$. Therefore, one can obtain a similar bound as in \cref{thm:cv-estimator-explicit-bound} when the cross-validation approach is modified. One alternative is to choose the optimum penalty using a single training and test split. Another option is to create $K$ random splits of the data into training and test sets and, similar to \cref{eq:cvestimator}, define $\EstimatorCV(\PenaltyRand)$ as a weighted average of the estimator for each fold. The proof of \cref{thm:cv-generalization-bound} stays valid since it requires only the convexity of $\NormLTP{\cdot}^2$ and the concentration of the test error for every split. However, the approach that uses a single split is expected to have a higher variance. Finally, often in practice after $\PenaltyCV$ is selected via some form of cross-validation, the estimator is refitted on the entire training set with the selected penalty parameter.  Our current theoretical analysis does not provide a guarantee for this version of cross-validation since the independence of $\Noise_i$ and $\lambda$ in our analysis would break down. However, we do compare our approach with this refitting version in the simulations of \cref{sec:supp-simulations}.
\end{rem}
\begin{rem}[Selecting the $\lambda$-sequence]\label{rem:finding-good-lambda}
It is important to note that the upper bound of \cref{thm:cv-estimator-explicit-bound} depends on $\lambda_{\ell_0}$. The bound is practically relevant if $\lambda_{\ell_0}$ is close to the theoretically optimal one. However, one can ensure that, up to constants, such an $\lambda_{\ell_0}$ exists in the sequence $\Lambda$ by selecting $\Lambda$ as follows: let $\lambda_{\max{}}$ be the minimum real number for which the only minimizer of the convex program is zero. It can be easily shown that $\lambda_{\max{}}=\NormOp{\sum_{i=1}^{n}Y_i\bX_i}$.
Then, we select the sequence of values of $(\lambda)_{i=1}^L$ as follows: $\lambda_1:=\lambda_{\max{}}$,
$\lambda_{\ell+1} = \lambda_\ell/2$, and $L$ is taken large enough such that $\lambda_L=2^{-L+1}\lambda_{\max{}}$ is close to $0$, thereby guaranteeing that the optimal penalty, up to a factor of $2$, is included in the sequence. Note that such a large $L$ impacts the upper bound only double logarithmically.  Specifically, a larger $L$ requires selecting a larger $t$ to make the tail probability
$
6KL\exp[-C_2\min({t^2}/{(\sigma^4\vee\ParamBound^4)},{t}/{(\sigma^2\vee\ParamBound^2)})\cdot (n/K)]
$
small, which means, up to constants, that the term $2t$ in the upper bound
of \cref{thm:cv-estimator-explicit-bound} is
\[
\sigma^2\vee\ParamBound^2\max\left(\frac{\log [K\log \frac{\lambda_{\max{}}}{\lambda_L}]}{n/K},\sqrt{\frac{\log [K\log \frac{\lambda_{\max{}}}{\lambda_L}]}{n/K}}\,\right)\,.
\]
This is the reason, in practice, why even $L=100$ leads to  a good cross-validated estimator. 
\end{rem}
In the next section, we numerically evaluate the effectiveness of this cross-validated estimator. 
\subsection{Simulations}
\label{sec:simulations}

In this section, we test the empirical performance of the cross-validated estimator, using synthetic and real data.

For the synthetic data setting, we generate a $d\times d$ matrix $\Param$ of rank $r$. Following a similar approach as in \citep{keshavan2010matrix}, we first generate $d\times r$ matrices $\bB_L^\star$ and $\bB_R^\star$ with independent standard normal entries and then set $\Param:=\bB_L^\star\cdot\bB_R^{\star\top}$.
For the real data setting, we use movie ratings from the MovieLens\footnote{The file {\tt ratings.csv} from \href{https://www.kaggle.com/datasets/grouplens/movielens-20m-dataset}{https://www.kaggle.com/datasets/grouplens/movielens-20m-dataset}.} data set, from which we take the $d$ most frequently watched movies and the $d$ users that rated the most movies. For $d$ equal to $50$ or $100$, this leads to a matrix with over 93\% observed ratings and we impute the remaining entries with the average of the observed ratings. We also center the matrix and scale it so that the overall mean and variance of the entries match the synthetic data,  and refer to the resulting rating as matrix $\Param$.
For the distribution of observations, $\SampleDistr$, we consider the matrix completion case. Specifically, for each $i\in[n]$, $r_i$ and $c_i$ are integers in $[d]$, selected independently and uniformly at random. Then,
$\bX_i=e_{r_i}e_{c_i}^\top$. This leads to $n$ observations
$Y_i=\Param_{r_ic_i}+\Noise_i$, where $\Noise_i$ are taken to be i.i.d. standard normal random variables.

\begin{figure}[t]
	\centering
	\begin{subfigure}[b]{0.49\textwidth}
		\centering
		\includegraphics[width=\textwidth]{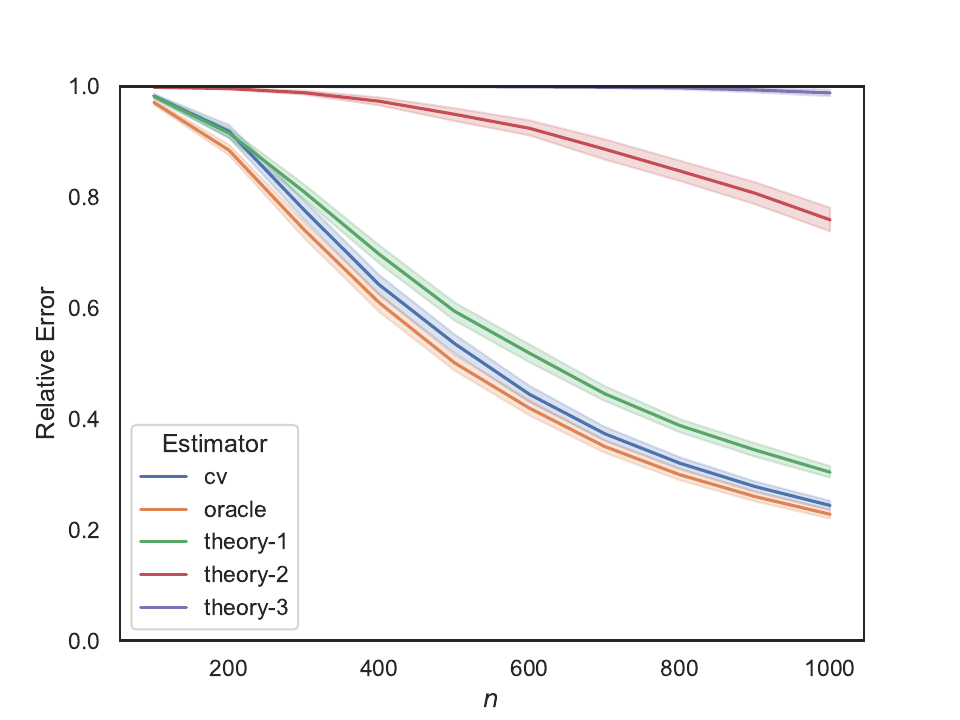}
		\caption{Synthetic data, $(d,r)=(50,2)$}
	\end{subfigure}
	\hfill
	\begin{subfigure}[b]{0.49\textwidth}
		\centering
	\includegraphics[width=\textwidth]{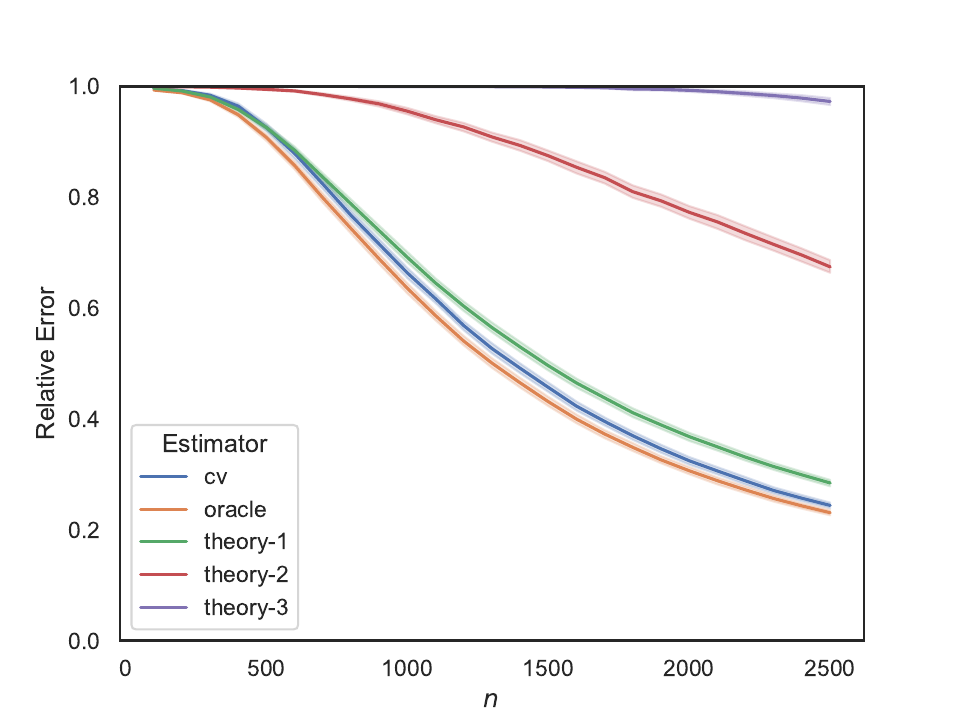}
		\caption{Synthetic data, $(d,r)=(100,3)$}
	\end{subfigure}
	\begin{subfigure}[b]{0.49\textwidth}
		\centering
		\includegraphics[width=\textwidth]{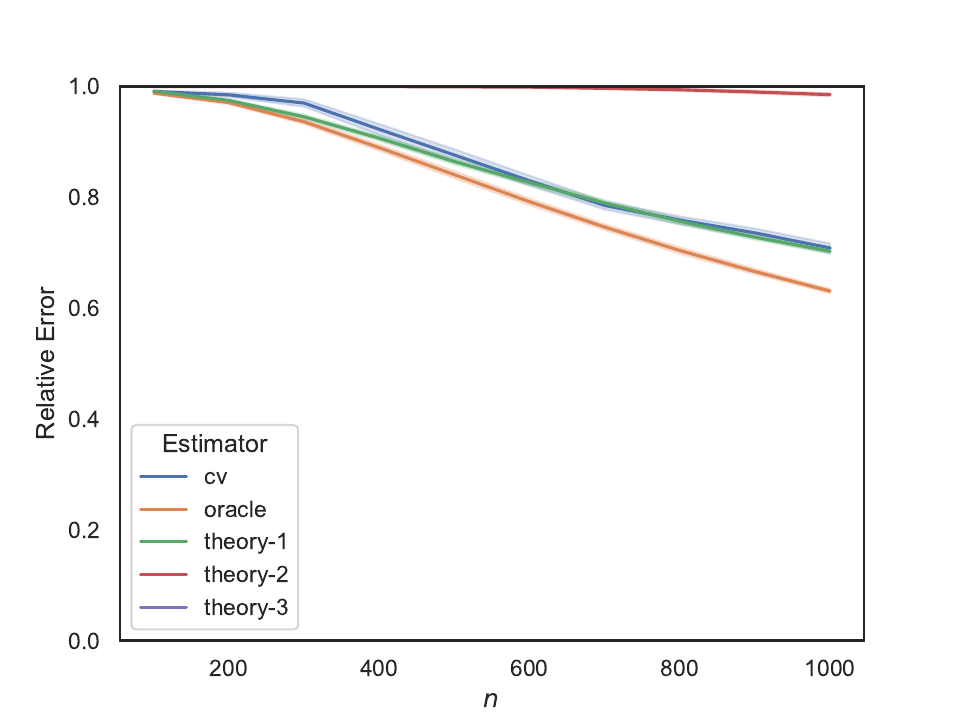}
		\caption{MovieLens data, $d=50$}
	\end{subfigure}
\hfill
	\begin{subfigure}[b]{0.49\textwidth}
	\centering
	\includegraphics[width=\textwidth]{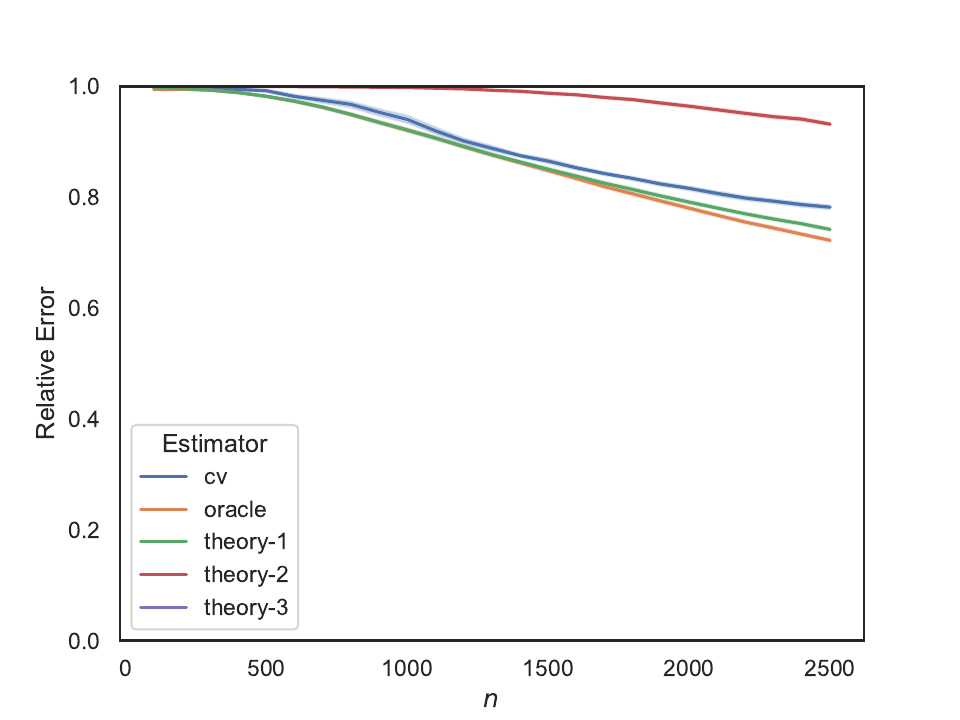}
	\caption{MovieLens data, $d=100$}
\end{subfigure}
	\caption{Comparison of the relative error (i.e., $\NormF[\big]{\Estimator-\Param}^2/\NormF{\Param}^2$) of the proposed estimators, for the synthetic and MovieLens data.}
	\label{fig:rel-error}
\end{figure}

\begin{figure}[t]
	\centering
	\begin{subfigure}[b]{0.49\textwidth}
		\centering
		\includegraphics[width=\textwidth]{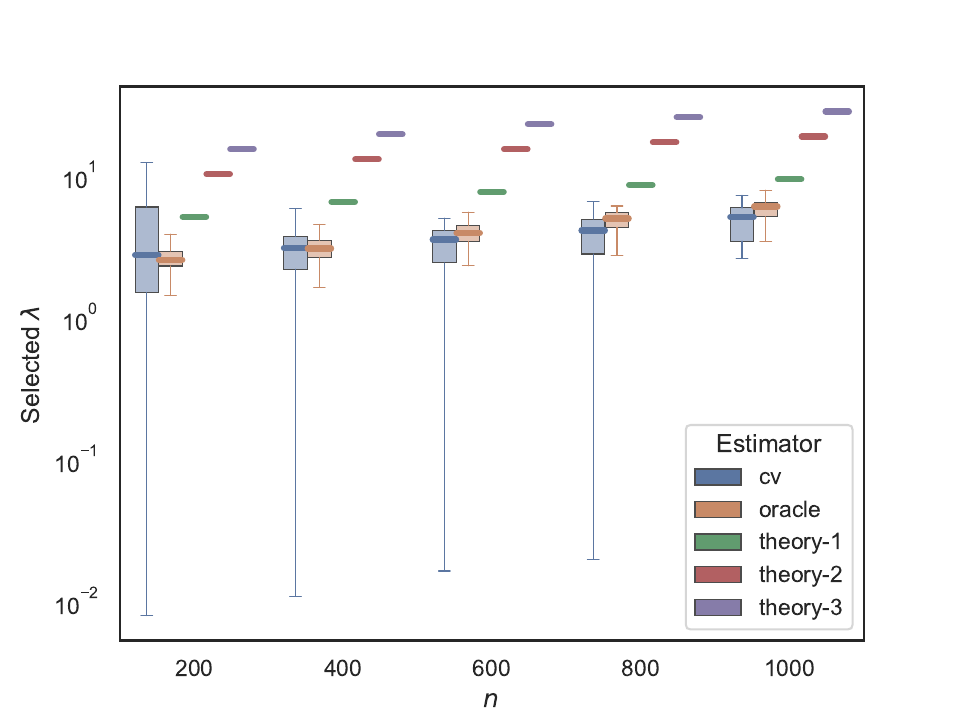}
		\caption{Synthetic data, $(d,r)=(50,2)$}
	\end{subfigure}
	\hfill
	\begin{subfigure}[b]{0.49\textwidth}
		\centering
		\includegraphics[width=\textwidth]{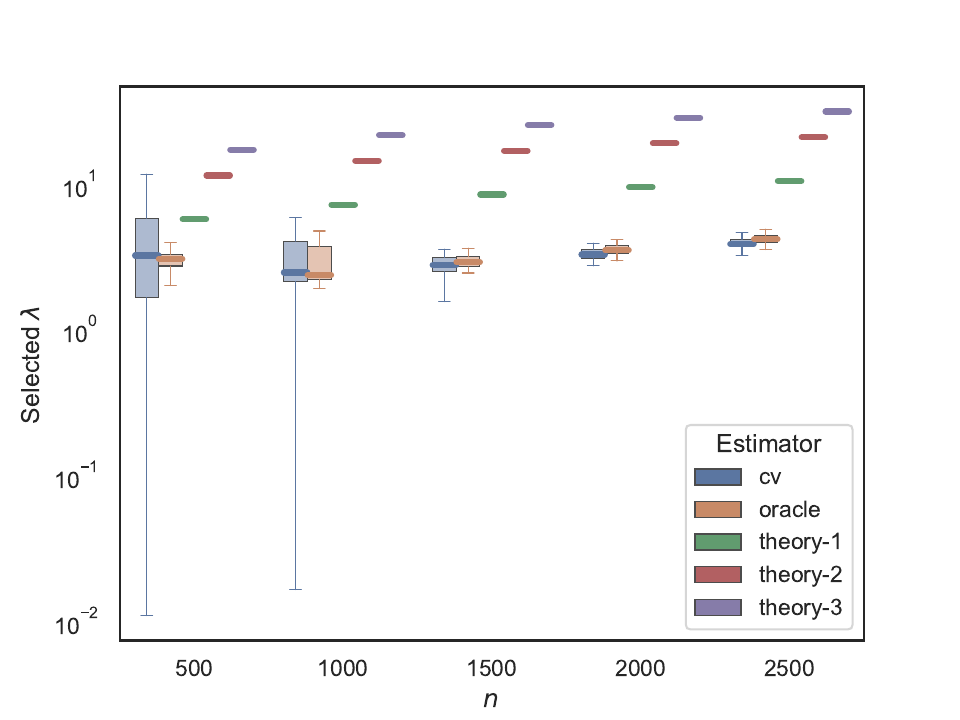}
		\caption{Synthetic data, $(d,r)=(100,3)$}
	\end{subfigure}\\
	\begin{subfigure}[b]{0.49\textwidth}
		\centering
		\includegraphics[width=\textwidth]{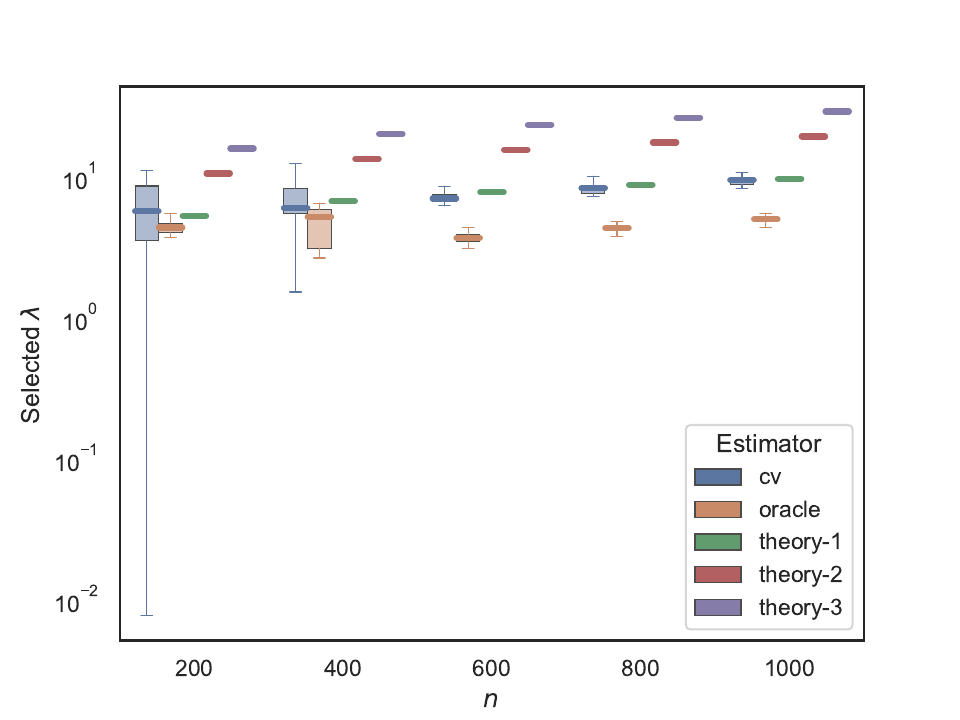}
		\caption{MovieLens data, $d=50$}
	\end{subfigure}
	\hfill
	\begin{subfigure}[b]{0.49\textwidth}
		\centering
		\includegraphics[width=\textwidth]{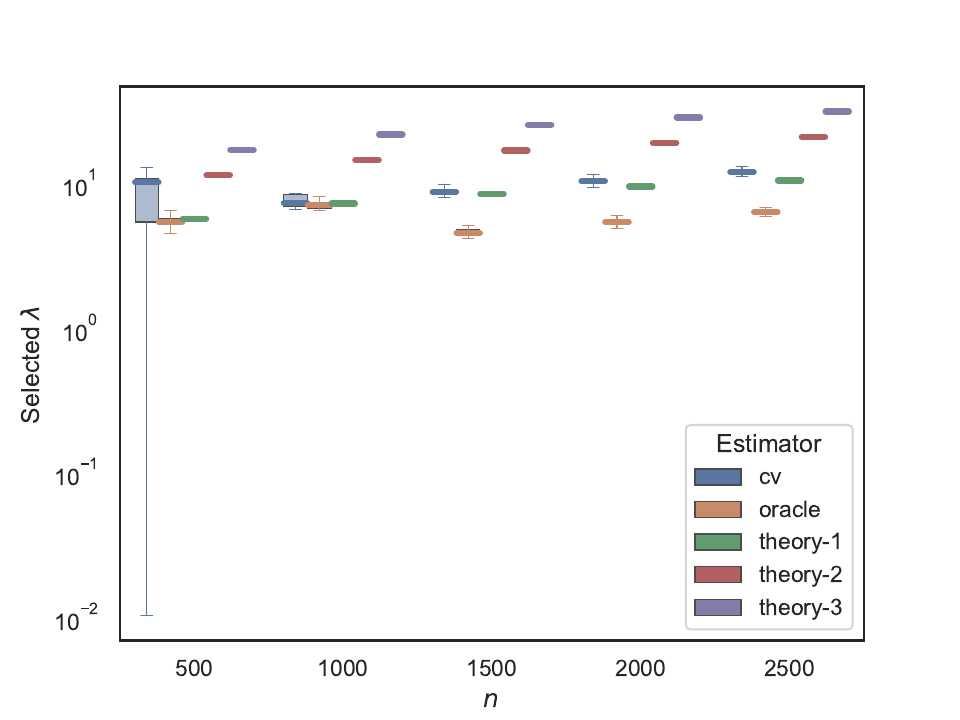}
		\caption{MovieLens data, $d=100$}
	\end{subfigure}
	\caption{Comparison of the selected penalty parameter $\lambda$ of the proposed estimators, for the synthetic and MovieLens data.}
	\label{fig:lam-sel}
\end{figure}

Given these observations, we compare the estimation error of the following five different estimators:
\begin{enumerate}
\item  \verb|theory-1|, \verb|theory-2|, and \verb|theory-3| estimators solve the convex program \cref{eq:convex-opt-version-1} for a given value of $\lambda=\lambda_0$ that is motivated by the theoretical results. Specifically, by \cref{rem:lambda}, we need $\lambda_0\geq3\NormOp{\bSigma}$ to hold with high probability, which means that we select $\lambda_0$ so that $\lambda_0\geq3\NormOp{\bSigma}$ holds with probability $0.9$. For each sample of size $n$, we find $\lambda_0$ by generating 1000 independent data sets of the same size and then, for the \verb|theory-3| estimator, we choose the $100^{\operatorname{th}}$ biggest value of $3\NormOp{\bSigma}$. While the main theory-inspired estimator is \verb|theory-3|, as we will see below, it performs very poorly since the constant $3$ as multiplier of $\NormOp{\bSigma}$ may be too conservative. Therefore, we also consider two other variants of this estimator where constant $3$ is replaced with constants $1$ and $2$ and denote these estimators by \verb|theory-1| and \verb|theory-2|, respectively. Overall, we highlight that these three estimators cannot be used in practice since they need access to $\NormOp{\bSigma}$.

\item The \verb|oracle| estimator solves the convex program \cref{eq:convex-opt-version-1} over a set of regularization parameters $\Lambda$. Then, the estimate is obtained by picking the matrix $\Estimator$ that has the minimum distance to the ground truth matrix $\Param$ under the Frobenius norm.  The set $\Lambda$ that is used in this estimator is obtained as follows: by \cref{rem:finding-good-lambda}, $\lambda_{\max{}}$ is the minimum real number for which the only minimizer of the convex program is zero, i.e., $\lambda_{\max{}}=\NormOp{\sum_{i=1}^{n}Y_i\bX_i}$.
    Then, we set $\lambda_{\min{}}:=\min(\lambda_0/10,10^{-5})$, and then the sequence of values of $(\lambda)_{i=1}^L$ is generated as follows: $\lambda_1:=\lambda_{\max{}}$ and $\lambda_{\ell+1} = \lambda_\ell/2$ such that $L$ is the smallest integer with $\lambda_L\leq\lambda_{\min{}}$.

\item The \verb|cv| estimator is introduced at the beginning of this section. We use $10$ folds (i.e., $K=10$) and use a set of regularization parameters $\Lambda'=\{\lambda\}_{i=1\in[L']}$ constructed exactly like the ones in the \verb|oracle| estimator; however, since \verb|cv| does not have access to $\lambda_0$, $L'$ is the smallest integer with $\lambda_{L'}\leq0.001\lambda_{\max{}}$.
\end{enumerate}
Finally, for each of these estimators, we compute the \emph{relative error} of the estimate $\Estimator$ from the ground truth $\Param$ defined as ${\NormF[\big]{\Estimator-\Param}^2}/{\NormF{\Param}^2}$ for a range of $n$. The results, averaged over 100 runs with 2SE error bars, are shown in \cref{fig:rel-error}, for two instances $(d,r)=(50,2)$ and $(d,r)=(100,3)$. We can see that \verb|cv| performs close to the \verb|oracle| and outperforms the theoretical estimators.
We also present the distribution of the selected $\lambda$ by each method, over all 100 runs, in Figure \ref{fig:lam-sel}. There is no variation for the selected $\Penalty$ in the theory-inspired estimators as we selected their $\lambda_0$ at the beginning of the experiment, using 1,000 data sets. It is also noticeable that the penalty selected by \verb|cv| is close to the one selected by \verb|oracle|, while the ones suggested by the theory are larger.
\paragraph{Sensitivity to a different noise level or cross-validation approach.} Recall that in the above simulations the noise standard deviation is $1$. We repeat the above simulations for two more values of the noise standard deviation, $0.5$ and $2$. In addition, and motivated by \cref{rem:other-cv-forms}, we also test three more variations of cross-validation. The results, presented in \cref{sec:supp-simulations}, validate the above observations.\footnote{All codes are available in this repository \href{https://github.com/mohsenbayati/cv-impute}{https://github.com/mohsenbayati/cv-impute}.}  

\section{Applications to Noisy Recovery}
\label{sec:applications}

In this section, we will show the benefit of proving \cref{cor:prob-trace-reg-error-bound}, which provides an upper bound with a more general norm $\SpikeNorm(\cdot)$. We will look at four different special cases for the distribution $\SampleDistr$ and in
two cases (matrix completion and compressed sensing with Gaussian ensembles) we will recover existing results by \cite{negahban2012restricted}, \cite{klopp2014noisy}, and \cite{candes2009exact}. For the other two (multi-task learning and compressed sensing with factored measurements), stated as open problems in \citep{rohde2011estimation} and \citep{recht2010guaranteed}, we will obtain the first (to the best of our knowledge) such results. Overall, in order to apply \cref{cor:prob-trace-reg-error-bound} in each case, we  need to perform the following:
\begin{enumerate}
  \item Choose a norm $\SpikeNorm(\cdot)$. In the examples below, we will be using $\Norm{\bB}_{\psi_p(\SampleDistr)}$ for parameters $p$ that are determined in each case.
  
  \item Compute $\NormLTP{\bB}$ to find an appropriate constant $\gmin$ for which \cref{as:gtmin-gtmax-bounds} holds.
  
  \item Compute $\SpikeNorm(\Param)$ to obtain the constant $\ParamBound$.
  
  \item Choose an appropriate constant $\Cutoff{}$ such that  \cref{as:x-tilde-boundedness-stochastic} holds.
  
  \item Apply \cref{lem:freedman} of \cref{subsec:bernstein} to obtain a bound for $\Prob{\lambda<3\NormOp{\bSigma}}$ as well as calculate $\Expect{\NormOp{\bSigma_R}}$.
\end{enumerate}
To simplify the notation, we assume that $d_r=d_c=d$ throughout this section; however, it is easy to see that the arguments hold for $d_r\neq d_c$ as well.
We also assume, for simplicity, that $\Noise_i\sim\Normal{0,\sigma^2}$ for all $i\in[n]$.

\subsection{Matrix completion}\label{subsec:matrix-completion}

Let $\Param$ be a $d\times d$ matrix and recall that $e_1,e_2,\ldots,e_d$ denotes the standard basis for $\IR^d$. Let, also, for each $i\in[n]$, $r_i$ and $c_i$ be integers in $[d]$, selected independently and uniformly at random. Then,
let $\bX_i=\xi_i\cdot e_{r_i}e_{c_i}^\top$ where, for each $i$, $\xi_i$ is an independent
$4d^2$-sub-Gaussian random variable that is also independent of $r_j$ and $c_j$, $j\in[n]$. If we set $\xi_i:=d$ almost surely, then
$\NormSubG{\xi_i}=d/\sqrt{\log2}\leq2d$. This corresponds to the problem studied in \citep{negahban2012restricted}. Here we show the bounds for the slightly more general case of $\xi_i\sim\Normal{0,d^2}$.

First, note that
\begin{align*}
\NormLTP{\bB}&=\NormF{\bB}\,,
\end{align*}
which means that $\gmin$ is equal to $1$. In order to find a suitable norm $\SpikeNorm(\cdot)$, we next study $\NormSGB{\bB}$ to see how heavy-tailed $\Inner{\bB,\bX_i}$ is. We have
\begin{align*}
\Expect*{\exp\left(\frac{\Abs{\Inner{\bB,\bX_i}}^2}{4d^2\NormInf{\bB}^2}\right)}
&=
\frac1{d^2}\sum_{j,k=1}^{d}\Expect*{\exp\left(\frac{\xi_i^2\bB_{jk}^2}{4d^2\NormInf{\bB}^2}\right)}\\
&=
\frac1{d^2}\sum_{j,k=1}^{d}\left[\frac{1}{\sqrt{\left(1-\frac{\bB_{jk}^2}{2\NormInf{\bB}^2}\right)_+}}\right]\\
&\leq 2\,,
\end{align*}
where the second equality follows from \cref{lem:expect-gaussian2} of \cref{sec:aux}. Therefore,
\begin{align*}
\NormSGB{\bB}\leq 2d\NormInf{\bB}\,,
\end{align*}
which guides our selection of $\SpikeNorm(\cdot)=d\NormInf{\cdot}$ and $\ParamBound = 2d\NormInf{\Param}$. We can now see that $\Cutoff=9$ fulfils \cref{as:x-tilde-boundedness-stochastic}. The reason for this is that, given $\SpikeNorm(\bB)=d\NormInf{\bB}=1$, we can condition on $r_i$ and $c_i$ and use \cref{cor:orlicz-trun-second-moment-bound,cor:normal-c-nu} of \cref{sec:aux} to obtain
\begin{align*}
\Expect*{\xi_i^2\bB_{r_ic_i}^2\cdot\II\left(\Abs{\xi_i\bB_{r_ic_i}}\leq 9\right)\Given r_i,c_i}
&\geq
\Expect*{\xi_i^2\bB_{r_ic_i}^2\cdot\II\left(\Abs{\xi_i\bB_{r_ic_i}}\leq 5\sqrt{\frac{8}{3}}d\Abs{\bB_{r_ic_i}}\right)\Given r_i,c_i}\\
&\geq\frac{\Expect*{\xi_i^2\bB_{r_ic_i}^2\Given r_i,c_i}}{2}\,.
\end{align*}
Therefore, we can take the expectation with respect to $r_i$ and $c_i$ and use the tower property to show that \cref{as:x-tilde-boundedness-stochastic} holds for $\Cutoff{}=9$.

The next step is to use \cref{lem:freedman} of \cref{subsec:bernstein}, with specification $\bZ_i:=\Noise_i\bX_i$, to find a tail bound inequality for $\Prob{\lambda<3\NormOp{\bSigma}}$. Define $\delta:=d\sigma e/(e-1)$ and let $G_1$ and $G_2$ be two independent standard normal random variables. It follows that
\begin{align*}
\Expect{\exp\left(\frac{\NormOp{\bZ_i}}{\delta}\right)}
&=
\Expect{\exp\left(\frac{d\sigma |G_1G_2|}{\delta}\right)}\\
&\leq
\Expect{\exp\left(\frac{d\sigma \left(G_1^2+G_2^2\right)}{2\delta}\right)}\\
&=
\Expect{\exp\left(\frac{d\sigma G_1^2}{2\delta}\right)}^2\\
&=
\Expect{\exp\left(\frac{(e-1) G_1^2}{2e}\right)}^2\\
&=
\left[\frac{1}{\sqrt{1-\frac{e-1}{e}}}\right]^2=
e\,.
\end{align*}
Next, notice that
\begin{align*}
\Expect{\bZ_i\bZ_i^\top}=\Expect{\bZ_i^\top\bZ_i}=d\sigma^2\,\Eye_d.
\end{align*}
Therefore, applying \cref{lem:freedman} of \cref{subsec:bernstein} gives
\begin{align}
\label{eq:mc-sigma-bound}
\Prob{\lambda<3\NormOp{\bSigma}}
\leq
2d\exp\left[-\frac{Cn\lambda}{d\sigma}\left(\frac{\lambda}{\sigma}\wedge\frac1{\log d}\right) \right]\,,
\end{align}
for some constant $C$.

We can follow the same argument for $\zeta_i\bX_i$, and use \cref{cor:freedman-4-expectation} of \cref{subsec:bernstein} to obtain
\begin{align}
\label{eq:mc-sigma-r-bound}
\Expect{\NormOp{\bSigma_R}}\leq C_1\sqrt{\frac{d\log d}{n}}\,,
\end{align}
provided that $n\geq C_2d\log^3d$ for constants $C_1$ and $C_2$. We can now combine \cref{eq:mc-sigma-bound}, \cref{eq:mc-sigma-r-bound}, and
\cref{cor:prob-trace-reg-error-bound} to obtain the following result: for any $\lambda\geq C_3\ParamBound\sqrt{d\log d/n}$ and $n\geq C_2d\log^3d$, the inequality
\begin{align*}
\NormF[\big]{\Estimator-\Param}^2
&\leq
C_4\lambda^2r
\end{align*}
holds with probability at least
\[
1-2d\exp\left[-\frac{Cn\lambda}{\sigma d}\left(\frac{\lambda}{\sigma}\wedge\frac1{\log d}\right) \right]-\exp\left(-\frac{C_5n\lambda^2r}{\ParamBound^2}\right)\,.
\]
In particular, setting
\begin{align}\label{eq:lambda-def-mc}
\lambda=C_6(\sigma\vee\ParamBound)\sqrt{\frac{\rho\,d}n},
\end{align}
for some $\rho\geq\log d$, we have that
\begin{align}
\label{eq:mc-error-bound}
\NormF[\big]{\Estimator-\Param}^2
&\leq
C_7(\sigma^2\vee\ParamBound^2)\:\frac{\rho\,dr}{n}
\end{align}
with probability at least $1-\exp(-C_8\rho)$ whenever $n\geq C_2d\log^3d$. This result closerly recovers Corollary 1 in \cite{negahban2012restricted} whenever $\rho=\log d$.

\subsection{Multi-task learning}\label{subsec:bandit-regression}

As in \cref{subsec:matrix-completion}, let $\Param$ be a $d\times d$ matrix and, for each $i\in[n]$, let $r_i$ be an integer in $[n]$, selected independently and uniformly at random. Then let $\bX_i=e_{r_i}\cdot X_i^\top$ where, for each $i$, $X_i$ is an independent $\Normal{0,d\cdot\Eye_d}$ random vector that is also independent of $\{r_j\}_{j=1}^n$. It follows that
\begin{align*}
\NormLTP{\bB}=\NormF{\bB}\,,
\end{align*}
which means that $\gmin=1$. Also,
\begin{align*}
\NormSGB{\bB}\leq2\sqrt{d}\Norm{\bB}_{2,\infty}\,.
\end{align*}
To see the latter, for any $\bX\sim\SampleDistr$ we follow the same steps as in the previous section and obtain,
\begin{align}
\Expect{\exp\left(\frac{\Abs{\Inner{\bB,\bX}}^2}{4d\Norm{\bB}_{2,\infty}^2}\right)}
&=
\frac1{d}\sum_{j=1}^{d}\Expect*{X_i}{\exp\left(\frac{\Abs{\Inner{\bB,e_{j}\cdot X_i^\top}}^2}{4d\Norm{\bB}_{2,\infty}^2}\right)}\nonumber\\
&=\frac1{d}\sum_{j=1}^{d} \left[\frac{1}{\sqrt{\left(1-\frac{\Norm{B_j}_{2}^2}{2\Norm{\bB}_{2,\infty}^2}\right)_+}}\right]\label{eq:invoke-gaussian-exp}\\
&\leq 2\,,\nonumber
\end{align}
where $B_j$ denotes the $j^{th}$ row of $\bB$ and \cref{eq:invoke-gaussian-exp} uses \cref{lem:expect-gaussian2} of \cref{sec:aux}, since  $\Inner{B_j,X_i}\sim\Normal{0,d\|B_j\|_2^2}$. The final step uses $\Norm{B_j}_{2}\leq \Norm{\bB}_{2,\infty}$, which follows from the definition of $\Norm{\bB}_{2,\infty}$. As in \cref{subsec:matrix-completion}, we use this to set $\SpikeNorm(\cdot)=2\sqrt{d}\Norm{\Param}_{2,\infty}$, which means that we can select $\ParamBound$ to be equal to $2\sqrt{d}\Norm{\Param}_{2,\infty}$. Also, as in \cref{subsec:matrix-completion}, we can condition on the random variable $r_i$ to show that
\[\Expect{(B_{r_i}X_i)^2\II\left(\Abs{B_{r_i}X_i}\leq 9\right)\Big| r_i}\geq\frac{1}{2}\Expect{(B_{r_i}X_i)^2\Big| r_i}\,,
\]
which means that $\Cutoff{}=9$ satisfies the requirement of \cref{as:x-tilde-boundedness-stochastic}.

Let $\bZ_i$ be as in \cref{subsec:matrix-completion}, $\delta=d\sigma e/(e-1)$, and  $(G_i)_{i=0}^{d}$ be a sequence of $d+1$ independent standard normal random variables. We see that
\begin{align*}
\Expect{\exp\left(\frac{\NormOp{\bZ_i}}{\delta}\right)}
&=
\Expect{\exp\left(\frac{\NormF{\bZ_i}}{\delta}\right)}\\
&=
\Expect{\exp\left(\frac{\sigma d\Abs{G_0}\sqrt{\frac{G_1^2+\cdots+G_d^2}{d}}}{\delta}\right)}\\
&\leq
\Expect{\exp\left(\frac{\sigma d\left(G_0^2+\frac{G_1^2+\cdots+G_d^2}{d}\right)}{2\delta}\right)}\\
&=
\left(1-\frac{\sigma d}{\delta}\right)^{-\frac12}\cdot\left(1-\frac{\sigma}{\delta}\right)^{-\frac d2}\\
&=
\left(1-\frac{e-1}{e}\right)^{-\frac12}\cdot\left(1-\frac{e-1}{ed}\right)^{-\frac d2}\\
&\leq
\sqrt{e}\cdot e^{\frac{e-1}{2e}} & \text{(using $1+x\leq e^x$)}\\
&\leq
e.
\end{align*}
Furthermore, we have
\begin{align*}
\Expect{\bZ_i\bZ_i^\top}
=
d\sigma^2\,\Eye_d
\qquad\text{and}\qquad
\Expect{\bZ_i^\top\bZ_i}
=
\sigma^2\,\Eye_d.
\end{align*}
This implies that \cref{eq:mc-sigma-bound} and \cref{eq:mc-sigma-r-bound} hold in this case as well. Since $\gmin$ and $\Cutoff{}$ are the same as \cref{subsec:matrix-completion}, we conclude that \cref{eq:mc-error-bound} holds, with the same probability as in \cref{subsec:matrix-completion}, when $n\geq C_2d\log^3d$.

\subsection{Compressed sensing via Gaussian ensembles}\label{subsec:gauss-ensemble}
Let $\Param$ be a $d\times d$ matrix.
Let each $\bX_i$ be a random matrix with entries filled with i.i.d. samples drawn from $\Normal{0,1}$. It follows that
\begin{align*}
\NormLTP{\bB}=\NormF{\bB}\,,
\end{align*}
which means that $\gmin=1$, and
\begin{align*}
\NormSGB{\bB}\leq 2\NormF{\bB}\,.
\end{align*}
To see the latter, since $\Inner{\bB,\bX_i}\sim\Normal{0,\NormF{\bB}^2}$ by \cref{lem:expect-gaussian2} of \cref{sec:aux}, it follows that
\begin{align*}
\Expect{\exp\left(\frac{\Abs{\Inner{\bB,\bX_i}}^2}{4\NormF{\bB}^2}\right)}
=
\left[\frac{1}{\sqrt{\left(1-\frac{1}{2}\right)_+}}\right]
\leq
2\,.
\end{align*}
Therefore, as before, we can use $\SpikeNorm(\cdot)=\NormSGB{\cdot}$ and $\Cutoff{}=9$. Therefore, a similar argument to those of \cref{subsec:matrix-completion,subsec:bandit-regression} shows that \cref{eq:mc-error-bound} holds in this setting as well. This bound closely recovers the bound of \cite{candes2011tight}.

\subsection{Compressed sensing via factored measurements}\label{subsec:factored-meas}

\citet{recht2010guaranteed} propose factored measurements to alleviate the need for a memory of size $nd^2$ to carry computations in the compressed sensing applications with large dimensions. The idea is to use measurement matrices of the form $\bX_i=UV^\top$, where $U$ and $V$ are random vectors of length $d$. Even though $UV^\top$ is a $d\times d$ matrix, we only need a memory of size $\Order{nd}$ to store all the input, which is a significant improvement over the Gaussian ensembles of \cref{subsec:gauss-ensemble}. We will now study this problem when $U$ and $V$ are both $\Normal{0,\Eye_d}$ vectors that are independent of each other. In this case we have
\begin{align*}
\NormLTP{\bB}^2
&=
\Expect*{\Inner{\bB, UV^\top}^2}\\
&=
\Expect*{(V^\top\bB^\top U)(U^\top\bB V)}\\
&=
\Expect*{V^\top\bB^\top\cdot \Expect*{UU^\top\Given V}\cdot\bB V}\\
&=
\Expect*{V^\top\bB^\top\bB V}\\
&=
\Expect*{\Trace{V^\top\bB^\top\bB V}}\\
&=
\Expect*{\Trace{\bB VV^\top\bB^\top}}\\
&=
\Trace{\bB \Expect{VV^\top}\bB^\top}\\
&=
\Trace{\bB\bB^\top}\\
&=
\NormF{\bB}^2\,,
\end{align*}
which means that the parameter $\gmin$ can be selected to be $1$ again. Next, let $\bB=\bO_1\bD\bO_2^\top$ be the singular value decomposition of $\bB$. Then, we get
\begin{align*}
\Inner{\bB,UV^\top}
&=
U^\top\bB V\\
&=
U^\top\bO_1\bD\bO_2^\top V\\
&=
(\bO_1^\top U)^\top\bD(\bO_2^\top V)\,.
\end{align*}
As the distribution of $U$ and $V$ is invariant by the multiplication of unitary matrices, for any $t>0$, we have
\begin{align*}
\Expect*{\exp\left(\frac{\Inner{\bB,UV^\top}}t\right)}
&=
\Expect*{\exp\left(\frac{\Inner{\bD,UV^\top}}t\right)}\\
&=
\Expect*{\exp\left(\frac{U^\top\bD V}t\right)}\\
&=
\Expect*{U}{\Expect*{V}{\exp\left(\frac{U^\top\bD V}t\right)\Given U}}\\
&=
\Expect*{\exp\left(\frac{\Norm{U^\top\bD}^2}{2t^2}\right)}\\
&=
\Expect*{\exp\left(\frac{\sum_{i=1}^{d}U_i^2\bD_{ii}^2}{2t^2}\right)}\\
&=
\prod_{i=1}^{d}\Expect*{\exp\left(\frac{U_i^2\bD_{ii}^2}{2t^2}\right)}\\
&=
\prod_{i=1}^{d}\frac{1}{\sqrt{\left(1-\frac{\bD_{ii}^2}{t^2}\right)_+}}.
\end{align*}
By \cref{lem:mgf-to-orlicz} of \cref{sec:aux}, the necessary condition for $\NormSubE{\Inner{\bB,UV^\top}}\leq t$ to hold is
\begin{align}
\label{eq:nec-cond-for-mgf}
\Expect{\exp\left(\frac{\Inner{\bB,UV^\top}}t\right)}\leq2\,.
\end{align}
This, in particular, implies that
\begin{align*}
\frac{1}{\sqrt{\left(1-\frac{\bD_{ii}^2}{t^2}\right)_+}}\leq2
\end{align*}
or equivalently
\begin{align*}
\frac{\bD_{ii}^2}{t^2}\leq \frac34\,,
\end{align*}
for all $i\in[d]$. By taking the derivatives and by applying the concavity of logarithm, we can observe that $-2x\leq\log (1-x)\leq -x$ for all $x\in[0,\frac34]$. This implies that, whenever \cref{eq:nec-cond-for-mgf} holds, we have
\begin{align*}
\frac{\bD_{ii}^2}{2t^2}\leq\log\left(1-\frac{\bD_{ii}^2}{t^2}\right)_+^{-\frac12}\leq \frac{\bD_{ii}^2}{t^2},
\end{align*}
and thus
\begin{align*}
\exp\left(\frac{\sum_{i=1}^{d}\bD_{ii}^2}{2t^2}\right)\leq\Expect{\exp\left(\frac{\Inner{\bB,UV^\top}}t\right)}\leq\exp\left(\frac{\sum_{i=1}^{d}\bD_{ii}^2}{t^2}\right)\,.
\end{align*}
Using $\NormF{\bB}^2=\sum_{i=1}^d\bD_{ii}^2$, we can simplify the above to
\begin{align}
\label{eq:factor-mgf-simple-bound}
\exp\left(\frac{\NormF{\bB}^2}{2t^2}\right)
\leq
\Expect{\exp\left(\frac{\Inner{\bB,UV^\top}}t\right)}
\leq
\exp\left(\frac{\NormF{\bB}^2}{t^2}\right)\,.
\end{align}
Combining the above results, we  conclude that $\NormSubE{\Inner{\bB,\bX}}\leq t$ implies
\begin{align*}
\frac{2\bD_{11}}{\sqrt3}\leq t
\qquad\text{and}\qquad
\frac{\NormF{\bB}}{\sqrt{2\log2}}\leq t\,.
\end{align*}
Therefore, we have
\begin{align}
\label{eq:factor-norm-lower-bound}
\frac{1}{\sqrt{2\log 2}}\NormF{\bB}\leq\Norm{\bB}_{\psi_1(\SampleDistr)}\,.
\end{align}
Next, define
\begin{align*}
t
:=
\max\{\frac{2\bD_{11}}{\sqrt{3}}, \frac{\NormF{\bB}}{\sqrt{\log2}} \}
=\frac{\NormF{\bB}}{\sqrt{\log2}}\,.
\end{align*}
Since $D_{11}^2/t^2\leq 3/4$, we can use \cref{eq:factor-mgf-simple-bound} to obtain
\begin{align*}
\Expect*{\exp\left(\frac{\Inner{\bB,UV^\top}}t\right)}\leq 2\,.
\end{align*}
Using \cref{lem:mgf-to-orlicz} of \cref{sec:aux}, we conclude that
\begin{align*}
\Norm{\bB}_{\psi_1(\SampleDistr)}
\leq
\frac{8}{\sqrt{\log 2}}\NormF{\bB}\,.
\end{align*}
Now, by setting $\SpikeNorm(\cdot)=\Norm{\cdot}_{\psi_1(\SampleDistr)}$, given that the ratio $\Norm{\bB}_{\psi_1(\SampleDistr)}/\NormF{\bB}$ is at most $8/\sqrt{\log(2)}$,
we apply \cref{cor:orlicz-trun-second-moment-bound} of \cref{sec:aux} to $Z=\Inner{\bB,\bX}$
to see that $\Cutoff{}=53$ satisfies \cref{as:x-tilde-boundedness-stochastic}.

Now, in order to bound $\Prob{\lambda<3\NormOp{\bSigma}}$ we need to use a truncation argument for the noise. Specifically, let
\[
\barNoise_i:=\Noise_i\,\II\left[\Abs{\Noise_i}\leq C_\Noise\sigma\sqrt{\log d}\right]\,,
\]
for a large enough constant $C_\Noise$. Next, using a union bound and defining $\tSigma:=\sum_{i=1}^{n}\barNoise_i\bX_i$, we have
\begin{align*}
  \Prob*{\lambda<3\NormOp{\bSigma}} &\leq \Prob*{\lambda<3\NormOp{\tSigma}} + \sum_{i=1}^{n}\Prob*{\Abs{\Noise_i}> C_\Noise\sigma\sqrt{\log d}} \\
  &\leq \Prob*{\lambda<3\NormOp{\tSigma}} + 2ne^{-\frac{C_\Noise^2}{2}\log d}\,.
\end{align*}
Now, we define $\delta: = 2C_\Noise\sigma d\sqrt{\log d}$ and $\bZ_i=\barNoise_i\bX_i$, as in \cref{subsec:matrix-completion}, and use \cref{lem:freedman}. Let $(G_i)_{i=1}^{2d}$ be a sequence of $2d+1$ independent standard normal random variables. We can use similar steps as in \cref{subsec:matrix-completion,subsec:bandit-regression} to obtain
\begin{align*}
\Expect*{\exp\left(\frac{\NormOp{\bZ_i}}{\delta}\right)}
&=
\Expect*{\exp\left(\frac{|\barNoise_i|\sqrt{\sum_{j=1}^{d}G_j^2}\sqrt{\sum_{j=d+1}^{2d}G_j^2}}{\delta}\right)}\\
&\leq \Expect*{\exp\left(\frac{|\barNoise_i|\sum_{j=1}^{2d}G_j^2}{2\delta}\right)}\\
&\leq \Expect*{\Expect*{\exp\left(\frac{|\barNoise_i|G_1^2}{2\delta}\right)}^{2d}\Big|\barNoise_i}\\
&\leq \Expect*{\exp\left(\frac{G_1^2}{4d}\right)}^{2d}\\
&\leq \left(\frac{1}{1-\frac{1}{2d}}\right)^{2d}\\
&\leq e\,.
\end{align*}
Furthermore, we have
\begin{align*}
\Expect*{\bZ_i\bZ_i^\top}
\preceq
d\sigma^2\,\Eye_d
\qquad\text{and}\qquad
\Expect{\bZ_i^\top\bZ_i}
\preceq
\sigma^2\,\Eye_d.
\end{align*}
Therefore, the following slight variation of \cref{eq:mc-sigma-bound} holds:
\begin{align}
\label{eq:fm-sigma-bound}
\Prob{\lambda<3\NormOp{\bSigma}}
\leq
2d\exp\left[-\frac{Cn\lambda}{d\sigma}\left(\frac{\lambda}{\sigma}\wedge\frac1{\log^{3/2} d}\right)\right]+2ne^{-\frac{C_\Noise^2}{2}\log d}\,.
\end{align}
However, \cref{eq:mc-sigma-r-bound} stays unchanged since for $\zeta_i$, unlike for $\Noise_i$, we do not need to use any truncation. This means that we can define $\lambda$ as in \cref{eq:lambda-def-mc} and obtain a bound as in \cref{eq:mc-error-bound}
with probability at least $1-\exp(-C_8\rho)$ whenever $\rho\ge \log d$ and $n\geq C_2d\log^4d$. This bound recovers Theorem 2.3 of \cite{cai2015rob}; however, their bound works for $n=O(rd)$ which is smaller than our bound when $r< \log^4d$. 

\section{Applications to Exact Recovery} \label{sec:exact-recovery}

In this section we study the trace regression problem when there is no noise. It is known that, under certain assumptions, it is possible to recover the true matrix $\Param$ exactly, with high probability \citep{candes2009exact,keshavan2009matrix}. The discussion of Section 3.4 in \citep{negahban2012restricted}, in the context of the matrix-completion problem, we can deduce that that upper bounds that rely on the spikiness of $\Param$ are not strong enough to guarantee exact recovery of $\Param$, even in the noiseless setting. However, will will show in this section that the methodology from \cref{sec:trace-reg} can be used to prove exact recovery for the two cases of compressed sensing studied in previous sections (\cref{subsec:gauss-ensemble,subsec:factored-meas}). After that, we will conclude this section with a brief discussion on exact recovery for the multi-task learning case (\cref{subsec:bandit-regression}).

For any arbitrary sampling operator $\SamplingOp(\cdot)$, let $\Sset$ be defined as follows:
\begin{align*}
\Sset:=\left\{\bB\in\IR^{d_r\times d_c}:\SamplingOp(\bB)=Y \right\}\,.
\end{align*}
Using $\sigma_\Noise=0$ and the linear model \cref{eq:generative-model}, one can verify that $\Param\in\Sset$ and thus that $\Sset$ is not empty. The definition of $\Sset$ implies that $\Sset$ is an affine space and is thus convex. Next, note that, for any $\bB\in\Sset$, the following identity holds:
\begin{align*}
\Loss(\bB)=\frac1n\Norm{Y-\SamplingOp(\bB)}_2^2+\lambda\NormNuc{\bB}=\lambda\NormNuc{\bB}\,.
\end{align*}
Therefore, the minimizers of the optimization problem
\begin{align*}
\Minimize\quad & \frac1n\Norm{Y-\SamplingOp(\bB)}_2^2+\lambda\NormNuc{\bB}\\
\SubjectTo\quad & \SamplingOp(\bB)=Y\,,
\end{align*}
are also the minimizers of
\begin{align*}
\Minimize\quad & \NormNuc{\bB}\\
\SubjectTo\quad & \SamplingOp(\bB)=Y\,.
\end{align*}
Note that, in the above formulation, the convex problem does not depend on $\lambda$ anymore, and so $\lambda$ can be chosen arbitrarily. In the noiseless setting, $\NormOp{\bSigma}=0$, and so any $\lambda>0$ satisfies \cref{eq:lambda-condition}. Therefore, if the RSC condition holds for $\SamplingOp(\cdot)$ with parameters $\alpha$ and $\beta$ on the set $\Cset(0,\eta)$, \cref{thm:deterministic-error-bound} leads to
\begin{align}
\label{eq:noiseless-bound-with-beta}
\NormF[\big]{\Estimator-\Param}^2
&\leq
\frac{8\ParamBound^2\beta}\alpha.
\end{align}
Now, defining $\nu_0$ as
\begin{align}
\label{eq:nu0-def}
\nu_0:=\inf_{\bB\neq0}\frac{\NormF{\bB}^2}{\SpikeNorm(\bB)^2},
\end{align}
one can easily observe that $\Cset(\nu_0,\eta)=\Cset(0,\eta)$. Moreover, assume that $n$ is large enough so that
\begin{align}
\label{eq:exact-n-condition}
\Expect*{\NormOp[\big]{\bSigma_R}}^2
\leq
\frac{\gmin^2\nu_0}{800\Cutoff^2\eta}\,,
\end{align}
where $\bSigma_R$ is defined in \cref{sec:trace-reg}. Combining this with \cref{cor:rsc-condition}, we obtain that
\begin{align*}
\frac{\Norm{\SamplingOp(\bA)}_2^2}{n}
&\geq
\frac\gmin4\NormF{\bA}^2-\frac{93\eta\Cutoff^2}{\gmin}\Expect{\NormOp[\big]{\bSigma_R}}^2\\
&\geq
\frac\gmin4\NormF{\bA}^2-\frac{\gmin\nu_0}{8}\\
&\geq
\frac\gmin4\NormF{\bA}^2-\frac{\gmin}{8}\NormF{\bA}^2\\
&=
\frac\gmin8\NormF{\bA}^2\,,
\end{align*}

with probability at least
\begin{align*}
	1-2\exp\left(-\frac{Cn\gmin\nu_0}{\Cutoff^2}\right)\,,
\end{align*}
for all $\bA\in\Cset(\nu_0,\eta)$. This shows that $\SamplingOp(\cdot)$ satisfies the RSC condition with $\alpha=\gmin/8$ and $\beta=0$. As a result, from \cref{eq:noiseless-bound-with-beta}, we can deduce the following proposition.
\begin{prop}
\label{prop:exact-recovery}
Let $\nu_0$ and $n$ be as in \cref{eq:nu0-def} and \cref{eq:exact-n-condition}. Then, the unique minimizer of the constraint optimization problem
\begin{align*}
\Minimize\quad & \NormNuc{\bB}\\
\SubjectTo\quad & \SamplingOp(\bB)=Y
\end{align*}
is $\bB=\Param$, and hence exact recovery of $\Param$ is possible with probability at least $1-2\exp\left(-\frac{Cn\gmin\nu_0}{\Cutoff^2}\right)$.
\end{prop}
Next, we can use the above proposition to prove that exact recovery is possible for the two problems of  compressed sensing with Gaussian ensembles (\cref{subsec:gauss-ensemble}) and compressed sensing with factored measurements (\cref{subsec:factored-meas}). Note that in both cases, we have $\gmin=1$, $\nu_0\geq0.1$, and $\Cutoff\leq60$.

Therefore, in order to use \cref{prop:exact-recovery}, all we need to do is to find a lower bound for $n$ such that \cref{eq:exact-n-condition} holds. We will study each case in turn.
\begin{enumerate}
\item {\bf Compressed sensing with Gaussian ensembles.} Here, since entries of $\bX_i$'s are i.i.d. $\Normal{0,1}$ random variables, the entries of $\bSigma_R$ are i.i.d. $\Normal{0,1/n}$  random variables. We can thus use Theorem 5.32 of \cite{vershynin2010introduction} to get
\begin{align*}
\Expect*{\NormOp{\bSigma_R}}\leq \frac{2\sqrt d}{\sqrt n}\,.
\end{align*}
Therefore, \cref{eq:exact-n-condition} is satisfied if $n\geq Crd$, where $C>0$ is a large enough constant.

\item {\bf Compressed sensing with factored measurements.} Here, the observation matrices are of the form $U_iV_i^\top$, where $U_i$ and $V_i$ are independent vectors distributed according to $\Normal{0,\Eye_d}$. Note that we have $\NormOp{\bX_i}=\Norm{U_i}_2\Norm{V_i}_2\leq\Norm{U_i}_2^2+\Norm{V_i}_2^2$. Then,
\begin{align*}
\NormSubE*{\NormOp{\bX_i}}
&\leq
\NormSubE*{\Norm{U_i}_2^2+\Norm{V_i}_2^2}\\
&\leq
2\NormSubE*{\Norm{U_i}_2^2}\\
&=
O(d)\,.
\end{align*}
By Eq. (3.9) of \cite{koltchinskii2011neumann} we have
\begin{align*}
\Expect*{\NormOp{\bSigma_R}}
= O\left(\sqrt{\frac{d\log(2d)}{n}}\vee\frac{d\log(2d)^2}{n}\right)\,.
\end{align*}
We can thus infer that \cref{eq:exact-n-condition} holds for all $n\geq Crd\log(d)$ where $C$ is a large enough constant.
\end{enumerate}
Therefore, \cref{prop:exact-recovery} guarantees that, for $n$ satisfying the conditions stated above, exact recovery is possible in each of the two aforementioned settings, with probability at least $1-\exp(C'n)$, where $C'>0$ is a numerical constant.

\paragraph{Implications for multi-task learning.} We can apply  \cref{prop:exact-recovery} to the multi-task learning case (\cref{subsec:bandit-regression}) as well. We have $\gmin=1$ and $\Cutoff=9$, but, for $\nu_0$, we have
\begin{align*}
\nu_0
&=
\inf_{\bB\neq0}\frac{\NormF{\bB}^2}{\SpikeNorm(\bB)^2}\\
&=
\inf_{\bB\neq0}\,\frac{\NormF{\bB}^2}{4d\Norm{\bB}_{2,\infty}^2}\\
&=
\frac{1}{4d}\,,
\end{align*}
where the infimum is achieved if and only if $\bB$ has exactly one non-zero row. Notice that,  in contrast to the previous examples where we had $\nu_0\geq0.1$, $\nu_0$ depends on the dimensions of the matrix.

It is straightforward to verify that
\begin{align*}
\NormSubG*{\NormOp{\bX_i}}
&=
\NormSubG[\Big]{\NormTwo{X_i}}\\
&=
O(\sqrt{d})\,.
\end{align*}
Therefore, a similar argument to the above shows that
\begin{align*}
\Expect*{\NormOp{\bSigma_R}}
= O\left(\sqrt{\frac{d\log(2d)}{n}}\vee\frac{\sqrt{d}\log(2d)^{\frac32}}{n}\right)\,.
\end{align*}
This, in turn, implies that, in order for \cref{eq:exact-n-condition} to hold, it suffices to have $n\geq Crd^2\sqrt{d}\log(2d)$, for a large enough constant $C$. In this case, \cref{prop:exact-recovery} shows that exact recovery is possible with probability at least $1-2\exp(-\frac{C'n}{d})$. However, this result is trivial, since $n\geq Crd^2\sqrt{d}\log(2d)$ means that with high probability, each row is observed at least $d$ times, and so each row can be reconstructed separately (without using the low-rank assumption). This result can not be improved without further assumptions, as it is possible in a rank-2 matrix that all rows are equal to each other except for one row, and that row can be reconstructed only if at least $d$ observations are made for that row. Since this must hold for all rows, at least $d^2$ observations would be needed. Nonetheless, one can expect that with stronger assumptions than generalized spikiness, such as incoherence, the number of required observations can be reduced to $rd\log(d)$.

\section*{Acknowledgments}
The authors gratefully acknowledge support of the National Science Foundation (CAREER award CMMI: 1554140) and Stanford Data Science Initiative.
\appendix
\section{Auxiliary proofs}
\label{sec:aux}

\begin{lem}
\label{lem:expect-gaussian2}
Let $Z$ be an $\Normal{0,\sigma^2}$ random variable. Then, for all $\eta>0$,
\begin{align*}
\Expect*{e^{\eta Z^2}} = \frac{1}{\sqrt{(1-2\sigma^2\eta)_+}}\,.
\end{align*}
\end{lem}
\begin{proof}
Easily follows by using the formula $\int_{-\infty}^{\infty} \exp(-\frac{t^2}{2a^2}) dt = \sqrt{2\pi a^2}$.
\end{proof}

\begin{lem}
\label{lem:orlicz-trun-second-moment-bound}
Let $Z$ be a nonnegative random variable such that $\Norm{Z}_{\psi_p}=\nu$ holds for some $p\geq 1$, and assume that $c>0$ is given. Then, we have
\begin{align*}
\Expect*{Z^2\cdot\II(Z\geq c)}
&\leq
(2c^2+4c\nu+4\nu^2)\cdot\exp\left(-\frac{c^p}{\nu^p}\right)\,.
\end{align*}
\end{lem}
\begin{proof}
Without loss of generality, we can assume that $Z$ has a density function $f(z)$ and $\nu=1$. Moreover, let $F(z):=\Prob{Z\leq z}$ be the cumulative distribution function of $Z$. The assumption that $\Norm{Z}_{\psi_p}=1$ together with the Markov inequality yields
\begin{align*}
F(z)\geq 1-2\exp\left(-z^p\right).
\end{align*}
Therefore,
\begin{align*}
\Expect*{Z^2\cdot\II(Z\geq c)}
&=
\int_{c}^{\infty}z^2f(z)\,dz\\
&=
\int_{c}^{\infty}(-z^2)[-f(z)]\,dz\\
&=
c^2\left[1-F(c)\right]+\int_{c}^{\infty}2z[1-F(z)]\,dz\\
&\leq
2c^2\cdot\exp\left(-c^p\right)+\int_{c}^{\infty}4z\exp\left(-z^p\right)\,dz\,.
\end{align*}
Next, note that the function $f_c(p)$ defined as
\begin{align*}
f_c(p)
:=
\frac{\int_{c}^{\infty}z\exp\left(-z^p\right)\,dz}{(c+1)\exp\left(-c^p\right)}
=
\frac{\int_{c}^{\infty}z\exp\left(c^p-z^p\right)\,dz}{c+1}\,,
\end{align*}
is decreasing in $p$. Therefore, we have
\begin{align*}
\int_{c}^{\infty}z\exp\left(-z^p\right)\,dz
&\leq
(c+1)\exp\left(-c^p\right)f_c(1)\\
&=
(c+1)\exp\left(-c^p\right)\,,
\end{align*}
using $f_c(1)=1$. Therefore, we have
\begin{align*}
\Expect*{Z^2\cdot\II(Z\geq c)}
&\leq
(2c^2+4c+4)\cdot\exp\left(-c^p\right)\,,
\end{align*}
which is the desired result.
\end{proof}
\begin{cor}
\label{cor:orlicz-trun-second-moment-bound}
Let $Z$ be a random variable such that $\Norm{Z}_{\psi_p}=\nu$ holds for some $p\geq 1$ and $\Expect*{Z^2}=\sigma^2$. Then, for
\begin{align}
c_{\sigma,p}&:=\nu\cdot\max\left\{5, \left[10\log(\frac{2\nu^2}{\sigma^2})\right]^{\frac1p}\right\}\,,\label{eq:constant-c=sigma}
\end{align}
we have
\begin{align*}
\Expect*{Z^2\cdot\II\left(\Abs{Z}\leq c_{\sigma,p}\right)}
&\geq\frac{\Expect*{Z^2}}{2}\,.
\end{align*}
\end{cor}
\begin{proof}
Without loss of generality, we can assume that $\nu=1$. Using \cref{lem:orlicz-trun-second-moment-bound} for $|Z|$ and any $c\geq5$, we have
\begin{align*}
\Expect*{Z^2\cdot\II\left(\Abs{Z}\leq c\right)}
&\geq
\Expect*{Z^2}-\Expect*{Z^2\cdot\II\left(\Abs{Z}\geq c\right)}\\
&\geq
\Expect*{Z^2}-3c^2\cdot\exp\left(-c^p\right)\,.
\end{align*}
Next, it is easy to show that, for any $c\geq5$,
\begin{align*}
3c^2\cdot\exp\left(-\frac{9c^p}{10}\right)
\leq
3c^2\cdot\exp\left(-\frac{9c}{10}\right)
\leq
1\,.
\end{align*}
Therefore, letting $c_\sigma$ be defined as in \cref{eq:constant-c=sigma}, we get
\begin{align*}
3c_{\sigma,p}^2\cdot\exp\left(-c_{\sigma,p}^p\right)
&=
3c_{\sigma,p}^2\cdot\exp\left(-\frac{9c_{\sigma,p}^p}{10}\right)\exp\left(-\frac{c_{\sigma,p}^p}{10}\right)\\
&\leq
\exp\left(-\frac{c_{\sigma,p}^p}{10}\right)\\
&\leq
\frac{\sigma^2}{2}\,,
\end{align*}
which completes the proof of this corollary.
\end{proof}
\begin{cor}
\label{cor:normal-c-nu}
Let $Z$ be an $\Normal{0,\sigma^2}$ random variable. Then, the constant $c_{\sigma,2}$ defined in \cref{eq:constant-c=sigma} satisfies $c_{\sigma,2}\leq 5\Norm{Z}_{\psi_2}$.
\end{cor}
\begin{proof}
By \cref{lem:expect-gaussian2}, we obtain $\nu=\Norm{Z}_{\psi_2}=\sqrt{8\sigma^2/3}$, which means that $\nu^2/\sigma^2=8/3$. The rest follows from \cref{cor:orlicz-trun-second-moment-bound}.
\end{proof}

The Orlicz norm of a random variable is defined in terms of the absolute value of that random variable, and it is usually easier to work with the random variable than with its absolute value. The next lemma relates the Orlicz norm to the moment-generating function of a random variable.
\begin{lem}
\label{lem:mgf-to-orlicz}
Let $X$ be a zero-mean random variable and
\begin{align*}
\alpha:=\inf\left\{t>0:\max\left\{\Expect*{\exp\left(\frac{X}{t}\right)},\Expect*{\exp\left(-\frac{X}{t}\right)}\right\}\leq2 \right\}.
\end{align*}
Then, we have
\begin{align*}
\alpha\leq\Norm{X}_{\psi_1}\leq8\alpha.
\end{align*}
\end{lem}
\begin{proof}
The first inequality, $\alpha\leq\Norm{X}_{\psi_1}$, follows from the monotonicity of the exponential function. For the second inequality, note that for any $t>0$,
\begin{align}
\Expect*{\exp\left(\frac{\Abs{X}}{t}\right)}
&=
\Expect*{\exp\left(\frac{\Abs{X}-\Expect*{\Abs{X}}}{t}\right)}\cdot\exp\left(\frac{\Expect*{\Abs{X}}}{t}\right)\label{eq:lemma-4-proof}\,.
\end{align}
Now, the union bound and Markov inequality lead to the following tail bound for $\Abs{X}$:
\begin{align*}
\Prob{\Abs{X}\geq x}&\leq \Prob{X\geq x}+\Prob{-X\geq x}\\
&\leq 4\exp\left(-\frac{x}{\alpha}\right)\,.
\end{align*}
Hence, we have
\begin{align*}
\Expect*{\Abs{X}}
&=
\int_0^\infty \Prob{\Abs{X}\geq x}dx\\
&\leq
4\alpha.
\end{align*}
Next, assuming that $X'$ is an independent copy of $X$ and $\Rademacher$ is a Rademacher random variable independent of $X$ and $X'$, we have
\begin{align*}
\Expect*{\exp\left(\frac{\Abs{X}-\Expect*{\Abs{X}}}{t}\right)}
&=
\Expect*{\exp\left(\frac{\Abs{X}-\Expect*{\Abs{X'}}}{t}\right)}\\
&\leq
\Expect*{\exp\left(\frac{\Abs{X}-\Abs{X'}}{t}\right)} & \text{(by Jensen's inequality)}\\
&=
\Expect*{\exp\left(\frac{\Rademacher(\Abs{X}-\Abs{X'})}{t}\right)}\\
&=
\Expect*{\Expect*{\exp\left(\frac{\Rademacher\Abs[\big]{\Abs{X}-\Abs{X'}}}{t}\right)\Given X,X'}}\\
&\overset{(*)}{\leq}
\Expect*{\Expect*{\exp\left(\frac{\Rademacher\Abs{X-X'}}{t}\right)\Given X,X'}}\\
&=
\Expect*{\Expect*{\exp\left(\frac{\Rademacher(X-X')}{t}\right)\Given X,X'}}\\
&=
\Expect*{\exp\left(\frac{\Rademacher(X-X')}{t}\right)}\\
&=
\frac12\Expect*{\exp\left(\frac{X-X'}{t}\right)}+\frac12\Expect*{\exp\left(\frac{X'-X}{t}\right)}\\
&=
\Expect*{\exp\left(\frac{X-X'}{t}\right)}\\
&=
\Expect*{\exp\left(\frac{X}{t}\right)}\cdot\Expect*{\exp\left(-\frac{X}{t}\right)}\,,
\end{align*}
where (*) follows from $\Abs[\big]{\Abs{a}-\Abs{b}}\leq\Abs[\big]{a-b}$ and the fact that the function $z\mapsto\frac12(\exp(-z)+\exp(z))$ is increasing in $z$ when $z>0$.

Therefore, from the above inequalities, we can deduce that
\begin{align*}
\Expect*{\exp\left(\frac{\Abs{X}}{t}\right)}
&\leq
\Expect*{\exp\left(\frac{X}{t}\right)}\cdot\Expect*{\exp\left(-\frac{X}{t}\right)}\cdot\exp\left(\frac{4\alpha}{t}\right).
\end{align*}
Now, by setting $t:=8\alpha$ and using Jensen's inequality, we get
\begin{align*}
\Expect*{\exp\left(\frac{\Abs{X}}{t}\right)}
&\leq
\Expect*{\exp\left(\frac{X}{\alpha}\right)}^{\frac18}\cdot\Expect*{\exp\left(-\frac{X}{\alpha}\right)}^{\frac18}\cdot\exp\left(\frac12\right)\\
&\leq
\exp\left(\frac{\log2+2}4\right)\\
&\leq
2.
\end{align*}
This implies that $\NormSubE{X}\leq t=8\alpha$.
\end{proof}
\begin{lem}
\label{lem:norm-se-exp-x-leq-norm-se-x}
For any subexponential random variable $X$, we have
\begin{align*}
\NormSubE{\Expect*{X}}\leq\NormSubE{X}.
\end{align*}
\end{lem}
\begin{proof}
\begin{align*}
\NormSubE*{\Expect*{X}}
&=
\inf\Set*{t>0:\exp\left(\frac{\Abs{\Expect*{X}}}{t}\right)\leq 2}\\
&=
\frac{\Abs{\Expect*{X}}}{\log 2}\\
&\leq
\frac{\Expect*{\Abs{X}}}{\log 2}\\
&=
\frac{\NormSubE{X}\cdot\log\exp\Expect*{\frac{\Abs{X}}{\NormSubE{X}}}}{\log 2}\\
&\leq
\frac{\NormSubE{X}\cdot\log\Expect*{\exp\frac{\Abs{X}}{\NormSubE{X}}}}{\log 2}\\
&\leq
\frac{\NormSubE{X}\cdot\log2}{\log 2}\\
&=
\NormSubE{X}.
\end{align*}
\end{proof}

\section{Trace regression proofs, adapted from \citep{klopp2014noisy}}
\label{sec:klopp-pf}

\subsection{Proof of \cref{thm:deterministic-error-bound}}
\label{subsec:pf-thm-deterministic-error-bound}

\begin{proof}
First, it follows from \cref{eq:goodness-condition} that
\begin{align*}
\frac{1}{n}\Norm[\big]{Y-\SamplingOp(\Estimator)}_2^2+\lambda\NormNuc[\big]{\Estimator}
\leq
\frac{1}{n}\Norm[\big]{Y-\SamplingOp(\Param)}_2^2+\lambda\NormNuc[\big]{\Param}.
\end{align*}
By substituting $Y$ for $\SamplingOp(\Param)+E$ and doing some algebra, we have
\begin{align*}
\frac1n\NormTwo[\big]{\SamplingOp(\Param-\Estimator)}^2
+2\Inner[\big]{\bSigma,\Param-\Estimator}
+\lambda\NormNuc[\big]{\Estimator}
\leq
\lambda\NormNuc[\big]{\Param}.
\end{align*}
Then, using the duality between the operator norm and the trace norm, we get
\begin{align}
\frac1n\NormTwo[\big]{\SamplingOp(\Param-\Estimator)}^2
+\lambda\NormNuc[\big]{\Estimator}
\leq
2\NormOp[\big]{\bSigma}\cdot\NormNuc[\big]{\Param-\Estimator}
+
\lambda\NormNuc[\big]{\Param}.
\label{eq:basic-inequality-2}
\end{align}
For a given set of vectors $S$, we denote by $\bP_S$ the orthogonal projection on the linear subspace spanned by elements of $S$ (i.e., $\bP_S=\sum_{i=1}^{k}u_iu_i^\top$ if $\{u_1,\ldots,u_k\}$ is an orthonormal basis for $S$). For matrix $\bB\in\IR^{d_r\times d_c}$, let $S_r(\bB)$ and $S_c(\bB)$ be the linear subspace spanned by the left and right orthonormal singular vectors of $\bB$, respectively. Then, for $\bA\in\IR^{d_r\times d_c}$ define
\begin{align*}
\bP_\bB(\bA):=\bA-\bP_{\bB}^\perp(\bA)
\quad\text{and}\quad
\bP_{\bB}^\perp(\bA):=\bP_{S_r^\perp(\bB)}\bA\bP_{S_c^\perp(\bB)}\,.
\end{align*}
We can alternatively express $\bP_\bB(\bA)$ as
\begin{align}
\bP_\bB(\bA)
&=
\bA-\bP_{\bB}^\perp(\bA)\nonumber\\
&=
\bP_{S_r(\bB)}\bA+\bP_{S_r^\perp(\bB)}\bA-\bP_{\bB}^\perp(\bA)\nonumber\\
&=
\bP_{S_r(\bB)}\bA+\bP_{S_r^\perp(\bB)}[\bA-\bA\bP_{S_c^\perp(\bB)}]\nonumber\\
&=
\bP_{S_r(\bB)}\bA+\bP_{S_r^\perp(\bB)}\bA\bP_{S_c(\bB)}\,.
\label{eq:PB-equivalent}
\end{align}
In particular, since $S_r(\bB)$ and $S_c(\bB)$ both have dimension $\Rank(\bB)$, it follows from \cref{eq:PB-equivalent} that
\begin{align}
\Rank(\bP_\bB(\bA))
&\leq
2\Rank(\bB)\,.
\label{eq:PB-rank}
\end{align}
Moreover, the definition of $\bP_{\bB}^\perp$ implies that the left and right singular vectors of $\bP_{\bB}^\perp(\bA)$ are orthogonal to those of $\bB$. We thus have
\begin{align*}
\NormNuc{\bB+\bP_{\bB}^\perp(\bA)}=\NormNuc{\bB}+\NormNuc{\bP_{\bB}^\perp(\bA)}\,.
\end{align*}
Set $\bB:=\Param$ and $\bA:=\Estimator-\Param$; then the above equality entails
\begin{align}
\NormNuc{\Param+\bP_{\Param}^\perp(\Estimator-\Param)}=\NormNuc{\Param}+\NormNuc{\bP_{\bB}^\perp(\Estimator-\Param)}\,.
\end{align}
We can then use the latter to get the following inequality:
\begin{align}
\NormNuc[\big]{\Estimator}
&=
\NormNuc[\big]{\Param+\Estimator-\Param}\nonumber\\
&=
\NormNuc[\big]{\Param+\bP_{\Param}^\perp(\Estimator-\Param)+\bP_{\Param}(\Estimator-\Param)}\nonumber\\
&\geq
\NormNuc[\big]{\Param+\bP_{\Param}^\perp(\Estimator-\Param)}
-\NormNuc[\big]{\bP_{\Param}(\Estimator-\Param)}\nonumber\\
&=\NormNuc[\big]{\Param}+\NormNuc[\big]{\bP_{\Param}^\perp(\Estimator-\Param)}-\NormNuc[\big]{\bP_{\Param}(\Estimator-\Param)}.
\label{eq:bhat-nuc-norm-lower-bound}
\end{align}
Combining \cref{eq:basic-inequality-2} with \cref{eq:bhat-nuc-norm-lower-bound}, we get
\begin{align*}
\frac1n\Norm[\big]{\SamplingOp(\Param-\Estimator)}_2^2
&\leq
2\NormOp[\big]{\bSigma}\cdot\NormNuc[\big]{\Param-\Estimator}
+
\lambda\NormNuc[\big]{\bP_{\Param}(\Estimator-\Param)}-\lambda\NormNuc[\big]{\bP_{\Param}^\perp(\Estimator-\Param)}\\
&\leq
(2\NormOp[\big]{\bSigma}+\lambda)\NormNuc[\big]{\bP_{\Param}(\Estimator-\Param)}
+
(2\NormOp[\big]{\bSigma}-\lambda)\NormNuc[\big]{\bP_{\Param}^\perp(\Estimator-\Param)}\\
&\leq
\frac53\lambda\NormNuc[\big]{\bP_{\Param}(\Estimator-\Param)}\,,
\end{align*}
where, in the last inequality, we have used \cref{eq:lambda-condition}. Now, using this and the fact that
$\Rank(\bP_{\Param}(\Estimator-\Param))\leq 2\Rank(\Param)$ (by \cref{eq:PB-rank}), we can apply the Cauchy--Schwartz inequality to singular values of $\bP_{\Param}(\Estimator-\Param)$ to obtain
\begin{align}
\frac1n\Norm[\big]{\SamplingOp(\Param-\Estimator)}_2^2
&\leq
\frac53\lambda\sqrt{2\Rank(\Param)}\NormF[\big]{\bP_{\Param}(\Estimator-\Param)}\nonumber\\
&\leq
\frac53\lambda\sqrt{2\Rank(\Param)}\NormF[\big]{\Estimator-\Param}\,.
\label{eq:basic-inequality-3}
\end{align}
The next lemma makes a connection between $\Estimator$ and the constraint set $\Cset(\nu,\eta)$.
\begin{lem}\label{lem:B1}
If $\lambda\geq3\NormOp{\bSigma}$, then
\begin{align*}
\NormNuc[\big]{\bP_{\Param}^\perp(\Estimator-\Param)}
\leq
5\NormNuc[\big]{\bP_{\Param}(\Estimator-\Param)}.
\end{align*}
\label{lem:compatibility}
\end{lem}
\begin{proof}
Note that $\Norm{\SamplingOp(\cdot)}_2^2$ is a convex function. We can then use the convexity at $\Param$ to get
\begin{align*}
\frac1n\Norm[\big]{\SamplingOp(\Estimator)}_2^2-\frac1n\Norm{\SamplingOp(\Param)}_2^2
&\geq
-\frac2n\sum_{i=1}^{n}\left(y_i-\Inner[\big]{\bX_i,\Param}\right)\Inner[\big]{\bX_i,\Estimator-\Param}\\
&=
-2\Inner[\big]{\bSigma,\Estimator-\Param}\\
&\geq
-2\NormOp[\big]{\bSigma}\NormNuc[\big]{\Estimator-\Param}\\
&\geq
-\frac{2\lambda}3\NormNuc[\big]{\Estimator-\Param}\,.
\end{align*}
Combining this with \cref{eq:goodness-condition} and \cref{eq:bhat-nuc-norm-lower-bound}, we have
\begin{align*}
\frac{2\lambda}3\NormNuc[\big]{\Estimator-\Param}
&\geq
\frac1n\Norm[\big]{\SamplingOp(\Param)}_2^2-\frac1n\Norm[\big]{\SamplingOp(\Estimator)}_2^2\\
&\geq
\lambda\NormNuc[\big]{\Estimator}-\lambda\NormNuc[\big]{\Param}\\
&\geq
\lambda\NormNuc[\big]{\bP_{\Param}^\perp(\Estimator-\Param)}-\lambda\NormNuc[\big]{\bP_{\Param}(\Estimator-\Param)}.
\end{align*}
By the triangle inequality, we have
\begin{align*}
\NormNuc[\big]{\bP_{\Param}^\perp(\Estimator-\Param)}\leq5\NormNuc[\big]{\bP_{\Param}(\Estimator-\Param)}.
\end{align*}
This finishes proof of \cref{lem:B1}.
\end{proof}

\noindent\cref{lem:compatibility}, the triangle inequality, and \cref{eq:PB-rank} imply that
\begin{align*}
\NormNuc[\big]{\Estimator-\Param}
&\leq
6\NormNuc[\big]{\bP_{\Param}(\Estimator-\Param)}\\
&\leq
\sqrt{72\Rank(\Param)}\NormF[\big]{\bP_{\Param}(\Estimator-\Param)}\\
&\leq
\sqrt{72\Rank(\Param)}\NormF[\big]{\Estimator-\Param}.
\end{align*}
Next, define $b:=\SpikeNorm(\Estimator-\Param)$ and $\bA:=\frac1b(\Estimator-\Param)$. We then have that
\begin{align*}
\SpikeNorm(\bA)=1
\qquad\text{and}\qquad
\NormNuc{\bA}\leq\sqrt{72\Rank(\Param)}\NormF{\bA}.
\end{align*}
In order to finish the proof, we consider the following two cases:

\textit{Case 1:}
If $\NormF{\bA}^2<\nu$, then
\begin{align*}
\NormF[\big]{\Estimator-\Param}^2
<
4\ParamBound^2\nu.
\end{align*}

\textit{Case 2:} Otherwise, $\bA\in\Cset(\nu,\eta)$.\\ 
We can now use the RSC condition and \cref{eq:basic-inequality-3} to get
\begin{align*}
\alpha\frac{\NormF[\big]{\Estimator-\Param}^2}{b^2}-\beta
&\leq
\frac{\Norm[\big]{\SamplingOp(\Estimator-\Param)}_2^2}{nb^2}\,,
\end{align*}
which leads to
\begin{align*}
\alpha\NormF[\big]{\Estimator-\Param}^2-4\ParamBound^2\beta
&\leq
\frac{\Norm[\big]{\SamplingOp(\Estimator-\Param)}_2^2}{n}\\
&\leq
\frac{5\lambda\sqrt{2\Rank(\Param)}}3\NormF[\big]{\Estimator-\Param}\\
&\leq
\frac{50\lambda^2\Rank(\Param)}{3\alpha}+\frac{\alpha}{2}\NormF[\big]{\Estimator-\Param}^2\,.
\end{align*}
Therefore, we have
\begin{align*}
\NormF[\big]{\Estimator-\Param}^2
&\leq
\frac{100\lambda^2\Rank(\Param)}{3\alpha^2}+\frac{8\ParamBound^2\beta}\alpha,
\end{align*}
which completes the proof of \cref{thm:deterministic-error-bound}.
\end{proof}

\subsection{Matrix Bernstein inequality}\label{subsec:bernstein}

The next proposition is a variant of the Bernstein inequality (Proposition 11 of \cite{klopp2014noisy}).
\begin{prop}
\label{lem:freedman}
Let $(\bZ_i)_{i=1}^n$ be a sequence of  $d_r\times d_c$ independent random matrices with zero mean, such that
\begin{align*}
\Expect{\exp\left(\frac{\NormOp{\bZ_i}}{\delta}\right)}\leq e\qquad\text{$\forall i\in[n]$}\,,
\end{align*}
and
\begin{align*}
\sigma_{\bZ}=\max\left\{\NormOp{\frac1n\sum_{i=1}^n\Expect{\bZ_i\bZ_i^\top}},\NormOp{\frac1n\sum_{i=1}^n\Expect{\bZ_i^\top\bZ_i}} \right\}^{\frac12}\,,
\end{align*}
for some positive values $\delta$ and $\sigma_{\bZ}$. Then, there exists a numerical constant $C>0$ such that, for all $t>0$,
\begin{align}
\label{eq:op-norm-tail-bound}
\NormOp{\frac1n\sum_{i=1}^n\bZ_i}
\leq
C\max\left\{
\sigma_{\bZ}\sqrt{\frac{t+\log(d)}{n}},
\delta\left(\log\frac{\delta}{\sigma_{\bZ}}\right)\frac{t+\log(d)}{n}
\right\}\,,
\end{align}
with probability at least $1-\exp(-t)$, where $d=d_r+d_c$.
\end{prop}
We also state the following corollary of the matrix Bernstein inequality.
\begin{cor}
\label{cor:freedman-4-expectation}
If \cref{eq:op-norm-tail-bound} holds and $n\geq \frac{\delta^2}{C\sigma^2_{\bZ}}\log d\left(\log\frac{\delta}{\sigma_{\bZ}}\right)^2$, then
\begin{align*}
\Expect{\NormOp{\frac1n\sum_{i=1}^n\bZ_i}}
\leq
C'\sigma_{\bZ}\sqrt{\frac{2e\log d}{n}},
\end{align*}
where $C'>0$ is a numerical constant.
\end{cor}
This corollary has been proved for the case of $\bZ=\zeta_i\bX_i$ in \cite{klopp2014noisy}. The proof can be adapted to the general case as well.

\subsection{Proof of \cref{thm:rsc-condition}}\label{subsec:pf-rsc-condition}

\begin{proof}
First, we reproduce a slightly modified version of proof Lemma 12 in \citep{klopp2014noisy}, adapted to our setting. Set
\[
\beta:=\frac{93\eta\Cutoff^2}{\gmin}\Expect[\big]{\NormOp[\big]{\bSigma_R}}^2\,.
\]
By $\Bset$, we denote a bad event defined as
\begin{align*}
\Bset:=\left\{\exists\bA\in\Cset'(\theta,\eta)\text{ such that } \frac12\Norm{\bA}_{L^2(\SampleDistr)}^2-\frac1n\Norm{\SamplingOp(\bA)}_2^2>\frac14\Norm{\bA}_{L^2(\SampleDistr)}^2+\beta \right\}.
\end{align*}
We thus need to bound the probability of this event. Set $\xi=6/5$. Then, for $T>0$, we define
\begin{align*}
\Cset'(\theta,\eta,T):=\{\bA\in\Cset'(\theta,\eta)\Given T\leq\Norm{\bA}_{L^2(\SampleDistr)}^2<\xi T \}.
\end{align*}
Clearly, we have
\begin{align*}
\Cset'(\theta,\theta)=\bigcup_{l=1}^\infty\Cset'(\theta,\eta,\xi^{l-1}\theta).
\end{align*}
Now, if the event $\Bset$ holds for some $\bA\in\Cset'(\theta,\eta)$, then $\bA\in\Cset'(\theta,\eta,\xi^{l-1}\theta)$ for some $l\in\IN$. In this case, we have
\begin{align*}
\frac12\Norm{\bA}_{L^2(\SampleDistr)}^2-\frac1n\Norm{\SamplingOp(\bA)}_2^2
&>
\frac14\Norm{\bA}_{L^2(\SampleDistr)}^2+\beta\\
&\geq
\frac14\xi^{l-1}\theta+\beta\\
&=
\frac5{24}\xi^{l}\theta+\beta.
\end{align*}
Next, we define the event $\Bset_l$ as
\begin{align*}
\Bset_l:=\left\{
\exists\bA\in\Cset'(\theta,\eta,\xi^{l-1}\theta)\text{ such that } \frac12\Norm{\bA}_{L^2(\SampleDistr)}^2-\frac1n\Norm{\SamplingOp(\bA)}_2^2
>
\frac5{24}\xi^{l}\theta+\beta
\right\}.
\end{align*}
It follows that
\begin{align*}
\Bset\subseteq\bigcup_{l=1}^\infty\Bset_l.
\end{align*}
The following lemma helps us obtain an upper bound for the probability of each of these events $\Bset_l$.
\begin{lem}
\label{lem:bound-Bset-l}
Define
\begin{align*}
Z_T:=\sup_{\bA\in\Cset'(\theta,\eta,T)}\left\{\frac12\Norm{\bA}_{L^2(\SampleDistr)}^2-\frac1n\Norm{\SamplingOp(\bA)}_2^2\right\}\,.
\end{align*}
Then, assuming that $(\Xtilde_i)_{i=1}^n$ are i.i.d. samples drawn from $\SampleDistr$, we get
\begin{align}
\Prob{Z_T\geq \frac{5\xi T}{24} + \beta}
&\leq
\exp\left(-\frac{Cn\xi T}{\Cutoff^2} \right)\,,\label{eq:Z_T-tail-bound}
\end{align}
for some numerical constant $C>0$.
\end{lem}
\begin{proof}
The proof is similar to the proof of Lemma 14 in \citep{klopp2014noisy}. For a $d_r\times d_c$ matrix $\bA$, define
\begin{align*}
f(\bX;\bA):=\Inner{\bX,\bA}^2\cdot\II\left(\Abs{\Inner{\bX,\bA}}\leq \Cutoff{}\right).
\end{align*}
Next, denote
\begin{align*}
W_T:=
\sup_{\bA\in\Cset'(\theta,\eta,T)}
\frac1n\sum_{i=1}^{n}\left\{
\Expect{f(\bX_i;\bA)}
-
f(\bX_i;\bA)
\right\}\,,\\
\tW_T:=
\sup_{\bA\in\Cset'(\theta,\eta,T)}
\Big|
\frac1n\sum_{i=1}^{n}\left\{
\Expect{f(\bX_i;\bA)}
-
f(\bX_i;\bA)
\right\}\Big|\,.
\end{align*}
It follows from \cref{as:x-tilde-boundedness-stochastic} (where $\Cutoff{}$ is defined) that $Z_T\leq W_T$, and clearly $W_T\leq \tW_T$.  Therefore, we have
\begin{align*}
\Prob{Z_T\geq t} \leq \Prob{\tW_T\geq t}
\end{align*}
for all $t$. Therefore, if we prove that \cref{eq:Z_T-tail-bound} holds when $Z_T$ is replaced with $\tW_T$, we will be done. In the sequel, we will aim to prove this using Massart's inequality (e.g., Theorem 3 of \cite{massart2000constants}). In order to invoke Massart's inequality, we need bounds for $\Expect{\tW_T}$ and $\Var{\tW_T}$.

First, we find an upper bound for $\Expect{\tW_T}$. It follows from the symmetrization argument (e.g., Lemma 6.3 of \cite{ledoux2013probability}) that
\begin{align}
\Expect{\tW_T}
&\leq
2\,\Expect{\sup_{\bA\in\Cset'(\theta,\eta,T)}
\Big|\frac1n\sum_{i=1}^n\zeta_if(\bX_i;\bA)\Big|}\,,\label{eq:WT-upper1}
\end{align}
where $(\zeta_i)_{i=1}^n$ is a sequence of i.i.d. Radamacher random variables. Note that Lemma 6.3 of \cite{ledoux2013probability} requires the use of a convex function and a norm. Here, the convex function is the identity function and the norm is the infinity norm applied to an infinite-dimensional vector (indexed by $\bA\in\Cset'(\theta,\eta,T)$).

Next, we will use the contraction inequality (e.g., Theorem 4.4 of \cite{ledoux2013probability}). First, we write $f(\bX_i;\bA)=\alpha_i\Inner{\bX_i,\bA}$, where $\alpha_i=\Inner{\bX_i,\bA}\cdot\II\left(\Abs{\Inner{\bX_i,\bA}}\leq \Cutoff{}\right)$. By definition, $|\alpha_i|\leq\Cutoff{}$. Now, for every realization of the random variables $\bX_1,\ldots,\bX_n$ we can apply Theorem 4.4 in \cite{ledoux2013probability} to obtain
\begin{align*}
\Expect*{\zeta}{\sup_{\bA\in\Cset'(\theta,\eta,T)}\Big |\frac1n\sum_{i=1}^n\zeta_if(\bX_i;\bA)\Big | } &\leq
\Cutoff{}\,\Expect*{\zeta}{\sup_{\bA\in\Cset'(\theta,\eta,T)}\Big |\frac1n\sum_{i=1}^n\zeta_i\Inner{\bX_i,\bA}\Big | } \,.
\end{align*}
Now, taking expectation of both sides with respect to $\bX_i$'s, using the tower property, and combining with \cref{eq:WT-upper1}, we obtain
\begin{align*}
\Expect{\tW_T}
&\leq
8\Cutoff{}\Expect{\sup_{\bA\in\Cset'(\theta,\eta,T)}\Big |\frac1n\sum_{i=1}^n\zeta_i\Inner{\bX_i,\bA}\Big |}\\
&\leq
8\Cutoff{}\Expect{\NormOp[\big]{\bSigma_R}\sup_{\bA\in\Cset'(\theta,\eta,T)}\NormNuc{\bA}}\\
&\leq
8\Cutoff{}\sqrt{\eta}\,\Expect{\NormOp[\big]{\bSigma_R}\sup_{\bA\in\Cset'(\theta,\eta,T)}\NormF{\bA}}\\
&\leq
8\Cutoff{}\sqrt{\frac\eta\gmin}\Expect{\NormOp[\big]{\bSigma_R}\sup_{\bA\in\Cset'(\theta,\eta,T)}\Norm{\bA}_{L^2(\SampleDistr)}}\\
&\leq
8\Cutoff{}\sqrt{\frac{\eta\xi T}\gmin}\Expect{\NormOp[\big]{\bSigma_R}}.
\end{align*}
In the above, we also used the definition of $\Cset'(\theta,\eta,T)$ as well as \cref{as:gtmin-gtmax-bounds}. We can now use $2ab\leq a^2+b^2$ to get
\begin{align*}
\Expect{\tW_T}
&\leq
\frac{8}{9}\left(\frac{5\xi T}{24}\right) +\frac{87\eta \Cutoff{}^2}{\gmin}\Expect{\NormOp[\big]{\bSigma_R}}^2.
\end{align*}
Next, we turn to find an upper bound for a certain variance term, specifically, the term
$
\sum_{i=1}^n \Var{f(\bX_i;\bA)/n}
$. Since
\begin{align*}
\Var{f(\bX_i;\bA)}
&\leq
\Expect{f(\bX_i;\bA)^2}\\
&\leq
\Cutoff{}^2\cdot\Expect{\Inner{\bX_i,\bA}^2}\\
&=
\Cutoff{}^2\cdot\Norm{\bA}_{L^2(\SampleDistr)}^2\,,
\end{align*}
it follows that
\begin{align*}
\sup_{\bA\in\Cset'(\theta,\eta,T)}\sum_{i\in[n]}\frac{1}{n^2}\Var{f(\bX_i;\bA)}
&\leq
\frac{\Cutoff{}^2}n\cdot\sup_{\bA\in\Cset'(\theta,\eta,T)}\Norm{\bA}_{L^2(\SampleDistr)}^2\\
&\leq
\frac{\xi T\Cutoff{}^2}n.
\end{align*}

Finally, noting that $\frac1nf(\bX_i;\bA)\leq\frac1n\Cutoff{}^2$ almost surely, we can use Massart's inequality (e.g., Theorem 3 of \cite{massart2000constants}) to conclude that
\begin{align*}
\Prob{\tW_T\geq \frac{5\xi T}{24} + \beta} & =
\Prob{\tW_T\geq \frac{5\xi T}{24} + \frac{93\eta \Cutoff{}^2}{\gmin}\Expect{\NormOp[\big]{\bSigma_R}}^2}\\
&\leq
\Prob{\tW_T\geq \frac{18}{17}\Expect{\tW_T}+\frac{1}{17}\left(\frac{5\xi T}{24}\right)}\\
&\leq
\exp\left(-\frac{Cn\xi T}{\Cutoff{}^2} \right)
\end{align*}
for some numerical constant $C_0>0$.
\end{proof}\\
\cref{lem:bound-Bset-l} entails that
\begin{align*}
\Prob{\Bset_l}
&\leq
\exp\left(-\frac{C_0n\xi^{l}\theta}{\Cutoff^2} \right)\\
&\leq
\exp\left(-\frac{C_0nl\log(\xi)\theta}{\Cutoff^2} \right)\,.
\end{align*}
Therefore, by setting the numerical constant $C>0$ appropriately, the union bound implies that
\begin{align*}
\Prob{\Bset}
&\leq
\sum_{l=1}^{\infty}\Prob{\Bset_l}\\
&\leq
\sum_{l=1}^{\infty}\exp\left(-\frac{Cnl\theta}{\Cutoff^2}\right)\\
&=
\frac{\exp\left(-\frac{Cn\theta}{\Cutoff^2}\right)}{1-\exp\left(-\frac{Cn\theta}{\Cutoff^2}\right)}.
\end{align*}
Finally, assuming that $Cn\theta>\Cutoff^2$, we get that
\begin{align*}
\Prob{\Bset}
&\leq
2\exp\left(-\frac{Cn\theta}{\Cutoff^2}\right),
\end{align*}
which completes the proof of \cref{thm:rsc-condition}.
\end{proof}

\section{Additional simulations}
\label{sec:supp-simulations}

\begin{figure}
	\centering
	\begin{subfigure}[b]{0.49\textwidth}
		\centering
		\includegraphics[width=\textwidth]{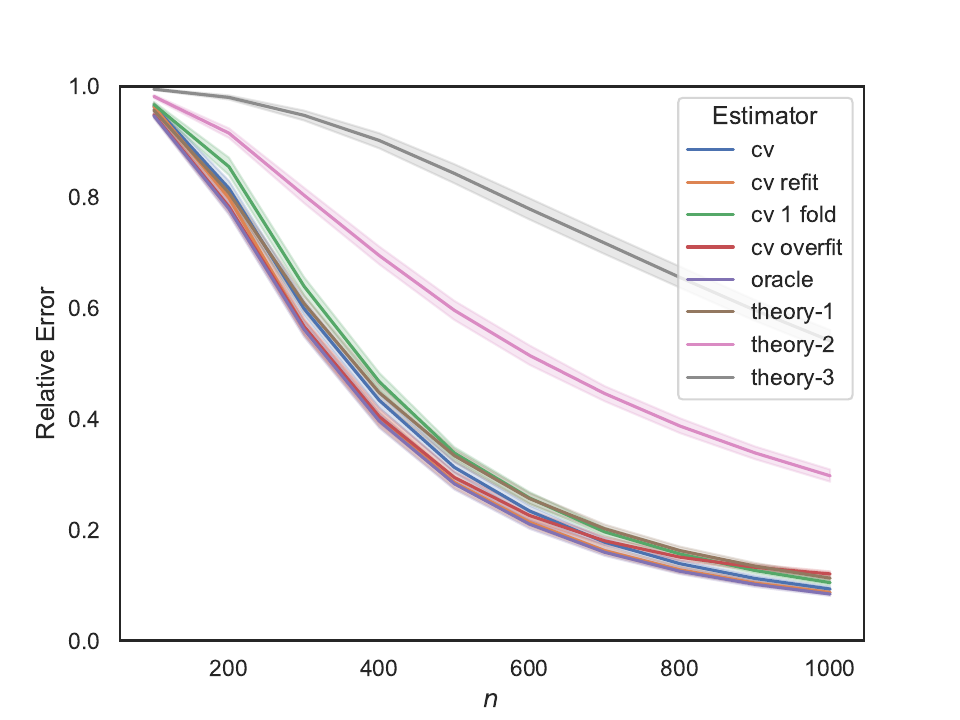}
		\caption{Synthetic, $(d,r)=(50,2)$, $\sigma=0.5$}
	\end{subfigure}
	\hfill
	\begin{subfigure}[b]{0.49\textwidth}
		\centering
		\includegraphics[width=\textwidth]{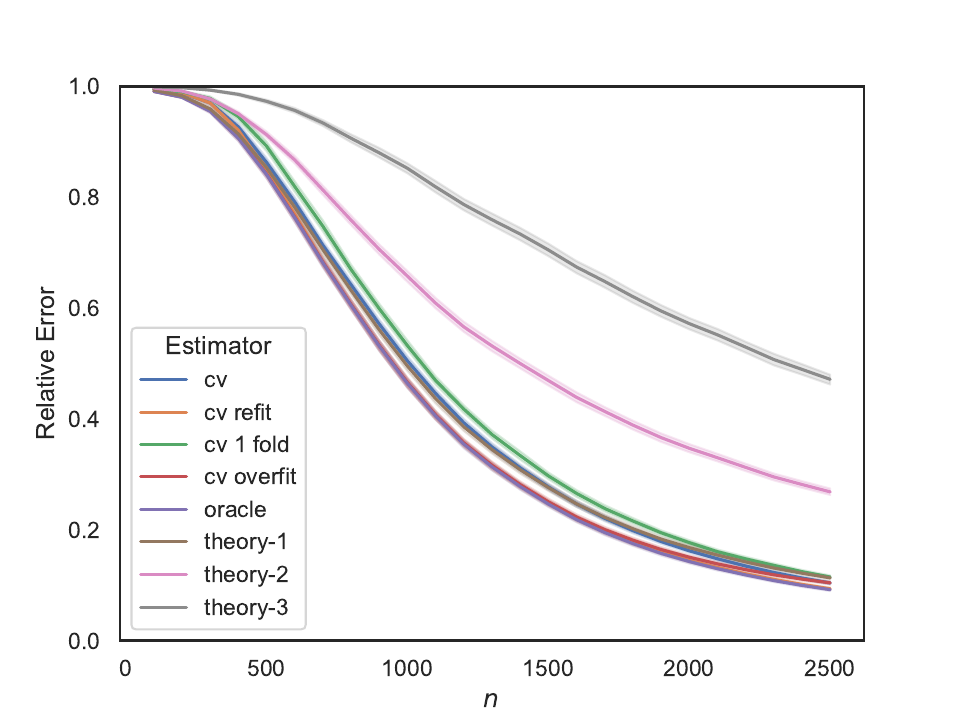}
		\caption{Synthetic, $(d,r)=(100,3)$, $\sigma=0.5$}
	\end{subfigure}\\
	\begin{subfigure}[b]{0.49\textwidth}
		\centering
		\includegraphics[width=\textwidth]{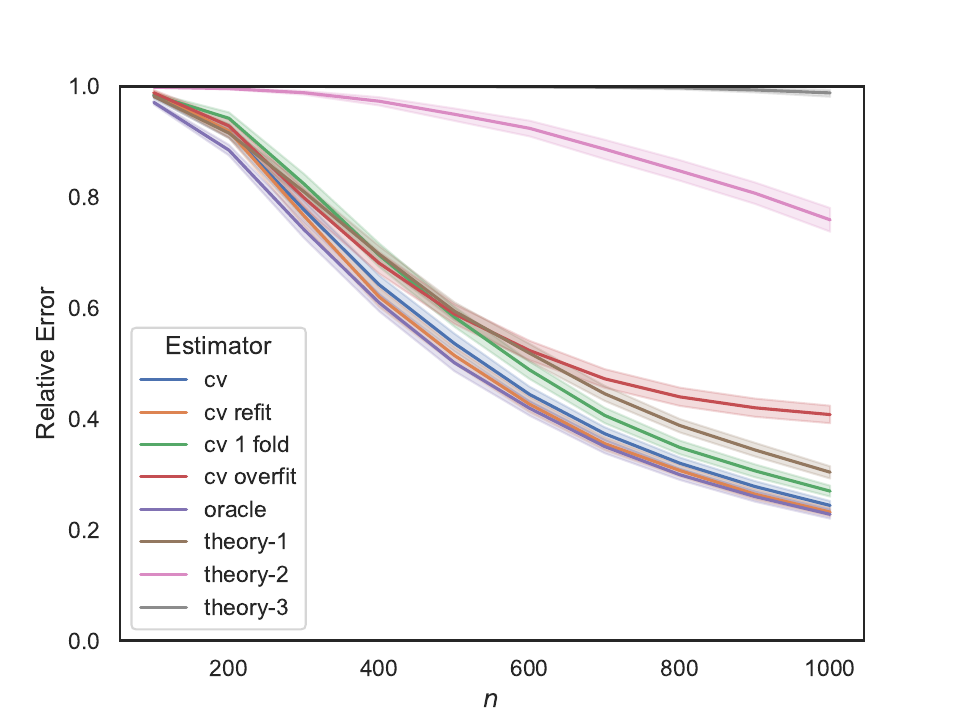}
		\caption{Synthetic, $(d,r)=(50,2)$, $\sigma=1$}
	\end{subfigure}
	\hfill
	\begin{subfigure}[b]{0.49\textwidth}
		\centering
		\includegraphics[width=\textwidth]{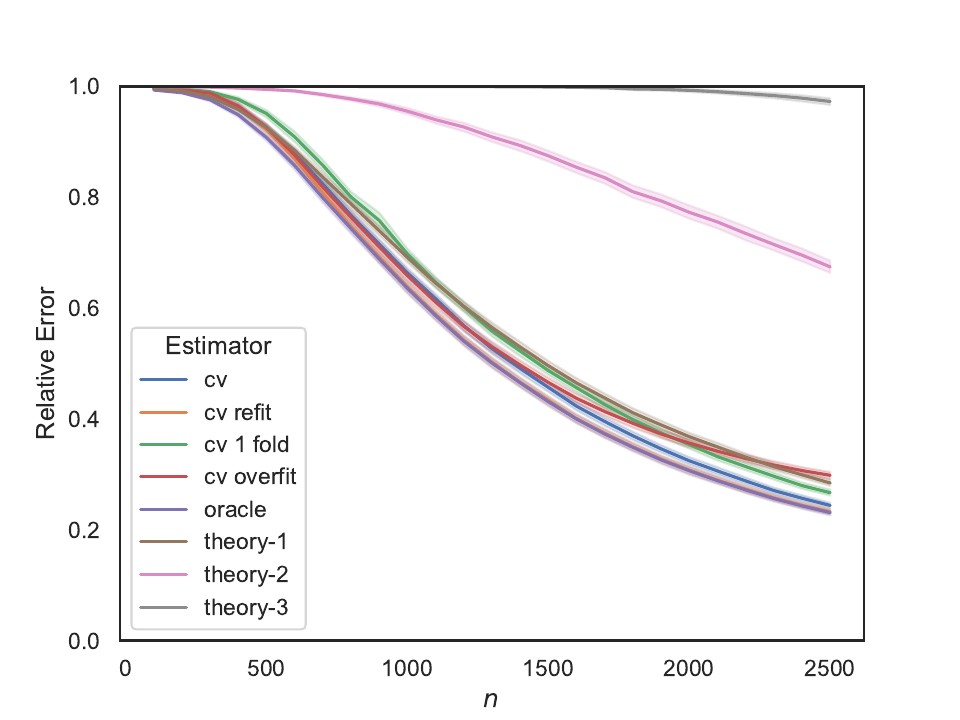}
		\caption{Synthetic, $(d,r)=(100,3)$, $\sigma=1$}
	\end{subfigure}\\
	\begin{subfigure}[b]{0.49\textwidth}
	\centering
	\includegraphics[width=\textwidth]{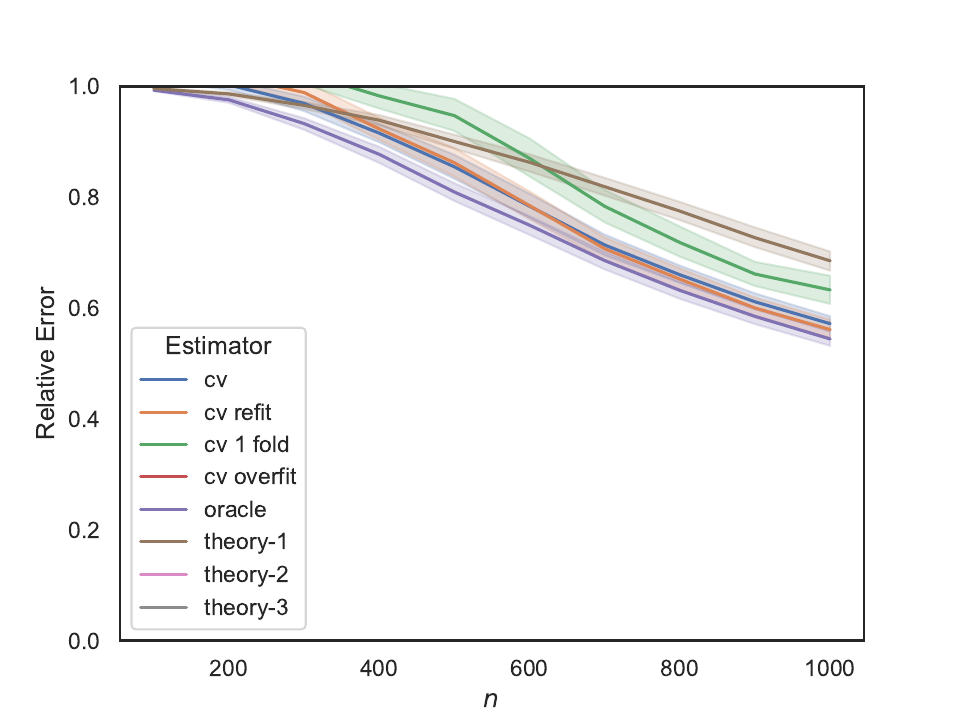}
	\caption{Synthetic, $(d,r)=(50,2)$, $\sigma=2$}
\end{subfigure}
\hfill
\begin{subfigure}[b]{0.49\textwidth}
	\centering
	\includegraphics[width=\textwidth]{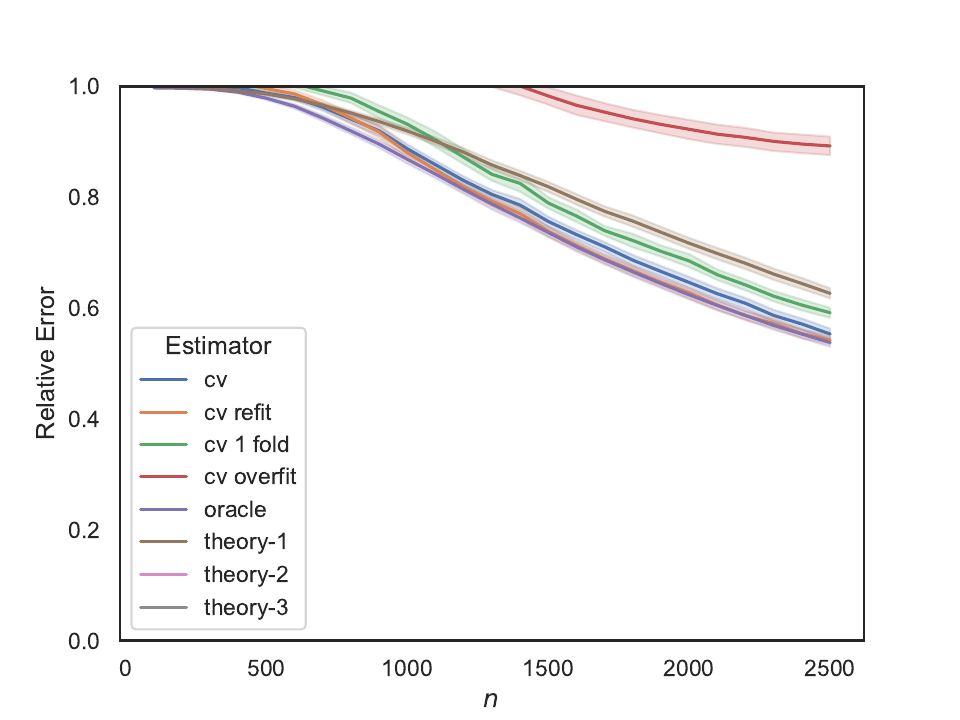}
	\caption{Synthetic, $(d,r)=(100,3)$, $\sigma=2$}
\end{subfigure}
	\caption{Comparison of the relative error (i.e., $\NormF[\big]{\Estimator-\Param}^2/\NormF{\Param}^2$) for the proposed estimators on synthetic data, for varying noise level and cross-validation methodology.}
	\label{fig:rel-error-full}
\end{figure}

\begin{figure}
	\centering
	\begin{subfigure}[b]{0.49\textwidth}
		\centering
		\includegraphics[width=\textwidth]{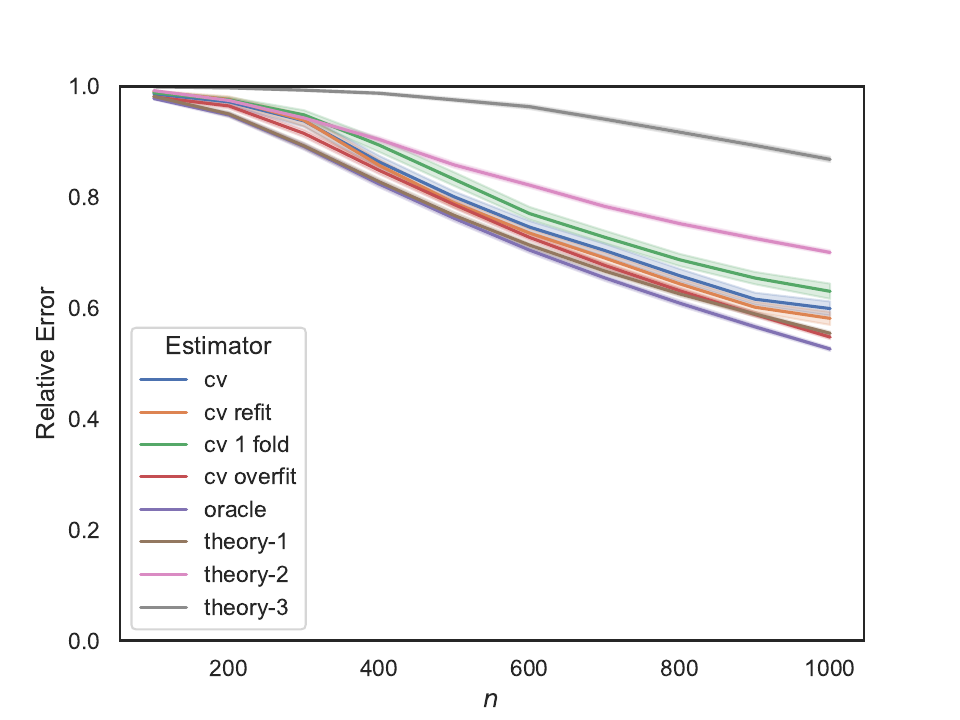}
		\caption{Movielens, $d=50$, $\sigma=0.5$}
	\end{subfigure}
	\hfill
	\begin{subfigure}[b]{0.49\textwidth}
		\centering
		\includegraphics[width=\textwidth]{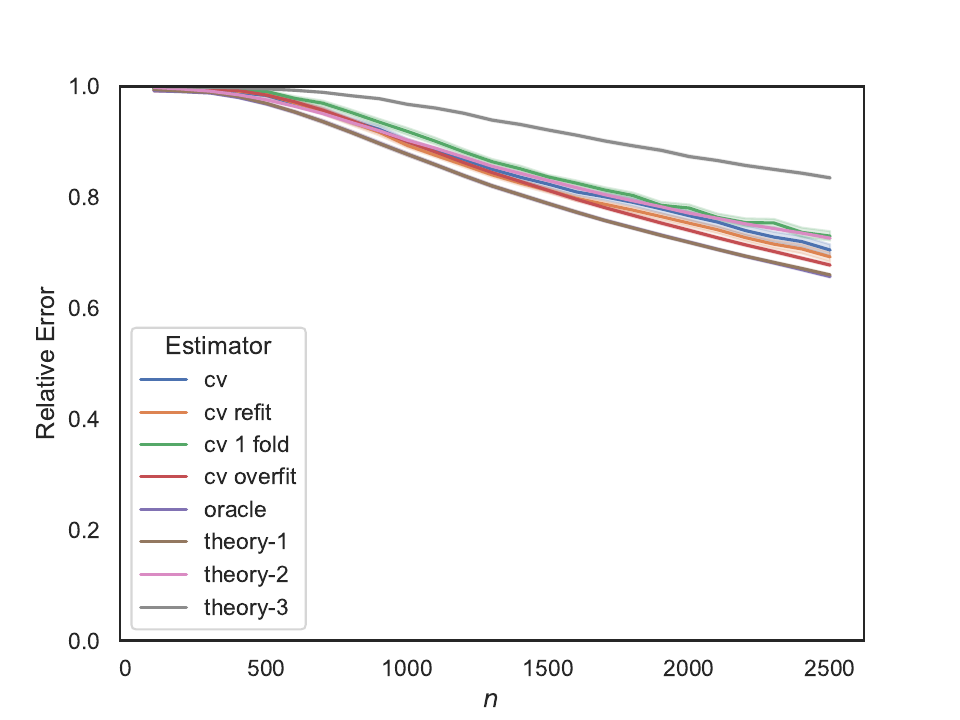}
		\caption{Movielens,  $d=100$, $\sigma=0.5$}
	\end{subfigure}\\
	\begin{subfigure}[b]{0.49\textwidth}
		\centering
		\includegraphics[width=\textwidth]{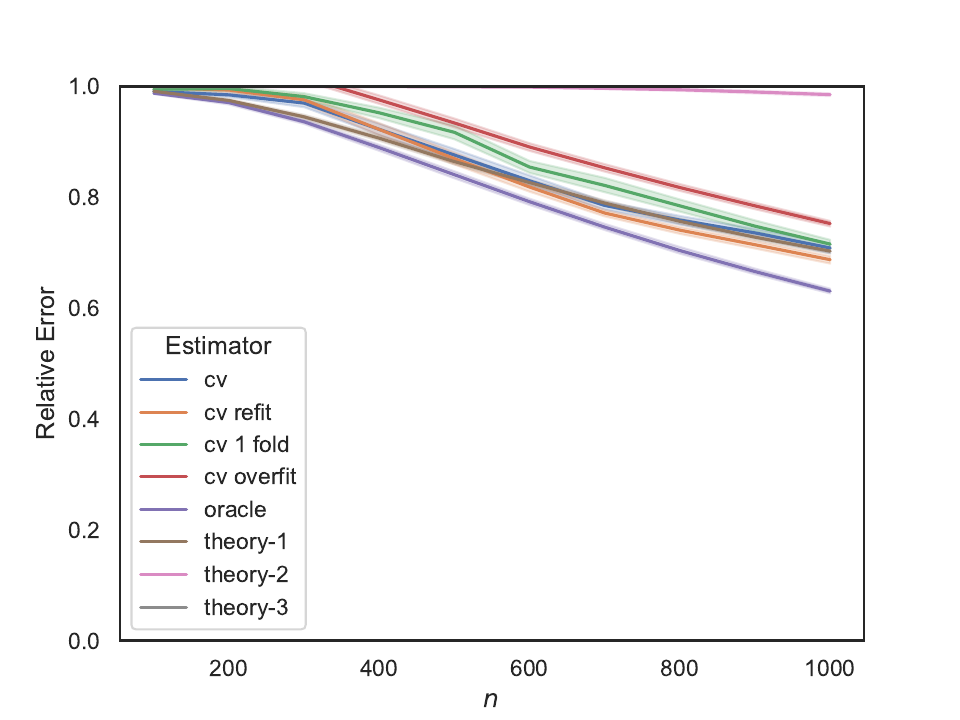}
		\caption{Movielens, $d=50$, $\sigma=1$}
	\end{subfigure}
	\hfill
	\begin{subfigure}[b]{0.49\textwidth}
		\centering
		\includegraphics[width=\textwidth]{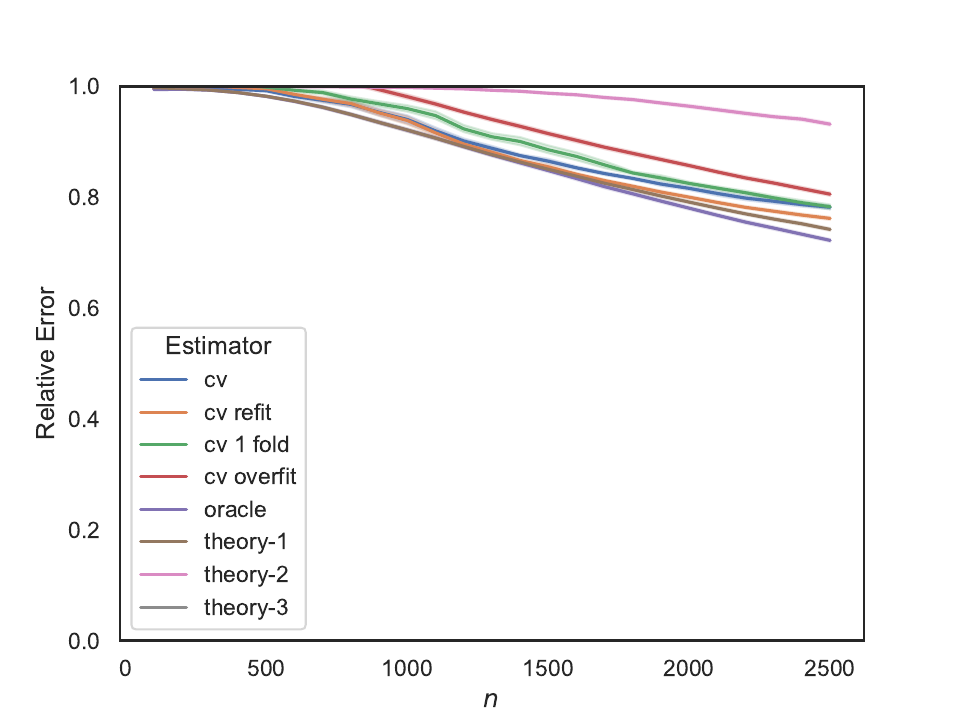}
		\caption{Movielens, $d=100$, $\sigma=1$}
	\end{subfigure}\\
	\begin{subfigure}[b]{0.49\textwidth}
		\centering
		\includegraphics[width=\textwidth]{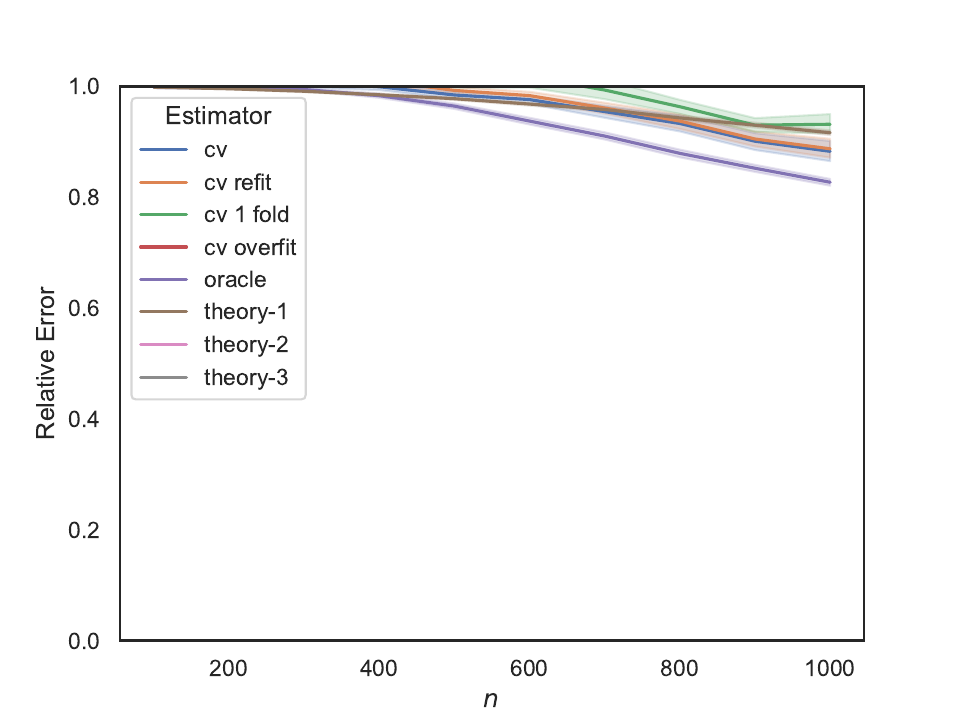}
		\caption{Movielens, $d=50$, $\sigma=2$}
	\end{subfigure}
	\hfill
	\begin{subfigure}[b]{0.49\textwidth}
		\centering
		\includegraphics[width=\textwidth]{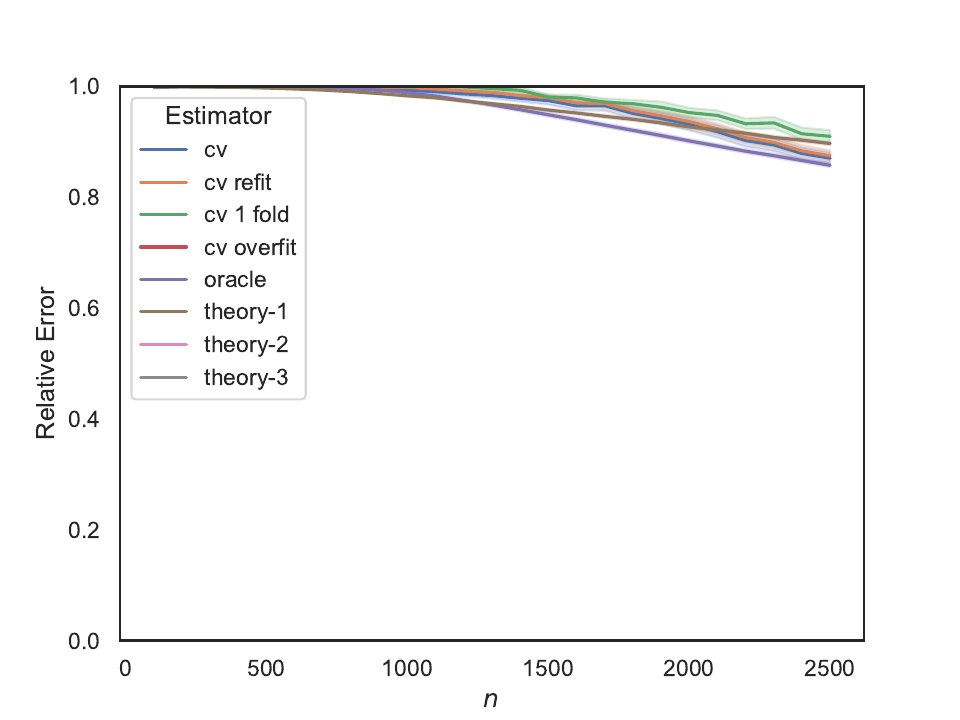}
		\caption{MovieLens, $d=100$, $\sigma=2$}
	\end{subfigure}
	\caption{Comparison of the relative error (i.e., $\NormF[\big]{\Estimator-\Param}^2/\NormF{\Param}^2$) for the proposed estimators on movielens data, for varying noise level and cross-validation methodology.}
	\label{fig:rel-error-full-real-data}
\end{figure}

\begin{figure}
	\centering
	\begin{subfigure}[b]{0.49\textwidth}
		\centering
		\includegraphics[width=\textwidth]{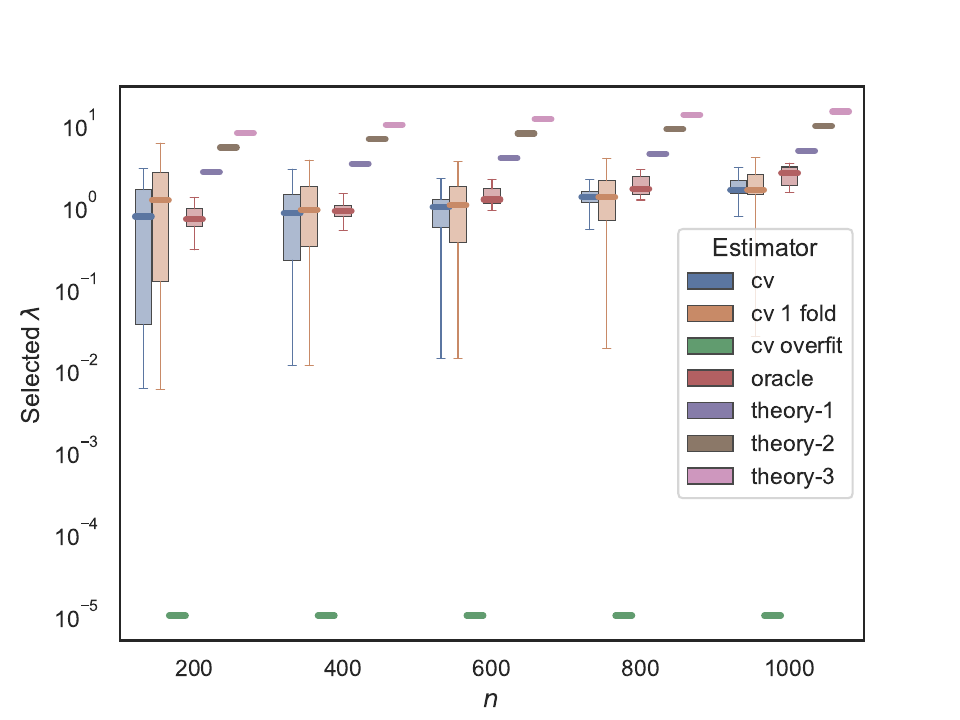}
		\caption{Synthetic, $(d,r)=(50,2)$, $\sigma=0.5$}
	\end{subfigure}
	\hfill
	\begin{subfigure}[b]{0.49\textwidth}
		\centering
		\includegraphics[width=\textwidth]{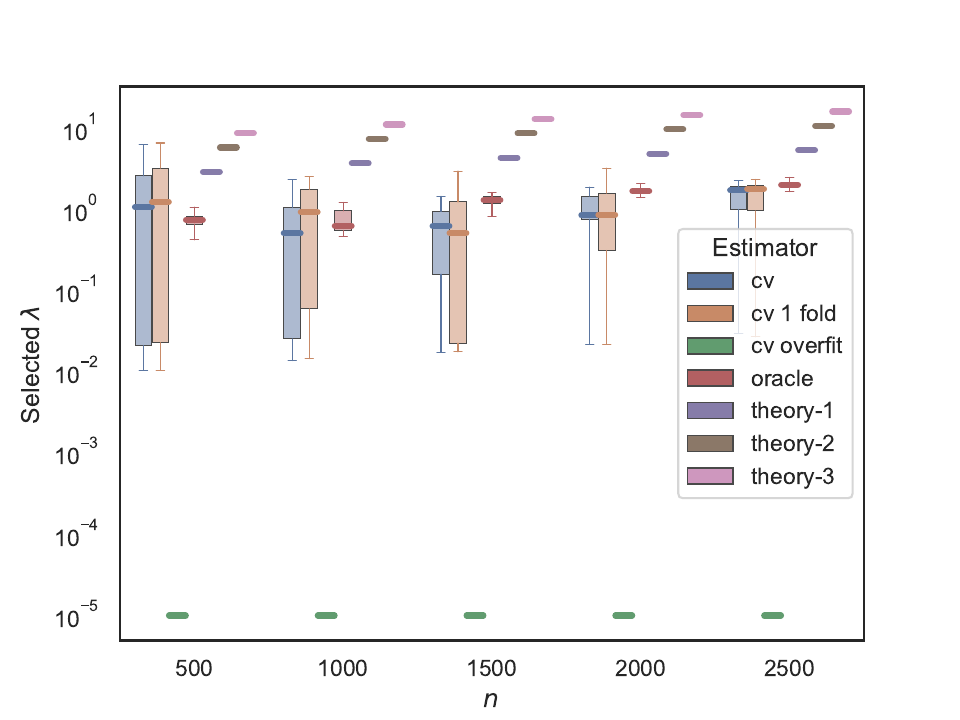}
		\caption{Synthetic, $(d,r)=(100,3)$, $\sigma=0.5$}
	\end{subfigure}\\
	\begin{subfigure}[b]{0.49\textwidth}
		\centering
		\includegraphics[width=\textwidth]{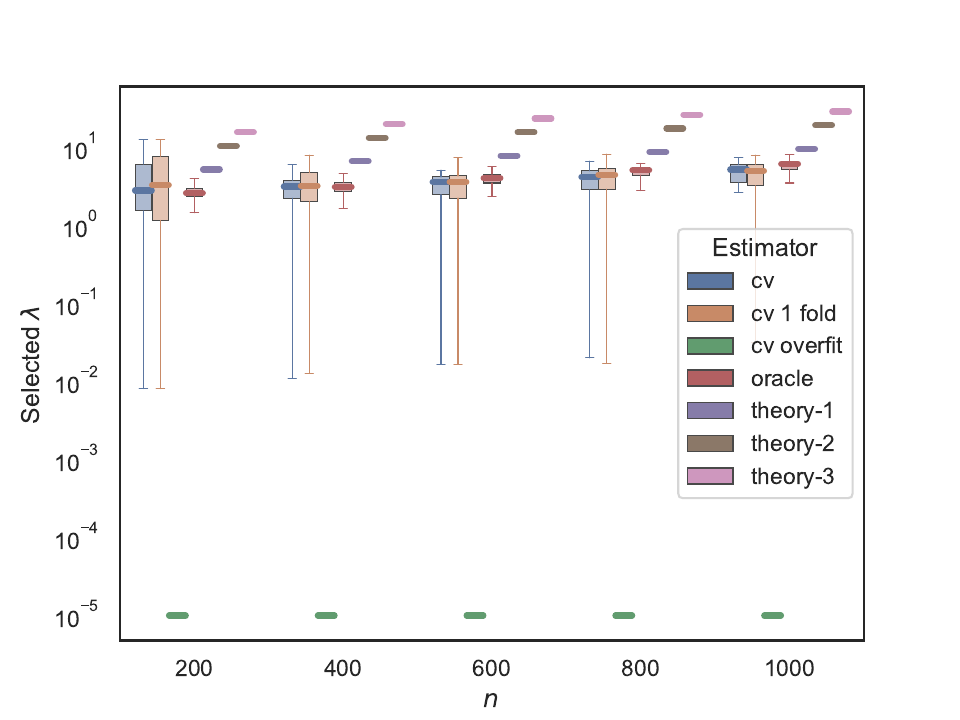}
		\caption{Synthetic, $(d,r)=(50,2)$, $\sigma=1$}
	\end{subfigure}
	\hfill
	\begin{subfigure}[b]{0.49\textwidth}
		\centering
		\includegraphics[width=\textwidth]{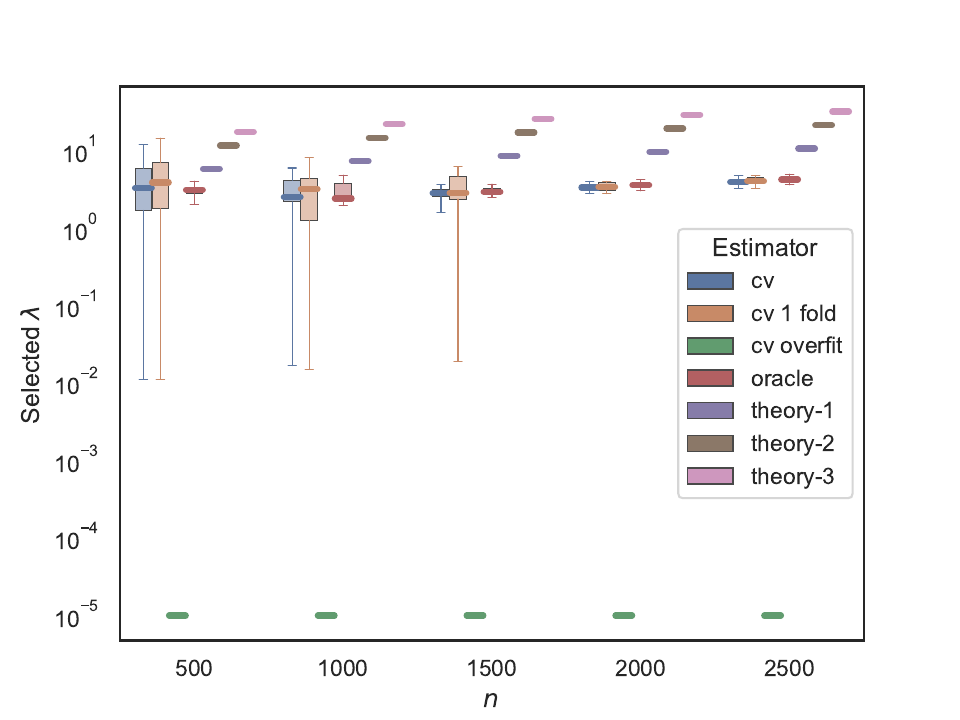}
		\caption{Synthetic, $(d,r)=(100,3)$, $\sigma=1$}
	\end{subfigure}\\
	\begin{subfigure}[b]{0.49\textwidth}
		\centering
		\includegraphics[width=\textwidth]{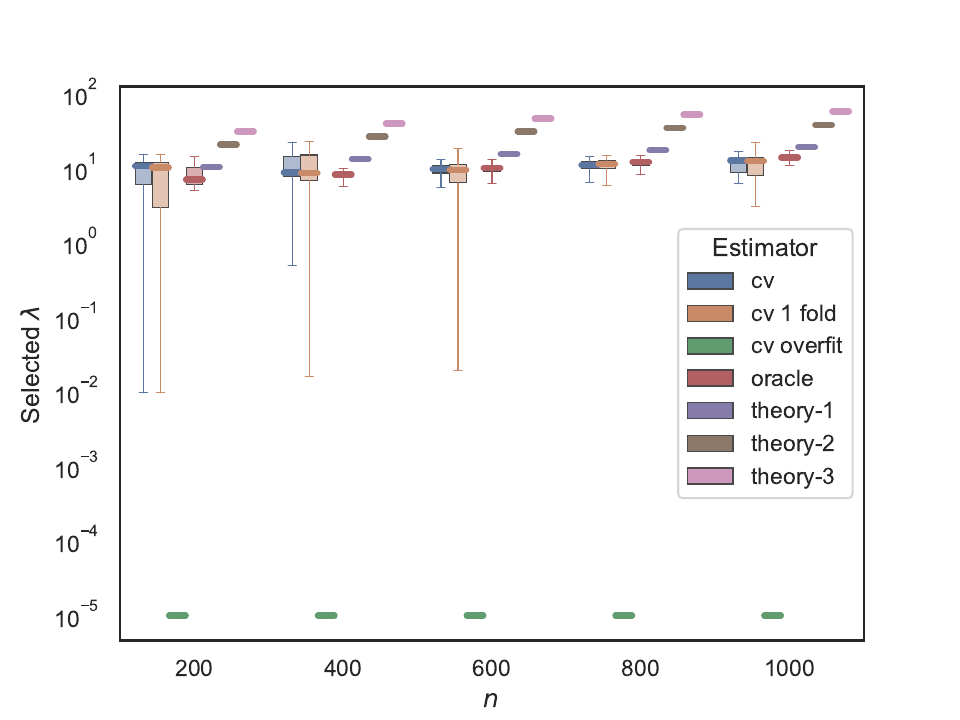}
		\caption{Synthetic, $(d,r)=(50,2)$, $\sigma=2$}
	\end{subfigure}
	\hfill
	\begin{subfigure}[b]{0.49\textwidth}
		\centering
		\includegraphics[width=\textwidth]{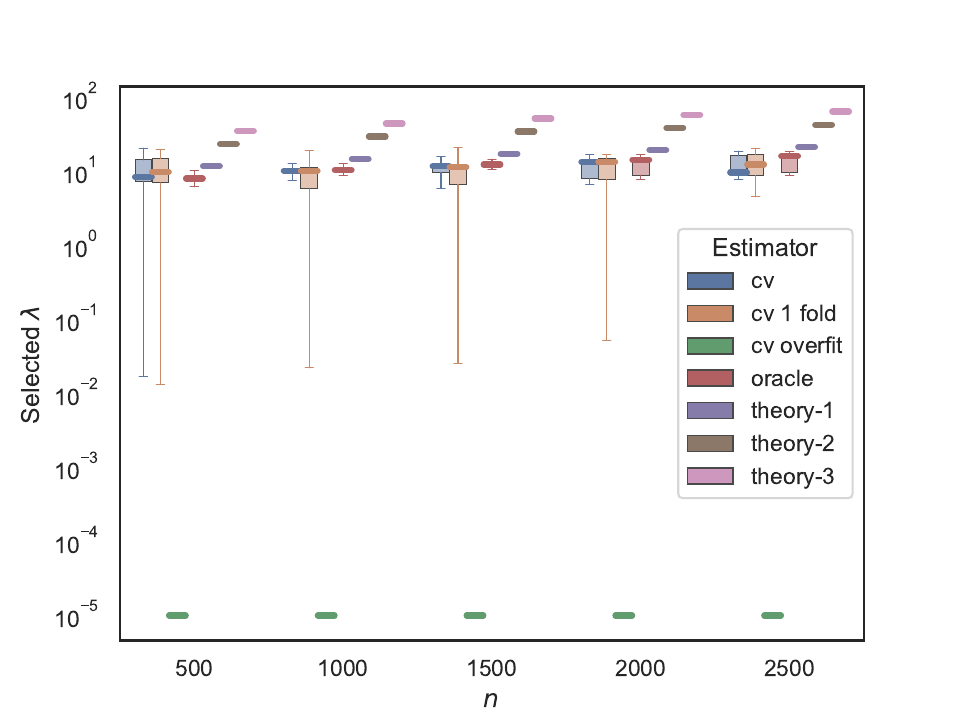}
		\caption{Synthetic, $(d,r)=(100,3)$, $\sigma=2$}
	\end{subfigure}
	\caption{Comparison of the selected penalty parameter $\lambda$ for the proposed estimators on synthetic data, for varying noise level and cross-validation methodology}
	\label{fig:lam-sel-full}
\end{figure}

\begin{figure}[H]
	\centering
	\begin{subfigure}[b]{0.49\textwidth}
		\centering
		\includegraphics[width=\textwidth]{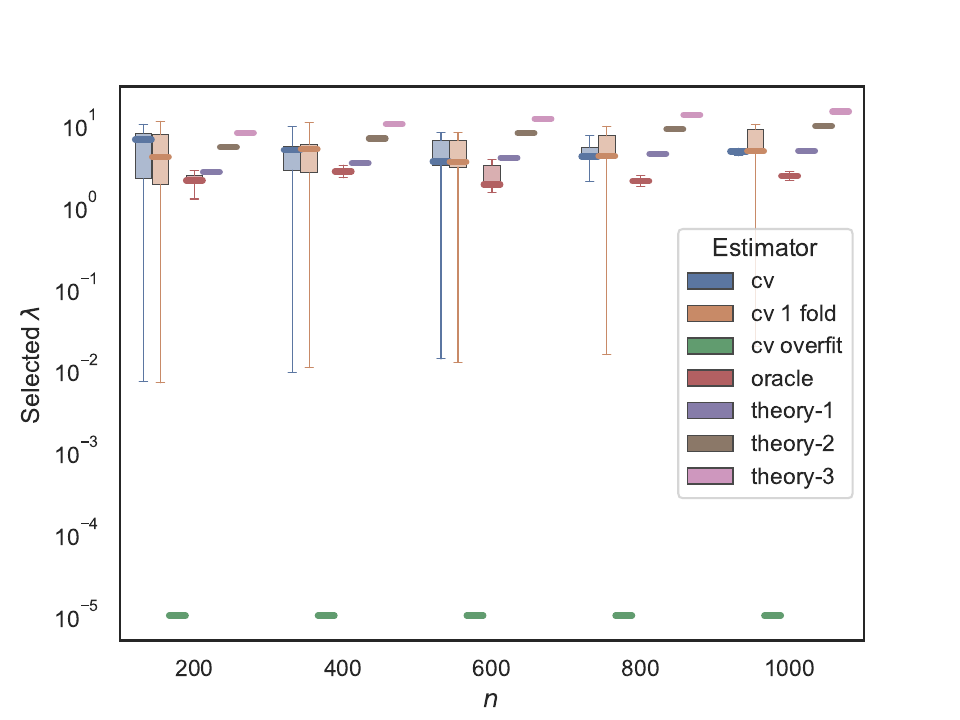}
		\caption{Movielens, $d=50$, $\sigma=0.5$}
	\end{subfigure}
	\hfill
	\begin{subfigure}[b]{0.49\textwidth}
		\centering
		\includegraphics[width=\textwidth]{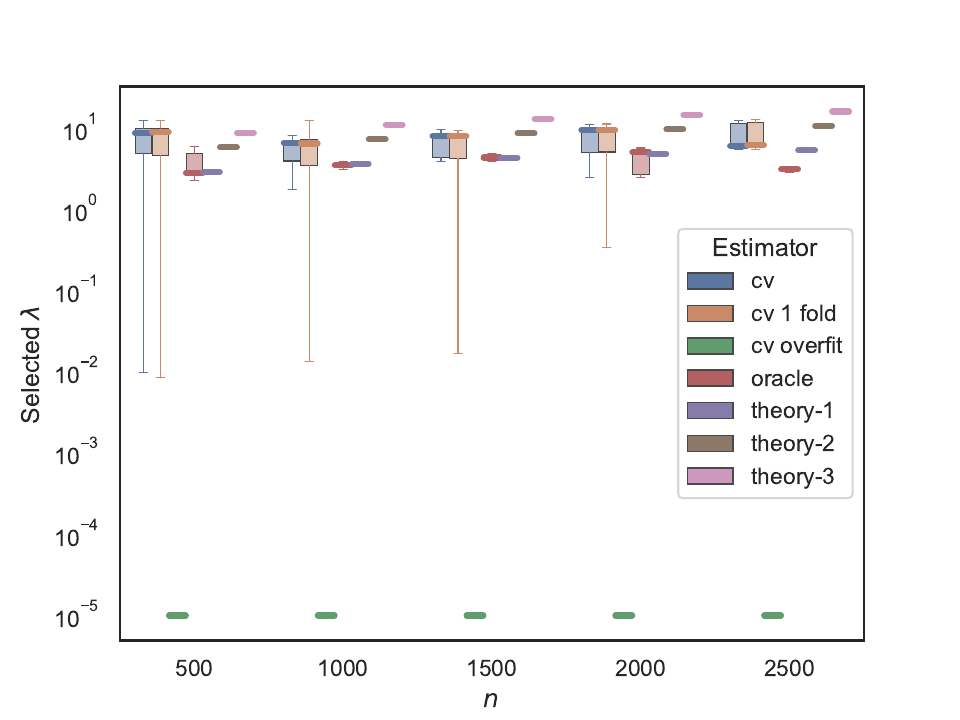}
		\caption{Movielens,  $d=100$, $\sigma=0.5$}
	\end{subfigure}\\
	\begin{subfigure}[b]{0.49\textwidth}
		\centering
		\includegraphics[width=\textwidth]{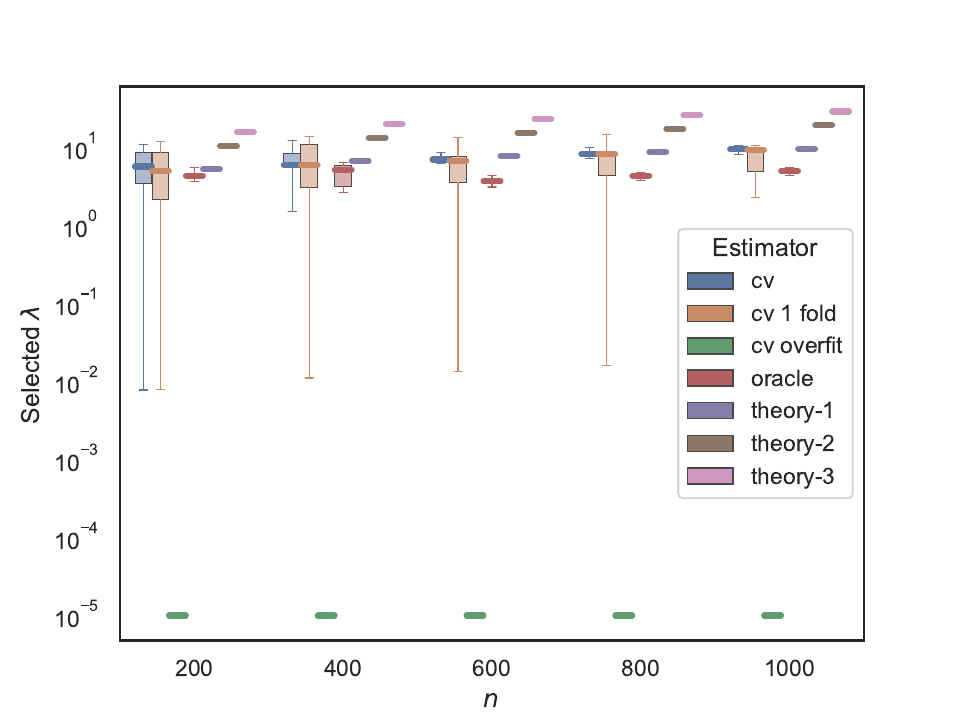}
		\caption{Movielens, $d=50$, $\sigma=1$}
	\end{subfigure}
	\hfill
	\begin{subfigure}[b]{0.49\textwidth}
		\centering
		\includegraphics[width=\textwidth]{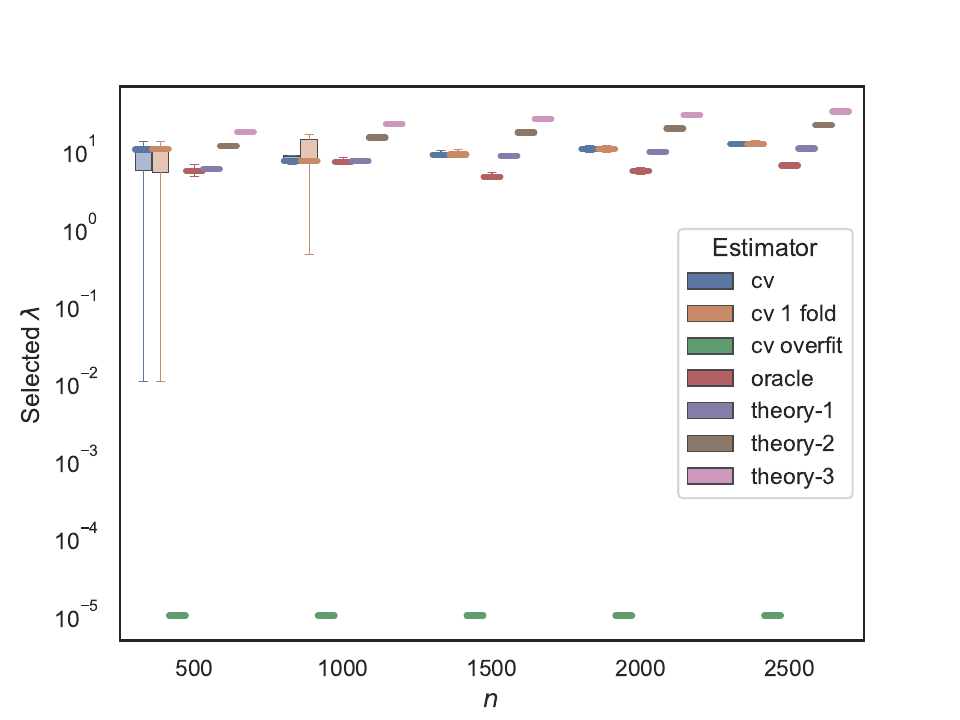}
		\caption{Movielens, $d=100$, $\sigma=1$}
	\end{subfigure}\\
	\begin{subfigure}[b]{0.49\textwidth}
		\centering
		\includegraphics[width=\textwidth]{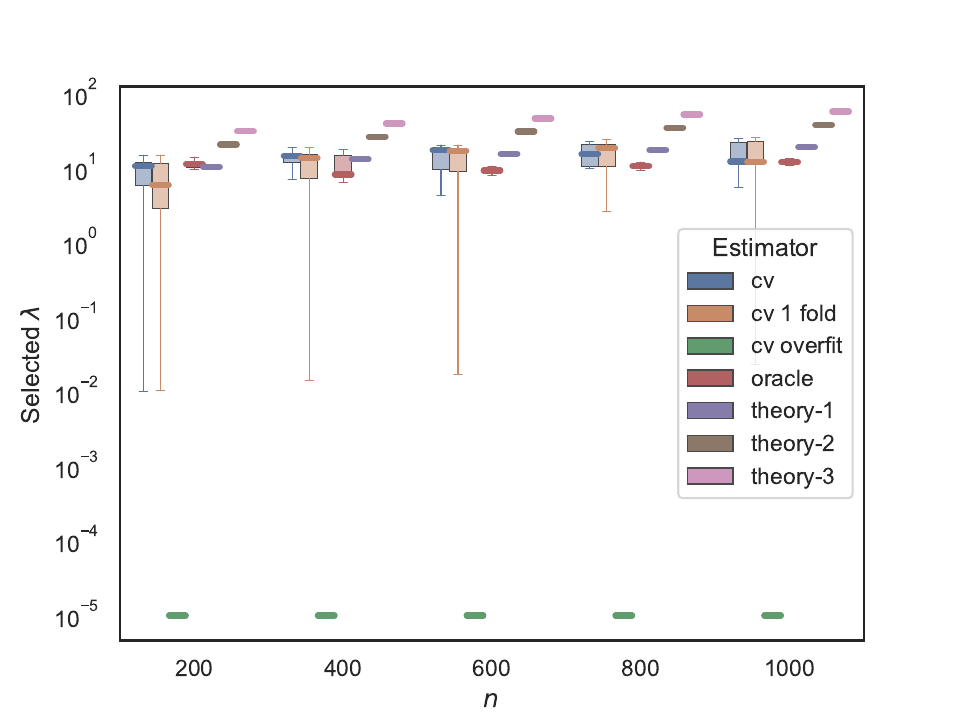}
		\caption{Movielens, $d=50$, $\sigma=2$}
	\end{subfigure}
	\hfill
	\begin{subfigure}[b]{0.49\textwidth}
		\centering
		\includegraphics[width=\textwidth]{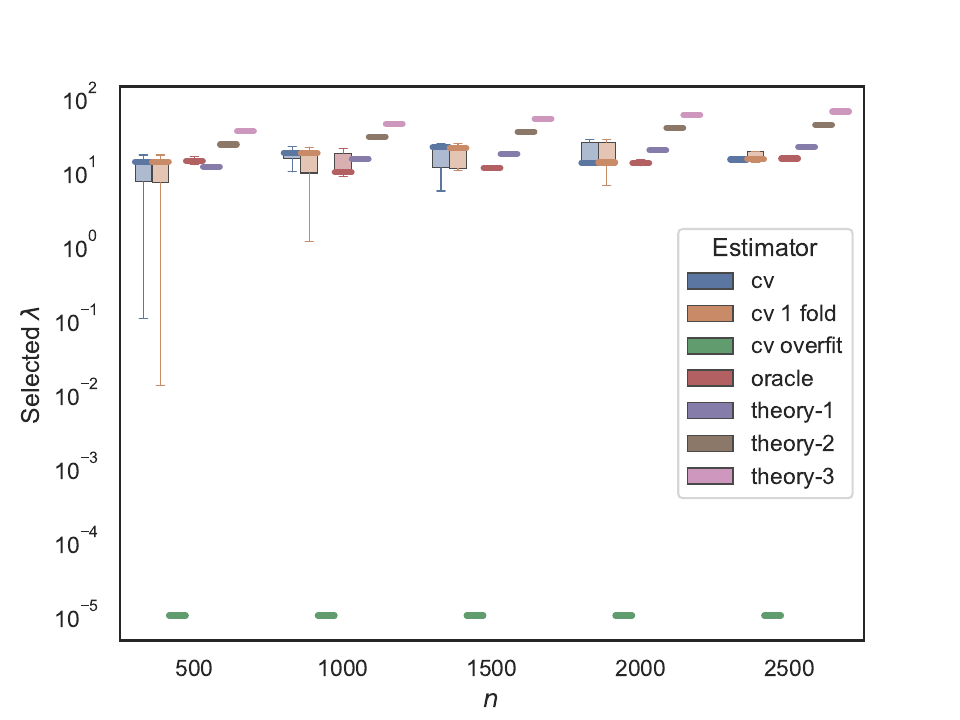}
		\caption{Synthetic, $d=100$, $\sigma=2$}
	\end{subfigure}
	\caption{Comparison of the selected penalty parameter $\lambda$ for the proposed estimators on movielens data, for varying noise level and cross-validation methodology.}
	\label{fig:lam-sel-full-real-data}
\end{figure}

In addition to the four estimators described in \cref{sec:simulations}, in this appendix we consider the following variants of the cross-validated estimator.
\begin{enumerate}
	\item \verb|cv 1 fold| follows the same procedure as the \verb|cv| estimator, expect that it uses only a single training and validation split. Specifically, this estimator is equal to $\Estimator_{-k}(\PenaltyCV')$ for a random $k$, $k\in[K]$, where 
	\[
	\PenaltyCV'\in\arg\min_{\lambda\in\Lambda}\Norm{Y_k-\SamplingOp_k(\Estimator_{-k}(\lambda))}_2^2\,.
	\]
	The motivation to consider this estimator goes back to \cref{rem:other-cv-forms} where we discussed that that \verb|cv 1 fold|  has the same theoretical properties as \verb|cv|, but one expects the latter to benefit from averaging and to perform better empirically.
	
	\item \verb|cv refit| follows the same procedure as the \verb|cv| estimator to choose the optimal penalty $\PenaltyCV$. However, it uses all of the data and refits the estimator. Specifically, \verb|cv refit| is a solution to \cref{eq:goodness-condition} when all of the observations $\{(\bX_i,y_i)\}_{i=1}^n$  and $\PenaltyCV$ are used. While  \verb|cv refit|
	is not covered by our theory (by \cref{rem:other-cv-forms}), it is a common way to perform cross-validation in practice. 
	
	\item \verb|cv overfit| is constructed as follows. For a sequence of $\Lambda$, as in \verb|cv| the estimator $\Estimator(\Penalty)$ is calculated using the entire observations $\{(\bX_i,y_i)\}_{i=1}^n$. Next, the penalty $\PenaltyCV''$ is selected such that the \emph{in-sample} error on the same set of observations is minimized. Thus,
	\[
    \PenaltyCV''\in\arg\min_{\lambda\in\Lambda}\Norm{Y-\SamplingOp(\Estimator(\lambda))}_2^2\,.
    \]
    The \verb|cv overfit| estimator $\Estimator(\PenaltyCV'')$ uses the same observations to fit $\Estimator$ and calculate the error, which is technically overfitted to the observations. The reason to include this estimator is to have a baseline for the effect of overfitting. 	
\end{enumerate}
We then repeat the simulations of \cref{sec:simulations} for all (seven) estimators, for three different noise levels, when the standard deviation of the noise belongs to $\{0.5, 1, 2\}$, and the results are presented in \crefrange{fig:rel-error-full}{fig:lam-sel-full-real-data}. The results are generally aligned with those in \cref{fig:rel-error} and \cref{fig:lam-sel}. However, a few interesting observations about various cross-validated estimators are worth highlighting: 
\begin{itemize}
	\item  \verb|cv refit|  performs very well in general and nearly ties with \verb|oracle|. But for small values of $n$, \verb|cv| slightly outperforms \verb|cv refit|, and this effect is more visible in the high-noise regime. 
	
	\item \verb|cv 1 fold| performs well compared to the theory inspired estimators, but generally loses to \verb|cv|, which underscores the benefit of averaging. It is clear from \crefrange{fig:lam-sel-full}{fig:lam-sel-full-real-data} that \verb|cv 1 fold| and \verb|cv| both select a penalty close to the one \verb|oracle| selects, but the selection has higher variance for \verb|cv 1 fold|, which could explain its empirical inferiority to \verb|cv|. 
	
	\item As expected, the penalty parameter selected by \verb|cv overfit| is close to $0$ due to overfitting. But the relative error of \verb|cv overfit| demonstrates a more nuanced behavior. The estimator performs very poorly in the high-noise regime (its relative error is above 1 and the curve is not visible in the plots), but in the low-to-medium noise regime, it performs well when $n$ is not too large. There are two explanations for this behavior. First, when $n$ is large, the impact of overfitting is increased as the estimator is perfectly fitting a larger number of noisy observations. Second, in the low-to-medium noise regime, the counterintuitively good performance of \verb|cv overfit| can be explained by the exact recovery results that are theoretically correct when the noise is zero. In these settings, the optimum penalty is indeed $0$; hence one can expect that, at $\lambda$ equal to $0$, the relative error only gradually deteriorates when the noise is increased.
	
\end{itemize}

\newpage


\bibliography{papers,books}
\bibliographystyle{plainnat}

\end{document}